\documentclass{article}

\PassOptionsToPackage{numbers, compress}{natbib}

\usepackage[preprint]{neurips_data_2023}

\usepackage[utf8]{inputenc} 
\usepackage[T1]{fontenc}    
\usepackage{hyperref}       
\usepackage{url}            
\usepackage{booktabs}       
\usepackage{amsfonts}       
\usepackage{nicefrac}       
\usepackage{microtype}      
\usepackage{xcolor}         
\usepackage{longtable}
\usepackage{graphicx}
\usepackage{array}
\usepackage{multirow}
\usepackage{xcolor,colortbl}
\usepackage{xspace}
\usepackage{makecell}
\usepackage{boldline}
\usepackage{pifont}
\usepackage{tabularx}
\usepackage{graphicx}
\usepackage{wrapfig}
\usepackage[export]{adjustbox}
\usepackage{fancyvrb}
\usepackage{fvextra}
\usepackage[most,breakable]{tcolorbox}
\usepackage{caption}
\usepackage{pgfplots}
\pgfplotsset{compat=newest}
\usepgfplotslibrary{units}

\newcommand{\ours}[0]{\textsc{OSWorld}\xspace}
\newcommand{\numexamples}[0]{369\xspace}
\newcommand{\numevals}[0]{134\xspace}
\newcommand{\numinitstates}[0]{302\xspace}
\newcommand{\numexampleswindows}[0]{43\xspace}
\newcommand{\cmark}{\textcolor{green}{\ding{51}}} 
\newcommand{\xmark}{\textcolor{red}{\ding{55}}} 

\title{
\ours: Benchmarking Multimodal Agents for Open-Ended Tasks in Real Computer Environments
}

\author{\textbf{Tianbao Xie}$^\spadesuit$  
        \textbf{Danyang Zhang}$^\spadesuit$
        \textbf{Jixuan Chen}$^\spadesuit$
        \textbf{Xiaochuan Li}$^\spadesuit$ 
        \textbf{Siheng Zhao}$^\spadesuit$\\  
        \textbf{Ruisheng Cao}$^\spadesuit$
        \textbf{Toh Jing Hua}$^\spadesuit$ 
        \textbf{Zhoujun Cheng}$^\spadesuit$
        \textbf{Dongchan Shin}$^\spadesuit$ 
        \textbf{Fangyu Lei}$^\spadesuit$ 
        \textbf{Yitao Liu}$^\spadesuit$\\ 
        \textbf{Yiheng Xu}$^\spadesuit$
        \textbf{Shuyan Zhou}$^\clubsuit$ 
        \textbf{Silvio Savarese}$^\heartsuit$
        \textbf{Caiming Xiong}$^\heartsuit$
        \textbf{Victor Zhong}$^\diamondsuit$ 
        \textbf{Tao Yu}$^\spadesuit$
       \\ 
  $^\spadesuit$The University of Hong Kong 
  \quad
  $^\clubsuit$CMU
  \quad
  $^\heartsuit$Salesforce Research 
  \quad
  $^\diamondsuit$University of Waterloo
  \vspace{-0.5cm}
}

\begin{document}

\maketitle

\begin{abstract} 
Autonomous agents that accomplish complex computer tasks with minimal human interventions have the potential to transform human-computer interaction, significantly enhancing accessibility and productivity.
However, existing benchmarks either lack an interactive environment or are limited to environments specific to certain applications or domains, failing to reflect the diverse and complex nature of real-world computer use, thereby limiting the scope of tasks and agent scalability. 
To address this issue, we introduce \ours, the \textit{first-of-its-kind scalable, real computer environment} for multimodal agents, supporting task setup, execution-based evaluation, and interactive learning across various operating systems such as Ubuntu, Windows, and macOS.
\ours can serve as a unified, integrated computer environment for assessing \textit{open-ended} computer tasks that involve arbitrary applications.
Building upon \ours, we create a benchmark of \numexamples computer tasks involving real web and desktop apps in open domains, OS file I/O, and workflows spanning multiple applications.
Each task example is derived from real-world computer use cases and includes a detailed initial state setup configuration and a custom execution-based evaluation script for reliable, reproducible evaluation.
Extensive evaluation of state-of-the-art LLM/VLM-based agents on \ours reveals significant deficiencies in their ability to serve as computer assistants.
While humans can accomplish over 72.36\% of the tasks, the best model achieves only 12.24\% success, primarily struggling with GUI grounding and operational knowledge.
Comprehensive analysis using \ours provides valuable insights for developing multimodal generalist agents that were not possible with previous benchmarks.
Our code, environment, baseline models, and data are publicly available at \url{https://os-world.github.io}.

\end{abstract}

\section{Introduction}
\vspace{-5pt}

Humans interact with computers to perform essential tasks in the digital realm, including web browsing, video editing, file management, data analysis, and software development.
These task workflows often involve multiple applications through graphical user interfaces (GUI) and command line interfaces (CLI). 
Autonomous digital agents, powered by advancements in large vision-language models (VLMs), have the potential to revolutionize how we interact with computer environments~\citep{licklider1960man, shi2017world, adept2022act1}. 
By following high-level natural language instructions, these agents can make digital interactions more accessible and vastly increase human productivity. 
However, a major challenge in developing such multimodal agents is the absence of a benchmark based on a real interactive environment that covers the diversity and complexity of real-world computer use across various operating systems, interfaces, and applications, consequently restricting task scope and agent scalability.

\begin{figure}[t]
    \centering
    \vspace{-0.8cm}
    \includegraphics[width=5.2in]{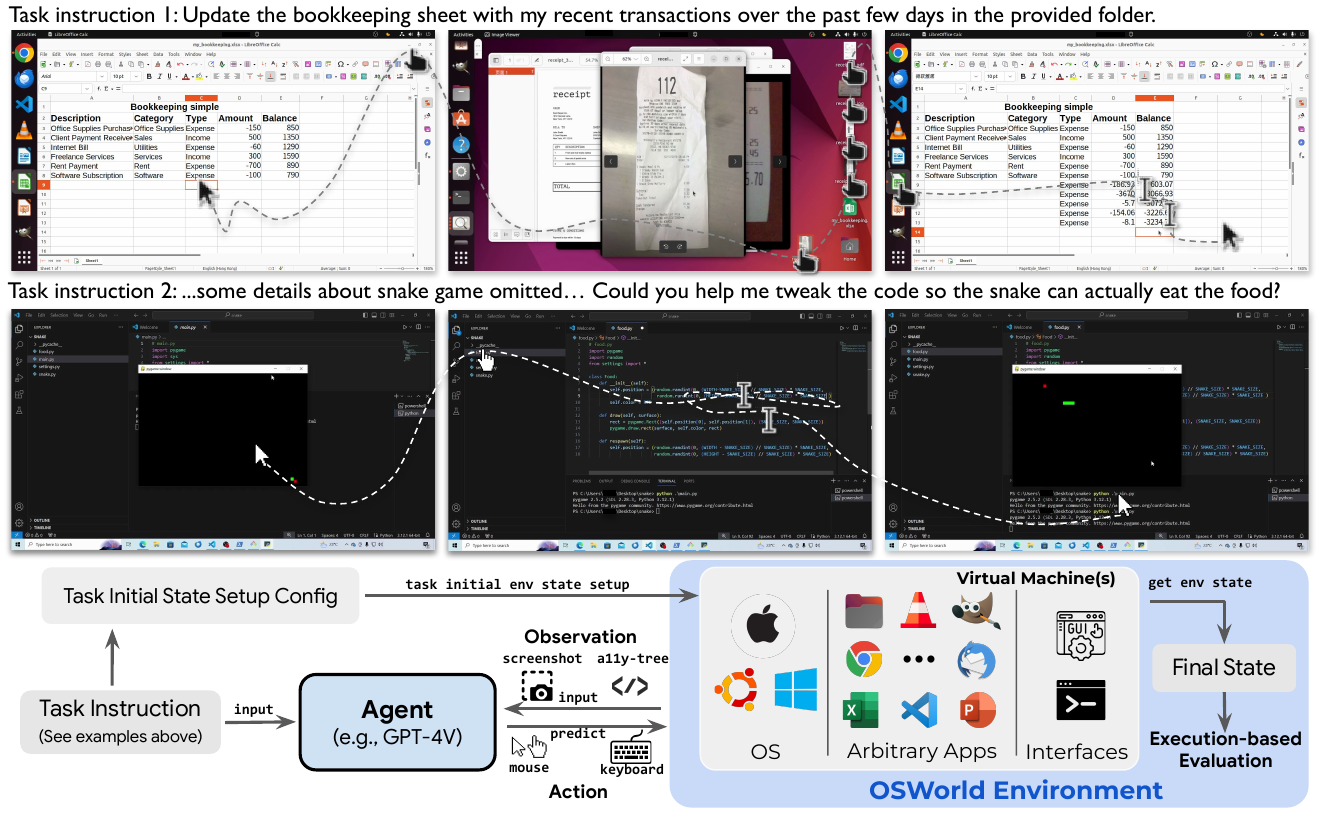}
    \caption{
    \ours is a \textit{first-of-its-kind scalable, real computer environment} for multimodal agents, supporting task setup, execution-based evaluation, and interactive learning across operating systems.
    It can serve as a unified environment for evaluating \textit{open-ended} computer tasks that involve arbitrary apps (e.g., task examples in the above Fig).
    We also create a benchmark of \numexamples real-world computer tasks in \ours with reliable, reproducible setup and evaluation scripts.
    }
    \label{fig:task_demonstrate}
    \vspace{-0.7cm}
\end{figure}

Previous benchmarks provide datasets of demonstrations without executable environments~\citep{deng2023mind2web, rawles2023android, Kapoor2024OmniACTAD}.
Their non-execution-based evaluation assumes a single solution for each task and wrongfully penalizes alternative correct solutions.
These benchmarks also miss opportunities for essential agent development methods like interactive learning and real-world exploration.
Building realistic interactive environments is a major challenge in developing multimodal agents.
Prior work that introduce executable environments simplify the observation and action spaces of human-computer interaction and limit task scope within specific applications or domains, such as web navigation in a few domains~\citep{shi2017world, liu2018reinforcement, yao2022webshop, zhou2023webarena}, coding~\citep{yang2023intercode} and the combination~\citep{liu2023agentbench, TianbaoXie2023_OpenAgents, ma2024agentboard}.
Agents developed in these restricted environments cannot comprehensively cover computer tasks, lacking the support of evaluating tasks in complex, real-world scenarios that require navigating between applications and interfaces in open domains (task examples in \textit{e.g.}, Fig.~\ref{fig:task_demonstrate}).

To address this gap, we introduce \ours, the \textit{first-of-its-kind scalable, real computer environment} designed for the development of multimodal agents capable of executing a wide range of real computer tasks beyond isolated interfaces and applications. 
This executable environment allows free-form raw keyboard and mouse control of real computer applications and supports initial task state configuration, execution-based evaluation, and interactive learning across mainstream operating systems (\textit{e.g.},~Ubuntu, Windows, macOS).
As shown in Fig.~\ref{fig:task_demonstrate}, \ours enables evaluation of \textit{open-ended} computer tasks that involve arbitrary applications, ranging from image viewing to software functionality integration and programming.
Thus, \ours can serve as a unified, real computer environment that allows users to define their agent tasks without the need to build application/domain-specific simulated environments.

Building upon \ours, we create a benchmark with \numexamples real-world computer tasks that involve widely-used web and desktop apps in open domains, OS file I/O, and multi-app workflows through both GUI and CLI.
Each task example is based on real-world computer use cases experienced by real users and often requires interactions with multiple applications and interfaces. 
To ensure reliable, reproducible assessment within the \ours environment, 9 authors with computer science backgrounds carefully annotate each example with an initial state setup configuration to simulate human work in progress and a custom execution-based evaluation script to verify task completion.
Our benchmark has a total of 134 unique evaluation functions, which are orders of magnitude larger than prior work~\citep{zhou2023webarena}, showcasing the complexity, diversity, and evaluation challenges of tasks in our benchmark.
The human performance study indicates that task examples from \ours are more time-consuming and challenging compared to those in prior work.

We extensively evaluate state-of-the-art LLM and VLM-based agent baselines, including the GPT-4V series~\citep{openai2023gpt}, the Gemini series~\citep{team2023gemini, reid2024gemini}, the Claude-3 Opus~\citep{claude3} and the Qwen-Max~\citep{bai2023qwen}, as well as Mixtral~\citep{jiang2024mixtral}, Llama-3~\citep{meta2024llama3} and CogAgent~\citep{hong2023cogagent} from the open-source community.
The performance of these experiments ranges from 0.99\% to 12.24\%, with subsets of applications even reaching 0\%,
for workflow tasks that involve cooperation from multiple apps, the highest performance of the baseline agent is only 6.57\%.
This indicates that current LLMs and VLMs are far from capable of serving as computer assistants (\S\ref{result-analysis}).
Results also show that while additional knowledge such as the accessibility tree and Set-of-Mark (\S\ref{baselines}) can be helpful, it can also lead to potential misguidance and varies across models.
We also observe performance changes in these agents compared to consistent human performance across different types of computer tasks.
Analysis reveals that VLM-based agents struggle to ground on screenshots to predict precise coordinates for actions, tend to predict repetitive actions, are unable to handle noise from unexpected application windows and exhibit limited knowledge of basic GUI interactions and domain-specific features of apps (\S\ref{sec:performance_by_mm_obs_variances}, \S\ref{qualitative_analysis}).
Feeding higher resolution and more trajectory history can help improve the performance by even doubling while requiring longer context length and efficient modeling (\S\ref{sec:performance_by_mm_obs_variances}).
We open-source \ours environment and benchmark, including environment initial state setup, reliable evaluation scripts, documentation, and our implementation of baseline models to promote research towards the goal of generalist capable computer agents~\footnote{\url{https://os-world.github.io}}.
Future work can focus on enhancing VLM GUI grounding abilities, including interaction commonsense knowledge, higher-resolution support, and coordinates accuracy for more robust GUI interactions. 
Additionally, efforts can be made to improve agent architectures to better handle complex computer tasks through exploration, memory, and reflection.

\section{\ours Environment}
\label{sec:environment}
\vspace{-5pt}

In this section, we will introduce the task definition of autonomous agents, the components and implementation of the \ours environment, and the supported observation and action spaces.

\subsection{Task Definition}
An autonomous digital agent task can be formalized as a partially observable Markov decision process (POMDP) $(\mathcal{S}, \mathcal{O}, \mathcal{A}, \mathcal{T}, \mathcal{R})$ 
with state space $\mathcal{S}$, observation space $\mathcal{O}$ (\S\ref{observation_space}, including natural language $\mathcal{I}$), action space $\mathcal{A}$ (\S\ref{action_space}), transition function $\mathcal{T}: \mathcal{S} \times \mathcal{A} \to \mathcal{S}$, and reward function $\mathcal{R}: \mathcal{S} \times \mathcal{A} \to \mathbb{R}$.
Given current observation $o_{t} \in \mathcal{O}$ (a natural language instruction observation and a screenshot observation (\textit{e.g.}, computer screenshot), accessibility (a11y) tree, or their combination according to facilities available), an agent generates executable action $a_t \in \mathcal{A}$ (\textit{e.g.}, clicking on the certain pixel of the screen --- \texttt{.click(300, 540, button=`right')}, press key combination --- \texttt{.hotkey(`ctrl', `alt', `t')}), which results in a new state $s_{t+1} \in \mathcal{S}$ (\textit{e.g.}, current Desktop environment) and a new partial observation $o_{t+1} \in \mathcal{O}$ (\textit{e.g.}, current screenshot).
The interaction loop repeats until an action that marks termination (\verb|DONE| or \verb|FAIL|, see Sec.~\ref{action_space}) is generated or the agent reaches the max number of steps (\textit{e.g.}, 15 in our experiments).
In this version of \ours, we implement an execution-based reward function $\mathcal{R}: \mathcal{S} \times \mathcal{A} \to [0, 1]$ (\S\ref{eval}). 
The reward function awards a value of 1 or a positive decimal under 1 at the final step if the state transitions meet the expectations of the task objective (i.e., the goal is successfully achieved or partially achieved), or if the agent accurately predicts failure for an infeasible task.
In all other scenarios, it returns 0.

\subsection{Real Computer Environment Infrastructure}
\vspace{-5pt}

\begin{figure}[htbp]
    \centering
    \includegraphics[width=\linewidth]{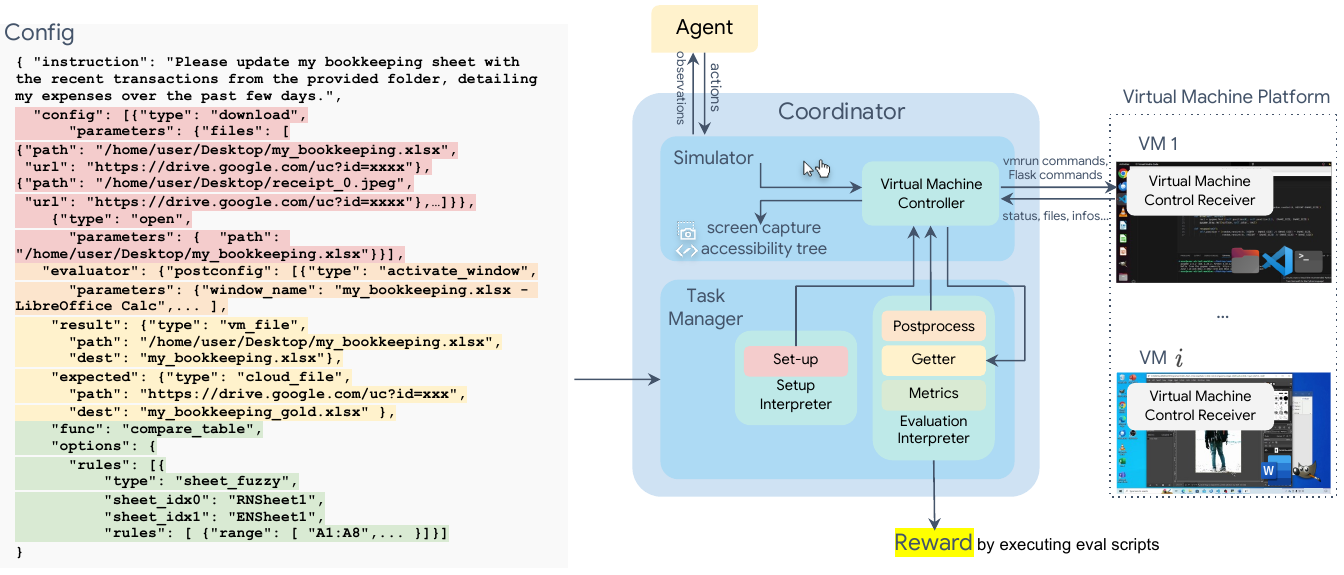}
    \caption{
    Overview of the \ours environment infrastructure. The environment uses a configuration file for initializing tasks (highlighted in red), agent interaction, post-processing upon agent completion (highlighted in orange), retrieving files and information (highlighted in yellow), and executing the evaluation function (highlighted in green). 
    Environments can run in parallel on a single host machine for learning or evaluation purposes. 
    Headless operation is supported.
    }
    \vspace{-15pt}
    \label{fig:arch_figure}
\end{figure}

\ours is an executable and controllable environment that supports task initialization, execution-based evaluation, and interactive agent learning in a range of \textit{real} operating systems (\textit{e.g.},~Ubuntu, Windows, macOS) using virtual machine techniques, shown in the middle and right of Fig.~\ref{fig:arch_figure}. 
Virtual machine offers a safe isolated environment and prevents the agent resulting in irreversible damaging effect on the real host machine. 
The snapshot feature also enables efficient reset of the virtual environment.
The environment is configured through a config file (shown in the left of Fig.~\ref{fig:arch_figure}) for interface initialization during the initialization phase (including downloading files, opening software, adjusting interface layout) (\S\ref{init_setup}, highlighted with red in Fig.~\ref{fig:arch_figure}), post-processing during the evaluation phase (activating certain windows, saving some files for easy retrieval of information, highlighted with orange), and acquiring files and information for evaluation (such as the final spreadsheet file for spreadsheet tasks, cookies for Chrome tasks, highlighted with yellow in Fig.~\ref{fig:arch_figure}), as well as the evaluation functions and parameters used (\S\ref{eval}, highlighted with green in Fig.~\ref{fig:arch_figure}).
See App.~\ref{app:env_infra} for more details.

\subsubsection{Overview}
\vspace{-5pt}

\ours environment runs on the host machine.
Its Coordinator accepts a configuration file at the initialization of a computer task, runs commands to automatically create a virtual machine instance, and initializes the required state for the task through the Task Manager.
The configuration file specifies the snapshot of the virtual machine to be used (which stores the complete state of a computer at a certain moment and can be restored to this state at any time) and also indicates the information needed for setup (such as downloading files and opening some software, making some additional settings, etc.).
Once the environment is set up, agents start to interact with the environment, receiving observations such as screenshots, the accessibility (a11y) tree, and customized streams such as terminal outputs.
Agents subsequently generate executable actions (\textit{e.g.},~\texttt{.click(300, 540)}) that manipulate the keyboard and mouse.
Each action of the agent is input into the environment as a code string, and the environment's Simulator executes them in the virtual machine. 
After the completion of a task, the Task Manager performs post-processing (such as file saving, or reopening certain apps) according to the task's post-config, retrieves data to the host machine (fetching images or configuration files from the virtual machine or cloud, etc.), and then runs evaluation scripts to assess the completion of the task.
Multiple virtual machines can run simultaneously on a single host machine, thereby parallelizing training and evaluation.

\subsubsection{Initial Task Environment Setup}
\label{init_setup}
Many real-world scenarios requiring assistance occur not at the beginning of digital activities, such as right after launching an application or when a computer has just been started, but rather at intermediate stages, such as when certain software is already open or the computer has experienced a crash. 
Therefore, we aim to simulate these intermediate states as closely as possible to replicate real-world scenarios.
The naturalness we bring in also leads to more challenges for agents to model and explore.
We adopted a hybrid approach for configuration instead of solely relying on example-wise snapshots for restoration since it would store much unnecessary hardware state information, resulting in each example requiring gigabytes of space.
The procedure is divided into three stages: start the VM emulator, prepare files (download the files or scripts from the cloud, \textit{etc.} optional), and execute reprocessing commands (open files or tabs, change the window size, \textit{etc.} optional).
We provide convenient APIs to configure initial conditions and world settings, standardizing our tasks to make this process user-friendly and easily extendable for scaling.
For more details on setup see App.~\ref{app:initial_state_setup_details}.

\subsubsection{Execution-Based Evaluation}
\label{eval}

Evaluating the successful execution of general computer tasks presents a significant challenge, as these tasks defy reduction to a uniform pattern or measurement by a single metric. 
To ensure a thorough assessment, we design example-specific evaluation metrics including pre-setup, post-processing, and dedicated functions, tailored to the software in use and the task's specific requirements.
This involves interpreting the software's internal files, utilizing specific packages, and preemptively setting up scaffolding based on the software's permissions (\textit{e.g.}, opening remote debugging ports for Chrome and VLC, creating extensions for VS Code). 
Occasionally, this process may also require assistance from reverse engineering tools, such as for decrypting account information in Thunderbird.

As a result, we construct a vast collection of functions that make final wrangling and retrieve files and data information of varying types, categories, and granularities from the cloud and software from virtual machines as well as evaluation functions covering different aspects and their combinations, inputting this information as parameters to assess the outcomes.
We show some evaluation examples in Tab.~\ref{tab:evaluation_examples}.
, demonstrate the retrieval of cookie data from virtual machines, obtaining files from both virtual machines and cloud services, fetching the current runtime interface's accessibility tree from the virtual machines, and determining success based on this information whether Amazon's cookies have been deleted, whether the generated table is accurate, and whether the correct interface has been accessed.
Need to note when the type of task has real-time characteristics (such as the number of citations of someone's paper, the content of blogs, \textit{etc}.), we include dynamic functions (such as crawler scripts) inside getter to obtain the real-time values at the moment of evaluation and then use them to compare with the results obtained by the agent upon task completion.
See more in App.~\ref{app:evaluation_configuration_details}.

\begin{table}[t]
\vspace{-10pt}
\caption{
Examples of our annotated evaluation scripts, which involve retrieving data from configuration files, the environment, and the cloud, and executing functions to assess functional correctness and obtain results.
The example-wise evaluation facilitates the diversity of tasks and reliable evaluation of complex, real-world, open-ended tasks.
}
\resizebox{\textwidth}{!}{
\begin{tabular}{lp{.35\textwidth}l}
  \toprule
  \textbf{Initial State} &   \textbf{Task Instruction}  & \textbf{Evaluation Script (Simplified)}  \\
  \midrule
  \multirow{3}{*}{\raisebox{-1cm}{\includegraphics[width=5cm]{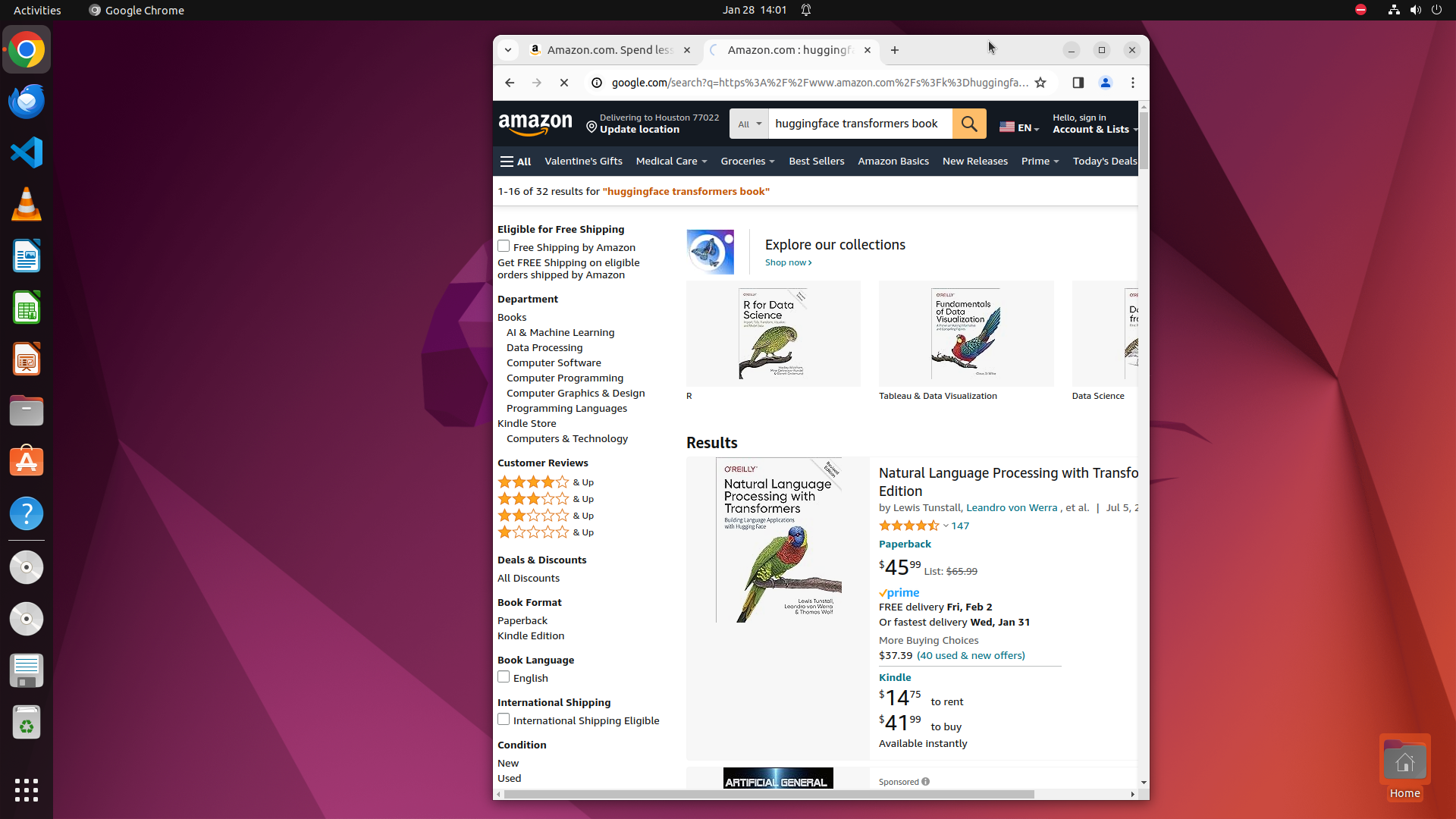}}}  & \multirow{3}{*}{\begin{minipage}{.35\textwidth}
  \textit{Can you help me clean up my computer by getting rid of all the cookies that Amazon might have saved?}
  \end{minipage}} &  \texttt{cookie\_data = get\_cookie\_data(env)}\\
    &   &  \texttt{rule = \{"type":"domains", }\\
    &   &  \texttt{"domains":[".amazon.com"]\}} \\
    &   &  \texttt{is\_cookie\_deleted(cookie\_data, rule)} \\ 
    &   &   \\
    &   &   \\
    &   &   \\
    &   &   \\
  \midrule
   \multirow{4}{*}{\raisebox{-1cm}{\includegraphics[width=5cm]{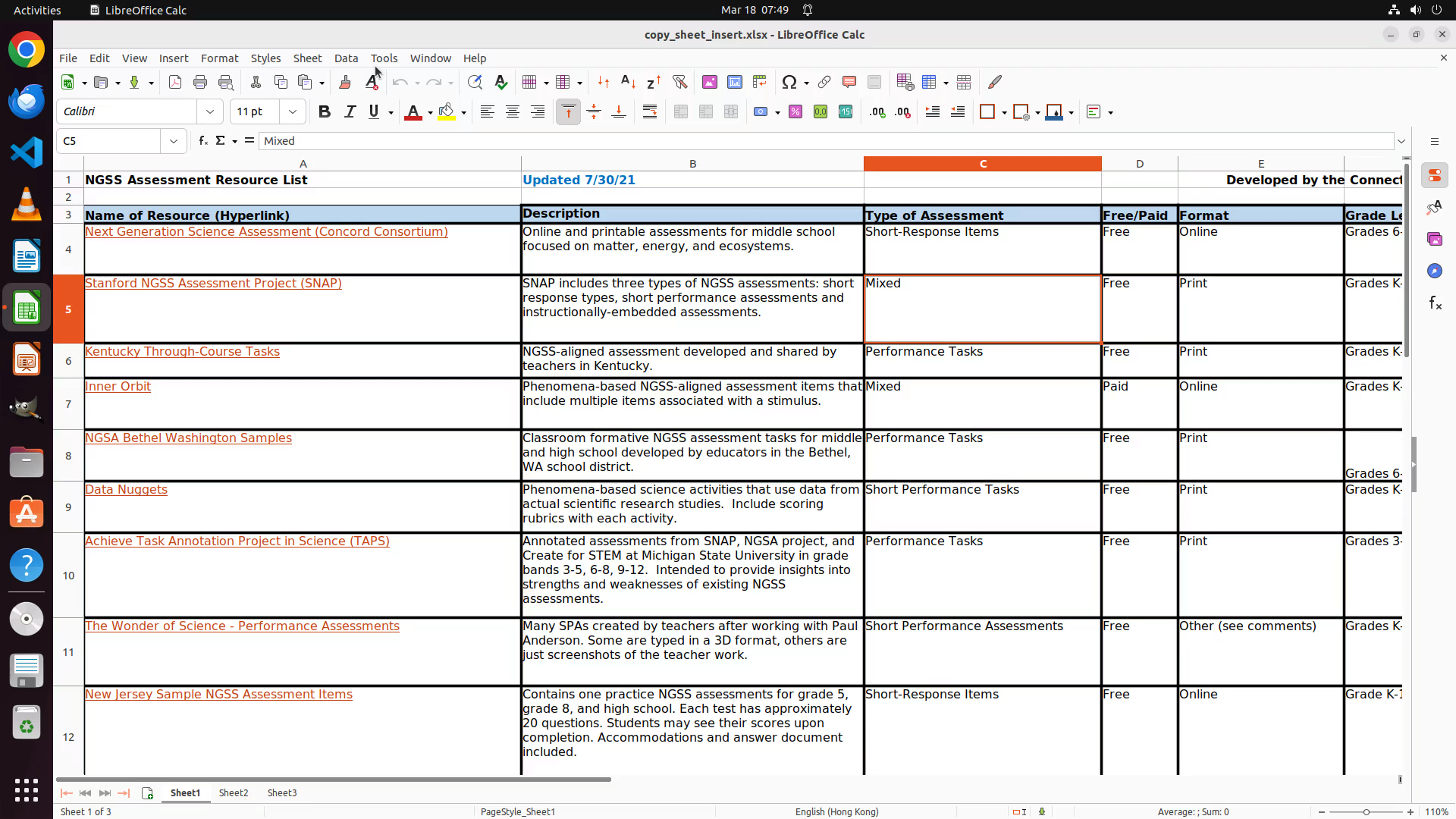}}} &  
   \multirow{4}{*}{
   \begin{minipage}{.35\textwidth}
   \textit{Rename ``Sheet 1'' to ``LARS Resources''. Then make a copy of it. Place the copy before ``Sheet 2'' and rename it by appending a suffix ``(Backup)'', ...}
   \end{minipage}}  & \texttt{result = get\_file(env)} \\
   &   & \texttt{expected = get\_file(cloud)} \\
   &   &  \texttt{rules = [\{"type":"sheet\_name"\},}\\
   &   &  \texttt{\ \  \  \  \  \  \  \ \ \{"type":"sheet\_data",} \\
   &   &  \texttt{\ \  \  \  \  \  \  \ \ \ "sheet\_idx0":0, } \\
   &   &  \texttt{\ \  \  \  \  \  \  \ \ \ "sheet\_idx1":1\}...]} \\
   &   &  \texttt{compare\_table(result, expected, rules)} \\
   &   & \\
    \midrule
    \multirow{4}{*}{\raisebox{-1cm}{\includegraphics[width=5cm]{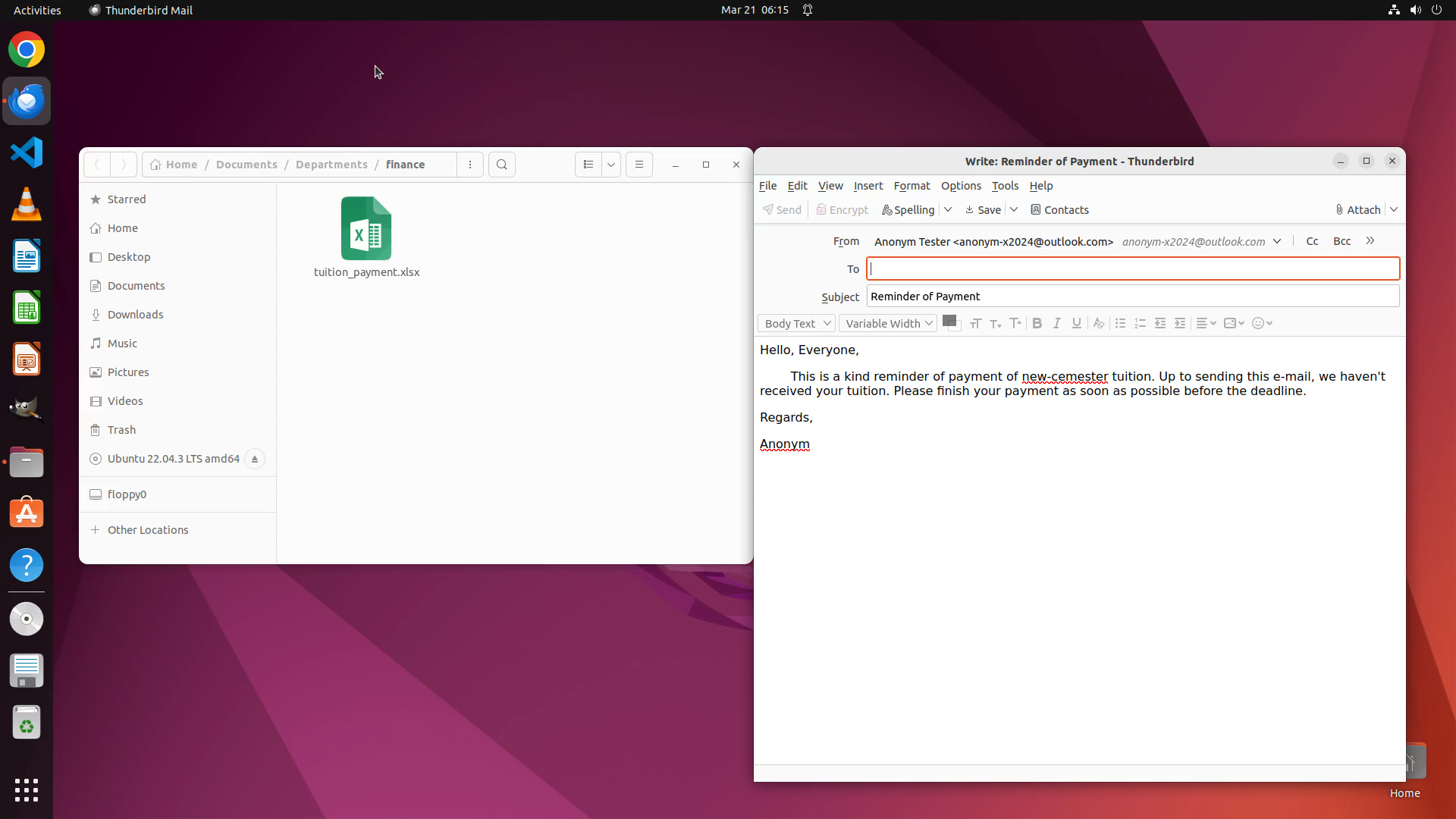}}} &  \multirow{4}{*}{
   \begin{minipage}{.35\textwidth}
   \textit{I've drafted an e-mail reminder for those who haven't paid tuition. Please help me to check out their e-mails from the payment record and add to the receiver field.}
   \end{minipage}} & \texttt{tree = get\_a11y\_tree(env)} \\
    &   & \texttt{rules = [\{"selectors": } \\
    &   & \texttt{\ \ ["tool-bar[attr|id=MsgHeadersToolbar]}\\ 
    &   & \texttt{\ \ label[name=To]~}\\
    &   & \texttt{\ \ [attr|class=\textbackslash"address-pill\textbackslash"]>}\\
    &   & \texttt{\ \ label[attr|class=\textbackslash"pill-label\textbackslash"]} \\
    &   & \texttt{\ \ [name*=\textbackslash"fox@someuniversity.edu...]} \\
    &   & \texttt{check\_a11y\_tree(tree, rules)} \\
  \bottomrule
\end{tabular}
}
\vspace{-18pt}
\label{tab:evaluation_examples}
\end{table}

\subsection{Observation Space}
\label{observation_space}
The observation space in \ours contains a \textbf{complete screenshot of the desktop screen}, including the mouse's position and shape, various application windows, files, and folders that are opened in different sizes and orders, maintaining the same perception as a human. 
Also, to be aligned with previous agent-building web and mobile research~\citep{liu2018reinforcement, li2020mapping, deng2023mind2web, zhou2023webarena} that provide and support the use of the webpage's DOM and app's view hierarchy, \ours also provides \textbf{XML-format accessibility (a11y) tree} (obtained via ATSPI~\footnote{\url{https://docs.gtk.org/atspi2/}} on Ubuntu, via PyWinAuto on Windows, \textit{etc.}), which can support additional information for modeling.
These raw observations allow rich interactions between multiple applications but induce challenges in long-horizon decision-making from high-resolution images (\textit{e.g.},~4k screenshots) and structured long text (\textit{e.g.},~accessibility trees).
For more detailed information on observation space, refer to App.~\ref{app:observation_space}.

\subsection{Action Space}
\label{action_space}

\begin{wraptable}{r}{8cm}
\vspace{-5pt}
\caption{Some examples of the mouse and keyboard actions $\mathcal{A}$ in \ours. 
See App.~\ref{app:action_space} for the complete list.}  
\centering
\resizebox{1.0\linewidth}{!}{%
\begin{tabular}{@{}ll@{}}
\toprule
Function & Description \\
\midrule
\texttt{moveTo(x, y)} & Moves the mouse to the specified coordinates. \\
\texttt{click(x, y)} & Clicks at the specified coordinates. \\
\texttt{write(`text')} & Types the specified text at the current cursor location. \\
\texttt{press(`enter')} & Presses the Enter key. \\
\texttt{hotkey(`ctrl', `c')} & Performs the Ctrl+C hotkey combination (copy). \\
\texttt{scroll(200)} & Scrolls up by 200 units. \\
\texttt{scroll(-200)} & Scrolls down by 200 units. \\
\texttt{dragTo(x, y)} & Drags the mouse to the specified coordinates. \\
\texttt{keyDown(`shift')} & Holds down the Shift key. \\
\texttt{keyUp(`shift')} & Releases the Shift key. \\
\texttt{WAIT} & Agent decides it should wait. \\
\texttt{FAIL} & Agent decides the task is infeasible. \\
\texttt{DONE} & Agent decides the task is finished. \\
\bottomrule
\end{tabular}
}
\vspace{-0.1in}
\label{tab:pyautogui_examples}
\end{wraptable}

Action space 
$\mathcal{A}$ in \ours encompasses all mouse and keyboard actions, including movement, clicks (left-key, right-key, multiple clicks), dragging, keystrokes, hotkeys, and others, covering all human-computer action space.
Some action examples are shown in Tab.~\ref{tab:pyautogui_examples} and the complete action list can be found in Appendix~\ref{app:action_space}. 
We use the widely used mouse and keyboard control library \texttt{pyautogui}\footnote{\url{https://pyautogui.readthedocs.io/en/latest/}} for our action space. 
This library leverages the high-level programming language Python to replicate and replay various human inputs into computers through code, allowing us to construct a universal and complete representation of actions. 
The agent must generate syntax-correct \texttt{pyautogui} Python code to predict valid actions. 
Basic actions, such as \texttt{press} and \texttt{moveTo}, can be integrated within program structures, such as for-loops, significantly improving the expressiveness of an action.
Timing is also crucial, as highlighted in previous studies on mobile devices~\citep{toyama2021androidenv}, as well as the ability to determine whether a task is infeasible or completed. 
Therefore, we add three special actions named \texttt{WAIT}, \texttt{FAIL}, and \texttt{DONE} to enhance the aforementioned action spaces.
Previous efforts towards creating domain-specific agents, such as MiniWoB++~\citep{shi2017world, liu2018reinforcement}, CC-Net~\citep{humphreys2022data}, and WebArena~\citep{zhou2023webarena, Koh2024VisualWebArenaEM}, have defined action spaces that include clicks and typing, as well as some actions specially designed for web browsing.
However, they do not model all possible actions on a computer, leading to limitations when attempting actions like right-clicking and clicking with the \texttt{ctrl} key held to select items.
This imposes an upper bound on agent learning capabilities.

\vspace{-5pt}
\section{\ours Benchmark}\label{sec:method}
\vspace{-5pt}

We introduce the \ours benchmark, which encompasses \numexamples real computing tasks defined and executed on Ubuntu. 
Additionally, we provide a set of \numexampleswindows tasks for Windows built on the \ours environment.~\footnote{Due to copyright issues, these Windows tasks require further activation by the user.}
The environment preparation, annotation process, data statistics, and human performance are described in this section.

\vspace{-5pt}
\subsection{Operating System and Software Environments}

\ours supports real operating systems, including Windows, macOS, and Ubuntu, for the development of automated computer agents. 
For development purposes, we offer an extensive set of examples on Ubuntu and its open-source applications, leveraging their open-source nature and more accessible APIs for task setting and evaluation. 
We also provide annotated testing examples for Windows, focusing on applications with similar functionalities.
For the first time, our real OS environments enable us to define all kinds of computer tasks, including those that involve interacting with multiple applications (e.g., Chrome and file manager) and interfaces (GUIs and CLIs).
Considering availability, the strength of the user community, and diversity, we mainly focus on eight representative applications as well as the basic ones system provide: 
Chrome for web browsing, VLC for media playback, Thunderbird for email management, VS Code as a coding IDE, and LibreOffice (Calc, Writer, and Impress) for handling spreadsheets, documents, and presentations respectively, GIMP for image editing, and other basic OS apps like terminal, file manager, image viewer, and PDF viewer.
Each example drawn from these applications separate or in combination showcases distinct operational logic and necessitates skills including commonsense knowledge, high-resolution perception, mastery of software shortcuts, and the precise controlling of mouse and keyboard movements.
For more details, check App.~\ref{app:os_selection} and~\ref{app:software_selection}.

\subsection{Tasks}
\vspace{-5pt}
We create a benchmark suite of \numexamples real-world computer tasks on Ubuntu environment collected from authors and diverse sources such as forums, tutorials, guidelines, \textit{etc.}, to show the capability for open-ended task creation within \ours.
Each example is carefully annotated with a natural language instruction, 
a setup configuration with corresponding files and setup actions for initialization of initial states upon our provided VM image, 
and a manually crafted evaluation script to check if the task is successfully executed.
We also adapt 43 tasks from the Ubuntu set for analytic usage on Windows.
Overall, it takes 9 computer science students (all student authors) over 3 months, consuming approximately 1800 man-hours (650 hours on single-app tasks, 750 hours on workflow tasks and 400 hours for double-checking).

\paragraph{Task instructions and scenarios}

To draw the most diverse and close-to-reality usage cases, we explore several types of resources, including official guidelines \& tutorials, video pieces giving tips and tutorials on the Internet (\textit{e.g.}, TikTok and YouTube), how-to websites (\textit{e.g.}, WikiHow), Q\&A forums (\textit{e.g.},
Reddit, Quora, Superuser, \& StackOverflow), formal video courses (\textit{e.g.}, Coursera and Udemy), and publicly-available personal blogs \& guidelines.
The detailed resources used in our benchmark are listed in 
App.~\ref{sub:task_exam_srcs}.
The examples are selected by judging their popularity, helpfulness, and diversity, revealed by the views and votes.
Meanwhile, we notice that it is challenging to find enough examples on the internet for tasks that involve the collaboration of multiple software applications. 
Therefore, the authors conducted extensive brainstorming, combining some existing examples or drawing inspiration from daily-life scenarios, to compile the tasks.
The instructions and task-related files are then crafted from these real-world guidelines and questions by the authors.
After the selection, each example will be cross-checked by the other two authors on the feasibility, ambiguity, and alignment with the source.
We not only collect tasks that can be finished, but also collect the infeasible ones that are inherently impossible to be completed due to deprecated features or hallucinated features raised by real users, which results in 30 infeasible examples in our benchmark.
Additionally, to demonstrate the unification ability of \ours environment for the creation of open-ended computer tasks, we also integrate 84 examples from other benchmarks focusing on single-application or domain-specific environments such as NL2Bash~\citep{lin2018nl2bash}, Mind2Web~\citep{deng2023mind2web}, SheetCopilot~\citep{li2023sheetcopilot}, PPTC~\citep{guo2023pptc}, and GAIA~\citep{mialon2023gaia}.
Refer to App.~\ref{app:examples_collection} for more details and~\ref{app:representitive_examples} for sampled examples for the showcase.
A total of about 400 man-hours were spent to collect these examples.

\paragraph{Initial state setup configs} 
To construct the initial state, we prepare the files required for the task and set up the initial state. 
For the files, we try to obtain them from the sources of the tasks we found, or, in cases where the files are not publicly available, we recreate them as realistically as possible based on scenarios. 
For the initial state setup, we also developed some functions based on the APIs of software and OS to control the opening and resizing of software windows and reimplement some functions that are difficult to achieve with APIs using \textit{pyautogui}. 
For different tasks, we write configs to set the files and initial steps in the virtual machine and verify them in the environment. 
For example, the setup stage (highlighted in red color, keyed as ``\verb|config|'') in 
Figure~\ref{fig:arch_figure} involves downloading files into the virtual machine
to prepare a close-to-reality initial environment, and then opening the
file of interest with the corresponding application.
The setup steps for each example take about 1 man-hours to construct.

\paragraph{Execution-based evaluation}
For each task, we select the appropriate getter functions, evaluator function, and parameters to compose the configuration file. 
The getter function is used to extract key components (\textit{e.g.}, the modified file, the text contents displayed in a window element) from the final state of the environment, and the evaluator function assesses success based on the extracted key components.
If a function does not exist, we will construct it and add it to the function library of the environment.
After completing each evaluation, the annotator conducts initial tests with self-designed test cases. 
Then, in the human evaluation and experiment running phases, each example is further scrutinized and iterated upon by different individuals three times from the perspective of alignment with the instruction and correctness under different solutions. 
As a result, we implement nearly sample-specific executable evaluation scripts, resulting in a total of \numevals unique evaluation functions for assessing functional correctness—significantly more than the previous benchmarks.
The average time spent on developing the evaluation for an example and its examination amounts to approximately 2 man-hours from graduate students.

\paragraph{Quality control}
Once annotation is finished, each example is attempted by two authors who did not participate in annotating that specific example, acting as agents to complete the task. 
This process evaluates the current example's quality and provides feedback to the annotators (such as unclear instructions or inability to complete the task, crashes in corner cases, serious instances of false positives and negatives, \textit{etc}.), and involves joint revisions and supplements. 
During experiments for human performance and baselines, we further fixed examples found to have issues, dedicating over 400 man-hours for four rounds of checks.
Further investment of time and a more red teaming could further reduce false positives and negatives, which we will leave to future work.

\subsection{Data Statistics}
\vspace{-5pt}

\begin{figure}[ht]
\vspace{-25pt}
    \centering
    \begin{minipage}{0.4\textwidth}
        \centering
        \captionof{table}{
        Key statistics in \ours. 
        The ``Supp. tasks'' refers to the Windows-based tasks, that could only be used after activation due to copyright restrictions.
        }
        \small
        \begin{tabular}{lc}
            \toprule
            \textbf{Statistic} & \textbf{Number} \\
            \midrule
            Total tasks (Ubuntu) & 369 (100\%) \\
            - Multi-App Workflow & 101 (27.4\%)\\
            - Single-App & 268 (72.6\%) \\
            - Integrated & 84 (22.8\%)\\
            - Infeasible  &30 (8.1\%)\\
            Supp. tasks (Windows) & \numexampleswindows \\
            \midrule
            Initial States & \numinitstates \\
            Eval. Scripts & \numevals \\
            \bottomrule
        \end{tabular}
        \label{tab:key_statistics}
    \end{minipage}
    \hspace{5mm}
    \begin{minipage}{0.50\textwidth}
        \centering
        \includegraphics[width=0.9\textwidth]{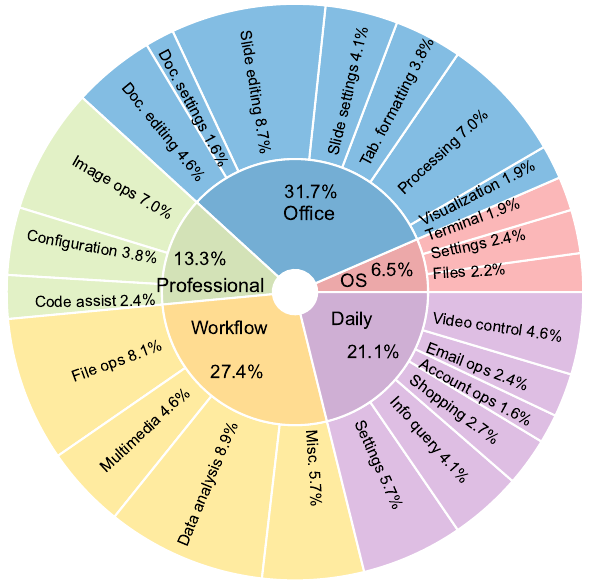}
        \caption{Distribution of task instructions in \ours based on the app domains and operation types to showcase the content intuitively.
        }
        \label{fig:instruction_pie_chart}
    \end{minipage}
    \vspace{-20pt}
\end{figure}

\paragraph{Statistics}
To facilitate the analysis and comprehension of the agent's capabilities, we cluster the examples into the software categories. 
Specifically, these categories include OS, Office (LibreOffice Calc, Impress, Writer), Daily (Chrome, VLC Player, Thunderbird), Professional (VS Code and GIMP), and Workflow (tasks involving multiple apps).
The main statistics of \ours are presented in Tab.~\ref{tab:key_statistics} and Fig.~\ref{fig:instruction_pie_chart}, showcasing the outline and a broad spectrum of tasks.
Specifically, \ours contains a total of 369 tasks (and an additional 43 tasks on Windows for analysis), with the majority (268 tasks or 72.6\%) aiming at single application functionalities and a remarkable section of workflow-related tasks (101 tasks or 27.4\%). 
The dataset's diversity is further affirmed by the inclusion of tasks considered infeasible, totaling 30 tasks or 8.1\% of the dataset.
Additionally, a total of 84 tasks (22.8\%) are integrated from related datasets, highlighting the dataset's applicability in universal modeling.
Remarkably, the dataset incorporates \numinitstates distinct initial states and \numevals different 
evaluation scripts, underscoring the comprehensive approach towards evaluating the tasks' complexity and requirements.
More statistic details are available in App.~\ref{app:examples_collection}.

\begin{table}[ht]
\centering
\caption{
Comparison of different environments for benchmarking digital agents. 
The columns indicate: 
the number of task instances and templates (if applicable) where the task instantiated from templates through configurations (\# Instances (\# Templates)), 
whether they provide a \emph{controllable} executable environment (Control. Exec. Env.), 
the ease of adding new tasks involving arbitrary applications in open domains  (Environment Scalability), support for multimodal agent evaluation (Multimodal Support), support for and inclusion of cross-app tasks (Cross-App), capability to start tasks from an intermediate initial state (Intermediate Init. State), and the number of execution-based evaluation functions (\# Exec.-based Eval. Func.).
}
\label{tab:benchmark_comparsion}
\resizebox{\textwidth}{!}{%
\begin{tabular}{@{}lccccccccc@{}}
\toprule
& \shortstack{\# Instances\\ (\# Templates)} & \shortstack{Control. \\ Exec. Env.?} & \shortstack{Environment\\Scalability?}  & \shortstack{Multimodal\\Support?} & \shortstack{Cross-\\App?} & \shortstack{Intermediate\\Init. State?} & \shortstack{\# Exec.-based\\ Eval. Func. } \\ \midrule
\textsc{GAIA}~\citep{mialon2023gaia} & 466 & \xmark & - & \xmark & \xmark & \xmark & 0 \\
\textsc{Mind2Web}~\citep{deng2023mind2web} & 2350 & \xmark & - & \cmark & \xmark & \cmark & 0 \\
\textsc{WebLINX}~\citep{lu2024weblinx} & 2337 & \xmark & - & \cmark & \xmark & \cmark & 0 \\
\textsc{PixelHelp}~\citep{li2020mapping} & 187 & \xmark & - & \cmark & \xmark & \xmark & 0 \\
\textsc{MetaGUI}~\citep{sun2022meta} & 1125 & \xmark & - & \cmark & \xmark & \xmark & 0 \\
\textsc{AitW}~\citep{rawles2023android} & 30$k$ & \xmark & - & \cmark & \xmark & \cmark & 0 \\
\textsc{OmniAct}~\citep{Kapoor2024OmniACTAD} & 9802 & \xmark & - & \cmark & \xmark  & \cmark & 0 \\
\midrule
\textsc{AgentBench}~\citep{liu2023agentbench} & 1091 & \small Multi-isolated & \xmark & \xmark & \xmark & \xmark & 7 \\
\textsc{InterCode}~\citep{yang2023intercode} & 1350 (3) & Code & \xmark & \xmark & \xmark & \xmark & 3 \\
\textsc{MiniWoB++}~\citep{liu2018reinforcement} & 125 & Web & \xmark & \cmark & \xmark & \xmark & 125 \\
\textsc{WebShop}~\citep{yao2022webshop} & 12$k$ (1) & Web & \xmark & \cmark & \xmark & \xmark & 1  \\
\textsc{WebArena}~\citep{zhou2023webarena} & 812 (241) & Web & \xmark & \cmark & \xmark & \xmark & 5 \\
\textsc{VWebArena}~\citep{Koh2024VisualWebArenaEM} & 910 (314) & Web & \xmark & \cmark & \xmark & \xmark & 6 \\
\textsc{WorkArena}~\citep{drouin2024workarena} & 23$k$ (29) & Web & \xmark & \cmark & \xmark & \cmark & 7 \\
\textsc{WikiHow}~\citep{zhang2023mobile} & 150 (16) & Mobile & \xmark & \cmark & \xmark & \xmark & 16 \\
\textsc{AssistGUI}~\citep{gao2023assistgui} & 100 & \xmark & \xmark & \cmark & \xmark & \cmark & 2 \\
\midrule
\ours & 369 & Computer & \cmark & \cmark & \cmark & \cmark & \textbf{\numevals} \\ \bottomrule
\end{tabular}
}
\vspace{-20pt}
\end{table}

\paragraph{Comparison with existing benchmarks}
\ours is compared with a number of existing benchmarks in Table~\ref{tab:benchmark_comparsion}.
\ours take utilizes raw mouse and keyboard actions that is universal to the computer environment, rather than focusing on specific computer applications~(\textit{e.g.,} a browser~\cite{zhou2023webarena, deng2023mind2web}), with multimodal observation including screenshot~(Multimodal Support column).
This universal action space enables the constructed agents to handle general tasks in the digital world. 
Our executable environment allows agents to freely explore during both the learning and evaluation phases, rather than providing only static demonstrations to evaluate an agent's prediction of the next step (Executable Env. column). 
Moreover, it does not solely focus on interactions within a single app but also considers interactions across multiple apps and the overall task (Cross-App column).
Unlike many evaluations that offer the same evaluation script or a few scripts for a certain type of task, the \ours benchmark provides example-wise, execution-based evaluation for tasks. 
Specifically, the total of \numevals unique execution-based evaluation functions in our benchmark is significantly more than previous work, demonstrating the complexity, diversity, and evaluation challenges of tasks in our benchmark (\# Exec.-based Eval. Func. column).
It also allow us to freely choose open-ended tasks and scale to new environments, rather than struggling in crafting new ones.
Constructing intermediate initial states as task setup increases realism and poses challenges to the agents' exploration capabilities (Intermediate Init. State column).

\subsection{Human Performance}

\begin{wrapfigure}{r}{0.5\textwidth}
    \centering
    \includegraphics[width=0.5\textwidth]{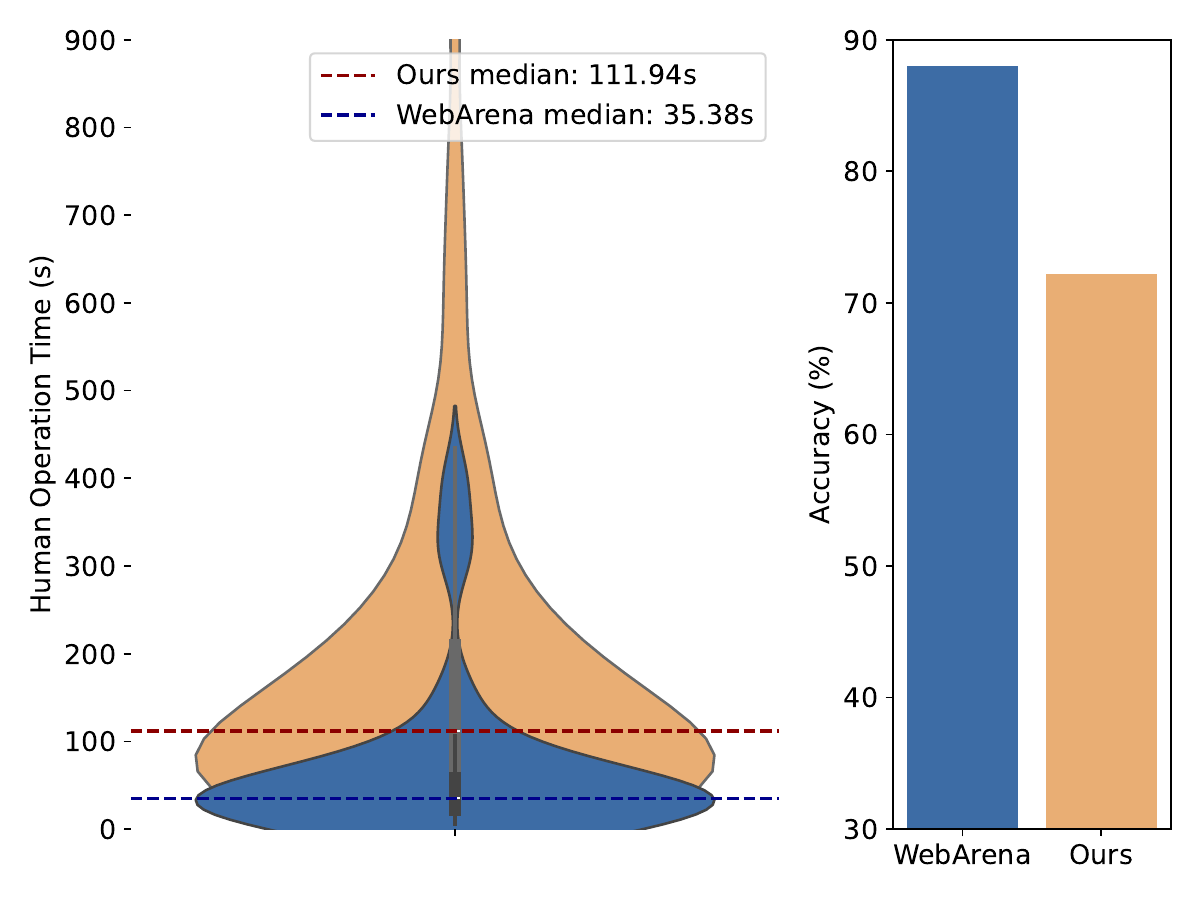}
    \vspace{-15pt}
    \caption{Human operation time and accuracy on \ours and WebArena. 
    }
    \label{fig:human_eval}
    \vspace{-15pt}
\end{wrapfigure}

We conduct human evaluations on each example in our dataset, with annotators being computer science major college students who possess basic software usage skills but have not been exposed to the samples or software before. 
We recorded the time required to complete each example and whether their completion of the example was correct. 
For comparison, we also sampled 100 examples from WebArena~\citep{zhou2023webarena} under the same evaluation setup.

As illustrated, tasks from our dataset generally required more time to complete, with a median completion time of 111.94 seconds (compared to 35.38 seconds in WebArena), and a significant number of examples distributed at 900 seconds or even more. 
In terms of accuracy, the human performance on our tasks was approximately 72.36\%, significantly lower than the 88\% observed on the pure web task dataset.
These findings highlight the complexity and challenge of tasks in our dataset, which demand more time and effort. 
The lower accuracy rate further indicates that our tasks require a higher level of understanding and proficiency, underscoring the need for advanced models and techniques to tackle them effectively.

\section{Benchmarking LLM and VLM Agent Baselines}\label{sec:experiments}
\vspace{-10pt}

In this section, we present the implementation details and experimental settings for several state-of-the-art LLM and VLM agent baselines on \ours benchmark, as well as their performance.

\begin{table*}[ht]
\centering
\vspace{-15pt}
\caption{
Success rates of baseline LLM and VLM agents on \ours, grouped by task categories: OS, Office (LibreOffice Calc, Impress, Writer), Daily (Chrome, VLC Player, Thunderbird), Professional (VS Code and GIMP) and Workflow (tasks involving multiple apps), for gaining insights from interfaces and operation logic.
See App.~\ref{app:prompting_details} and \ref{app:full_results} for more details.
}
\label{tab:exp_baseline}
\small
\scalebox{0.9}{
\renewcommand{\arraystretch}{1.1} 
\begin{tabularx}{\linewidth}{>{\centering\arraybackslash}p{2cm}>{\centering\arraybackslash}p{2.2cm}>{\centering\arraybackslash}X>{\centering\arraybackslash}X>{\centering\arraybackslash}X>{\centering\arraybackslash}X>
{\centering\arraybackslash}X>{\centering\arraybackslash}X>{\centering\arraybackslash}X}
\toprule
\multirow{2}{*}{\textbf{Inputs}} & \multirow{2}{*}{\textbf{Model}} & \multicolumn{6}{c}{\textbf{Success Rate (↑)}} \\ 
\cline{3-8}
~ & ~ & {OS} & {Office} & {Daily} & {Profess.} & \textbf{Workflow} & \textbf{Overall} \\
\midrule
A11y tree & Mixtral-8x7B  & 12.50\% & 1.01\% & 4.79\% & 6.12\% & 0.09\% & 2.98\% \\
& Llama-3-70B & 4.17\% & 1.87\% & 2.71\% & 0.00\% & 0.93\% & 1.61\% \\
& GPT-3.5 & 4.17\% & 4.43\% & 2.71\% & 0.00\% & 1.62\% & 2.69\% \\
& GPT-4 & 20.83\% & 3.58\% & 25.64\% & 26.53\% & 2.97\% & \textbf{12.24\%} \\
& Gemini-Pro & 4.17\% & 1.71\% & 3.99\% & 4.08\% & 0.63\% & 2.37\% \\
& Gemini-Pro-1.5 & 12.50\% & 2.56\% & 7.83\% & 4.08\% & 3.60\% & 4.81\% \\
& Qwen-Max & 29.17\% & 3.58\% & 8.36\% & 10.20\% & 2.61\% & 6.87\% \\
& GPT-4o   & 20.83\% & 6.99\% & 16.81\% & 16.33\% & \textbf{7.56\%} & 11.36\% \\
\midrule
Screenshot & CogAgent & 4.17\% & 0.85\% & 2.71\% & 0.00\% & 0.00\% & 1.11\% \\
& GPT-4V & 12.50\% & 1.86\% & 7.58\% & 4.08\% & \textbf{6.04\%} & 5.26\% \\
& Gemini-ProV & 8.33\% & 3.58\% & 6.55\% & 16.33\% & 2.08\% & \textbf{5.80\%} \\
& Gemini-Pro-1.5 & 12.50\% & 6.99\% & 2.71\% & 6.12\% & 3.60\% & 5.40\% \\
& Claude-3-Opus & 4.17\% & 1.87\% & 2.71\% & 2.04\% & 2.61\% & 2.42\% \\
& GPT-4o   & 8.33\% & 3.58\% & 6.07\% & 4.08\% & 5.58\% & 5.03\% \\
\midrule
Screenshot & CogAgent & 4.17\% & 0.85\% & 2.71\% & 0.62\% & 0.09\% & 1.32\%\\
+ A11y tree & GPT-4V & 16.66\% & 6.99\% & 24.50\% & 18.37\% & 4.64\% & \textbf{12.17\%} \\
& Gemini-ProV & 4.17\% & 4.43\% & 6.55\% & 0.00\% & 1.52\% & 3.48\%\\
& Gemini-Pro-1.5 & 12.50\% & 3.58\% & 7.83\% & 8.16\% & 1.52\% & 5.10\% \\
& Claude-3-Opus & 12.50\% & 3.57\% & 5.27\% & 8.16\% & 1.00\% & 4.41\% \\
& GPT-4o   & 41.67\% & 6.16\% & 12.33\% & 14.29\% & \textbf{7.46\%} & 11.21\% \\
\midrule
Set-of-Mark & CogAgent & 4.17\% & 0.00\% & 2.71\% & 0.00\% & 0.53\% & 0.99\% \\
& GPT-4V & 8.33\% & 8.55\% & 22.84\% & 14.28\% & \textbf{6.57\%} & \textbf{11.77\%} \\
& Gemini-ProV & 4.17\% & 1.01\% & 1.42\% & 0.00\% & 0.63\% & 1.06\% \\
& Gemini-Pro-1.5 & 16.67\% & 5.13\% & 12.96\% & 10.20\% & 3.60\% & 7.79\% \\
& Claude-3-Opus & 12.50\% & 2.72\% & 14.24\% & 6.12\% & 4.49\% & 6.72\% \\
& GPT-4o   & 20.83\% & 3.58\% & 3.99\% & 2.04\% & 3.60\% & 4.59\% \\
\midrule
\rowcolor{gray!20}
\multicolumn{2}{c}{Human Performance}  & 75.00\% & 71.79\% & 70.51\% & 73.47\% & 73.27\% & 72.36\% \\
\bottomrule
\end{tabularx}
}
\vspace{-15pt}
\end{table*}

\subsection{LLM and VLM Agent Baselines}
\label{baselines}

We adopt state-of-the-art LLM and VLM from open-source representatives such as Mixtral~\citep{jiang2024mixtral}, CogAgent~\citep{hong2023cogagent} and Llama-3~\citep{meta2024llama3}, and closed-source ones from GPT, Gemini, Claude and Qwen families on \ours, to serve as the foundation of agent.
We also explore methods such as the Set-of-Marks aided approach~\citep{yang2023set, gpt4vact}, which has been demonstrated to improve spatial capabilities for visual reasoning.
Our prior experiments following VisualWebArena~\citep{Koh2024VisualWebArenaEM} adopt few-shot prompting, which involves using (observation, action) pairs as few-shot examples and inputting the current observation to generate the action, but this resulted in poor performance (success rate of 2.79\% under pure-screenshot setting).
We attribute the result to a lack of history encoding and change in the prompting scheme.
Therefore, in the experiments, we opt to utilize the context window by providing the most recent 3 observations and actions in chat mode, \textit{i.e.}, alternating between ``user'' prompts and ``assistant'' prompts, instead of the (observation, action) pairs.
We use a temperature of 1.0 and top-p of 0.9 and truncate from the beginning of the input if still exceeding the max tokens limit required by the models.
The prompts used in the experiments are provided in App.\ref{app:prompting_details}.
We heuristically request the agents to complete the tasks within a max step limit of 15, which is
enough for most tasks.
We present a summary of the results in Tab.~\ref{tab:exp_baseline} and analysis in Sec.~\ref{result-analysis}.
We implement the following four types of input settings on LLM and VLM.

\paragraph{Accessibility tree}
We aim to evaluate whether the current advanced text-based language models can reason and ground themselves in the context to generate the correct action. 
Since the original XML format of accessibility tree contains millions of tokens, caused by countless
elements, redundant attributes, and a mass of markups, we opt to filter out
non-essential elements and attributes, and represent the elements in a more compact
tab-separated table format. To be specific, we filter the elements by their 
tag, visibility, availability, existence of text or image contents, \textit{etc}.
The detailed filtering method is elaborated on in 
App.~\ref{app:a11y_tree_handling}.
Only the \textit{tag}, \textit{name}, \textit{text}, \textit{position}, and \textit{size} of
the remaining elements are kept and concatenated by tab character in the input.
As the raw coordinates are provided within the accessibility tree, the LLM is
required to ground its action predictions to accurate coordinates.

\paragraph{Screenshot}
This is the input format that is closest to what humans perceive.
Without special processing, the raw screenshot of the 
virtual machine is directly sent to the VLM. The VLM is to 
understand the screenshot and predict
correct actions with precise coordinates. 
The raw resolution of the screen is set to $1920\times 1080$. 
In order to investigate the impact of input resolution, ablation studies are also conducted with different resolutions by manually downsampling the screenshot.

\paragraph{Screenshot + accessibility tree}
To check if a combination with the accessibility tree can improve the capacity of VLM for spatial grounding, we take this setting by inputting both raw screenshots and a simplified accessibility tree.

\paragraph{Set-of-Marks}
Set-of-Marks (SoM)~\citep{yang2023set} is an effective method for enhancing the grounding capabilities of VLMs such as GPT-4V, by segmenting the input image into different sections and marking them with annotations like alphanumerics, masks, or boxes.
We leverage the information from the filtered accessibility tree and mark the 
elements on the screenshot with a numbered bounding box.
Following VisualWebArena~\citep{Koh2024VisualWebArenaEM} and 
UFO~\citep{Zhang2024UFOAU}, we further combine
the annotated screenshot with the text metadata from accessibility tree, including
the \textit{index}, \textit{tag}, \textit{name}, and \textit{text} of the 
elements\footnote{This metadata is similar to but kind of different from that provided in
the single a11y tree setting. To be specific, the coordinates and size are replaced with element
index.}.
Instead of predicting precise coordinates, the VLM is supposed to specify
the action object by its number index, which will be 
mapped into our action space by post-processing. Ablation studies are also
conducted with different resolutions for SoM setting.

\subsection{Results}
\label{result-analysis}
\vspace{-10pt}
\paragraph{LLMs and VLMs are still far from being digital agents on real computers.}
The results from Table~\ref{tab:exp_baseline} show that when only using screenshots as input and adopting \texttt{pyautogui} as the code space, the success rate of the model is only 5.26\% to 5.80\% even with the strongest VLMs GPT-4V and Gemini-Pro-vision.
Meanwhile, the most advanced batch of language models, when using the a11y tree as input, has a success rate ranging from 2.37\% to 12.24\%.
Overall, these figures of performance are significantly lower than the human-level performance which is 72.36\% overall for individuals not familiar with the software.
These gaps indicate that current LLMs and VLMs may still have a significant gap from humans in performance, necessitating further research in this area.
Another surprising finding is that although Claude-3 Opus is reported to be competitive with GPT-4V on common benchmarks~\citep{Anthropic2023Claude}, it falls far behind when used as a digital agent in \ours. 
We will present a qualitative analysis and infer reasons in Sec.~\ref{qualitative_analysis}.

\paragraph{Agent performance has much higher variance than human across different types of computer tasks.}
\ours is capable of simulating and evaluating the various software types and combination scenarios involved in people's daily lives in an open-ended manner. 
We observe performance based on software type grouping and find that agents based on LLMs show significant differences across different subsets. 
As shown in Table~\ref{tab:exp_baseline}, performance tends to be better in tasks oriented towards CLI interfaces (such as OS-type tasks) compared to those based on GUI (such as Office tasks involving clicks on spreadsheet interfaces and document processing). 
Moreover, the biases between different models and settings are inconsistent, with gaps even exceeding 20\%; another point is that performance on workflow-type tasks involving multiple software is far below the figures on a single software, generally below 5\%. 
However, human performance is consistent across these tasks, fluctuating around 70\% without exceeding a 5\% variance, forming a significant contrast with the models. 
This suggests that the way humans understand and complete tasks may differ significantly from the current logic and methods based on LLMs and VLMs.

\paragraph{A11y tree and SoM's effectiveness varies by models.} 
The a11y tree contains some attribute information of visible elements, including window position and size, as well as some semantic labels of the window. 
The performance gap illustrated in Table~\ref{tab:exp_baseline} between GPT-4V and Claude-3 with additional a11y tree information and under a pure screenshot setup suggests that it still has significant room for improvement in accurately perceiving and reasoning GUI elements.
Conclusions are reversed for Gemini-Pro.

While applying SoM setting, there is a decline for GPT-4V in performance compared to directly providing the model with screenshots and a11y tree inputs, which contradicts the widely shown effectiveness of SoM in classic image understanding tasks~\citep{yang2023set}, as well as in application areas like web agents \citep{zheng2024gpt, he2024webvoyager}. 
We speculate that this is due to the tasks performed within operating systems having higher resolution and much more elements, (\textit{e.g.}, the cells in a spread table), leading to a significant amount of noise that counteracts the auxiliary role of bounding boxes. 
Some tasks also require detailed operation on coordinate-level, which cannot be modeled by the bounding box that SoM marks.

\paragraph{VLM agents with screenshot-only setting show lower performance, but it should be the ultimate configuration in the long run.}
The setting that relies solely on screenshots exhibits the lowest performance, at only 5.26\%, among all. 
Surprisingly, it still achieves a decent outcome when managing workflow tasks (involving multiple applications) that involve multiple applications. 
Despite the performance, it is worth mentioning that this is the only configuration that does not require additional information, such as an accessibility (a11y) tree, making it concise and in alignment with intuitive human perception since the a11y tree may not be well-supported across all software or cannot be obtained under noisy conditions (\textit{e.g.}, when the agent is restricted to viewing the computer through peripheral screens), and the massive amount of tokens contained in the a11y tree (even just the leaf nodes can have tens of thousands of tokens) can also impose an additional inference burden on the model.
Future work on purely vision-based agents could lead to stronger generalization capabilities and, ultimately, the potential for integration with the physical world on a larger scale.

\section{Analysis}
\vspace{-10pt}
In this section, we aim to delve into the factors influencing the performance of VLMs in digital agent tasks and their underlying behavioral logic. 
We will investigate the impact of task attributes (such as difficulty, feasibility, visual requirement, and GUI complexity), input measurements (such as screenshot resolution, the influence of trajectory history, and the effect of UI layout), explore whether there are patterns in the agent's performance across different operating systems, and make a qualitative analysis in the aspect of models, methods, and humans.
All experiments, unless specifically mentioned otherwise, are conducted using GPT-4V under the Set-of-Mark setting.
Some takeaways from the analysis are:
1) higher screenshot resolution typically leads to improved performance; 
2) encoding more a11y (text) trajectory history can boost performance, while not working for screenshots (image); 
3) current VLMs are not adept at image-based trajectory history context;
4) current VLM agents are not robust to UI layout and noise;
5) the performance of VLM agents across OS is in strong correlation;
6) VLM agents have common error types like mouse-clicking inaccuracies, limited domain knowledge, and more types discussed in Sec.~\ref{qualitative_analysis}.

\subsection{Performance by Task Difficulty, Feasibility and App Involved}
We analyze the success rate across several additional subsets of tasks, as summarized in Tab.~\ref{tab:task_subset} and will be discussed in the following sections.

\begin{wraptable}{r}{6cm}
\vspace{-15pt}
\caption{Success rate (SR) of GPT-4V (SoM) across different types of tasks.}
\centering
\resizebox{1.0\linewidth}{!}{%
\begin{tabular}{@{}lcc@{}}
\toprule
    Task Subset             & \% of Total & SR (↑) \\ \midrule
    Easy            & 28.72\%      & \textbf{16.78\%} \\
    Medium       & 40.11\%      & 13.12\% \\
    Hard        & 30.17\%      & 4.59\% \\
    \midrule
    Infeasible    & 8.13\% & \textbf{16.67\%} \\
    Feasible      & 91.87\% & 13.34\% \\
    \midrule
    Single-App    & 72.63\% & \textbf{13.74\%} \\
    Multi-App Workflow    & 27.37\% & 6.57\% \\
\bottomrule
\end{tabular}
}
\vspace{-18pt}
\label{tab:task_subset}
\end{wraptable}

\paragraph{Task difficulty}
We categorize the tasks based on the time required for human completion into three groups: 0$\sim$60s (Easy), 60s$\sim$180s (Medium), and greater than 180 seconds (Hard), as an indicator of difficulty. 
Across these groups, the model's success rate drops as the required time increases, with tasks taking longer than 180 seconds becoming almost impossible to complete (considering we have infeasible examples for agent's luckiness), whereas human performance across these three groups is 84.91\%, 81.08\% and 49.57\%, showing a slight decline of the same trend but not to the extent of being unachievable.

\paragraph{Feasibility}
We also divide tasks into groups of tasks infeasible (\textit{e.g.}, deprecated features or hallucinated features) and tasks feasible, which requires the agents to have the ability to judge based on their own knowledge and exploration results.
As shown in Tab.~\ref{tab:task_subset}, we observe that agents currently perform slightly better in terms of infeasibility (16.67\% to 13.34\%), but overall, they are at a relatively low level. 
It is noteworthy that we also observe in some methods and settings (such as under the pure screenshot setting with the Gemini-Pro model), agents tend to easily output \texttt{FAIL} and refuse to continue trying. 
This situation leads to some false positives in infeasible tasks.
The focus needs to be on improving overall performance.

\paragraph{Number of apps involved}
We also examined the performance based on whether the task involved apps software or within a single app. 
As shown in Tab.~\ref{tab:task_subset}, the average performance for tasks involving a single app is low, at 13.74\%, but still more than double the 6.57\% observed for subsets of tasks involving workflows across multiple apps. 
Within single-app scenarios, tasks involving GUI-intensive Office apps generally performed the worst, with subsets such as LibreOffice Calc often scoring zero (we show more detailed results in App.~\ref{app:full_results}).
These findings highlight the need for improved collaboration capabilities between software and enhanced proficiency in specific scenarios.

\subsection{Performance by Multimodal Observation Variances}
\label{sec:performance_by_mm_obs_variances}
\paragraph{Higher screenshot resolution typically leads to improved performance}

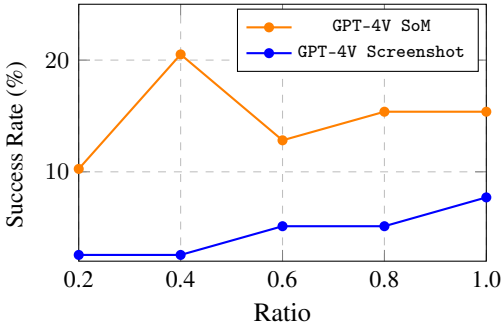
\begin{wrapfigure}{l}{0.5\textwidth}
\begin{tikzpicture}
\begin{axis}[
    width=7cm, height=5cm,
    xlabel={Ratio},
    xlabel style={yshift=0.2ex},
    ylabel={Success Rate (\%)},
    ylabel near ticks,
    ylabel style={xshift=-1ex},
    grid=both,
    xmin=1, xmax=5,
    ymin=2, ymax=25,
    xtick={1, 2, 3, 4, 5},
    xticklabels={0.2, 0.4, 0.6, 0.8, 1.0},
     ylabel style={font=\footnotesize},
    xticklabel style={font=\footnotesize},
    yticklabel style={font=\footnotesize},
    ymajorgrids=true,
    grid style=dashed,
    legend style={
        at={(1,1)},
        anchor=south east,
        xshift=-1.0mm,
        yshift=-9.0mm,
        font=\scriptsize,
    },
]

\addplot[
    color=orange,
    mark=*,
    mark size=1.5pt,thick
    ]
    coordinates {
    (1,10.26)(2,20.51)(3,12.82)(4,15.38)(5,15.38)
    };
    \addlegendentry{\texttt{GPT-4V SoM}}

\addplot[
    color=blue,
    mark=*,
    mark size=1.5pt,thick
    ]
    coordinates {
    (1,2.56)(2,2.56)(3,5.13)(4,5.13)(5,7.71)
    };
    \addlegendentry{\texttt{GPT-4V Screenshot}}
    
\end{axis}
\end{tikzpicture}
\vspace{-5pt}
\caption{The effect of downsampling on the screenshot on performance with down-sampling ratios of 0.2, 0.4, 0.6 and 0.8 and run on a subset (10\%) of examples.}
\vspace{-10pt}
\label{fig:resolution_effect}
\end{wrapfigure}

Despite the significant progress in display technology (1080P, 2K, and 4K), most VLMs are still trained on data far below these resolutions. 
We select the screenshot-only input and SoM setting to test the method's performance under different screen input down-sampling ratios (i.e., 0.2, 0.4, 0.6 and 0.8 of the original resolution), to evaluate the impact of resolution changes on model recognition ability and accuracy.
The output coordinates of the model for the screenshot setting are still expected to align with the original resolution (\textit{i.e.}, 1080P).
The effects of varying input resolutions on performance are shown in Figure~\ref{fig:resolution_effect}. 
For inputs based on pure screenshots, it is observed that an increase in resolution directly correlates with enhanced performance. 
This issue may arise from the discrepancy between the resolution of the screenshot and the coordinates of the output.
However, the scenario slightly differs on SoM.
Interestingly, a reduction in resolution to 768$\times$432 (down-sampling ratio of 0.4) leads to an improvement 
in the agent's performance and further diminishing the resolution even more to a down-sampling ratio of 0.2 results in a noticeable decline in performance. 

\paragraph{Longer text-based trajectory history context improves performance, unlike screenshot-only history, but poses efficiency challenges}

\begin{wrapfigure}{l}{0.5\textwidth}
\vspace{-12pt}
    \centering
    \begin{minipage}[b]{0.5\textwidth}
        \includegraphics[width=\textwidth]{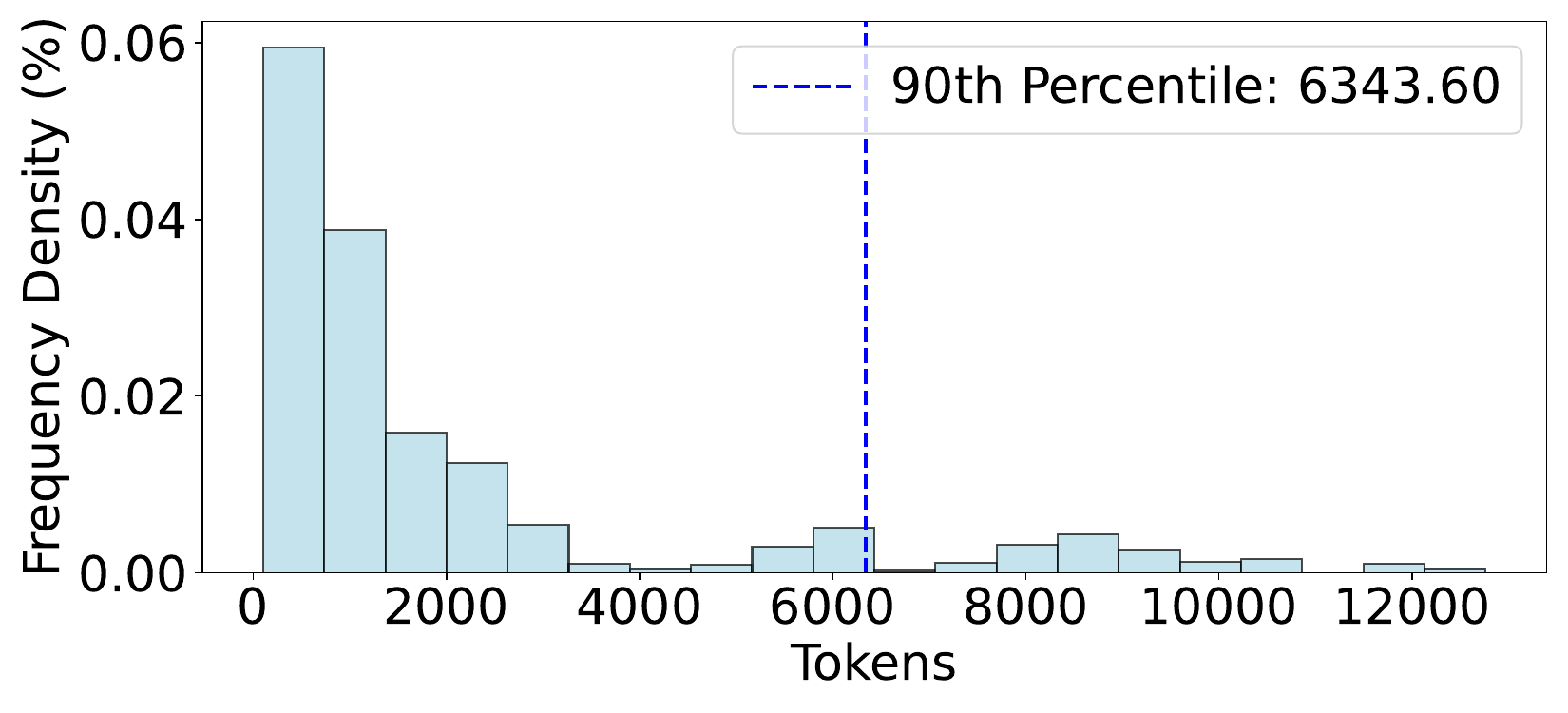}
        \vspace{-10pt}
        \caption{
        The length distribution of a11y tree as observation from sampled trajectories.
        }
        \label{fig:obs_length_distribution}
        \vspace{5pt}
    \end{minipage}
    \begin{minipage}[b]{0.5\textwidth}
    \begin{tikzpicture}
    \begin{axis}[
        width=7cm, height=5cm,
        xlabel={History Trajectory Length},
        xlabel style={yshift=1ex},
        ylabel={Success Rate (\%)},
        ylabel near ticks,
        ylabel style={xshift=-1ex},
        grid=both,
        xmin=1, xmax=4,
        ymin=2, ymax=16,
        xtick={1, 2, 3, 4},
        xticklabels={1, 2, 3, >3},
         ylabel style={font=\footnotesize},
        xticklabel style={font=\footnotesize},
        yticklabel style={font=\footnotesize},
        ymajorgrids=true,
        grid style=dashed,
        legend style={
            at={(1,1)},
            anchor=south east,
            xshift=-1.0mm,
            yshift=-17.0mm,
            font=\scriptsize,
        },
    ]
    
    \addplot[
        color=orange,
        mark=*,
        mark size=1.5pt,thick
        ]
        coordinates {
        (1,7.69)(2,12.82)(3,15.38)(4,15.40)
        };
        \addlegendentry{\texttt{GPT-4V SoM}}

    \addplot[
        color=blue,
        mark=*,
        mark size=1.5pt,thick
        ]
        coordinates {
        (1,7.69)(2,5.13)(3,7.71)(4,2.58)
        };
        \addlegendentry{\texttt{GPT-4V Screenshot}}
        
    \end{axis}
    \end{tikzpicture}
    \end{minipage}
    \vspace{-10pt}
    \caption{The effect of length of history on performance with the history encoding length of 1, 2, 3, and $>3$ and run on a subset (10\%) of examples.}
    \label{fig:history_effect}
    \vspace{-20pt}
\end{wrapfigure}

The main experiment revealed the decisive role of the a11y tree in performance within the current technological context. 
Even when we retain key attribute elements based on heuristic rules (keep nodes with tags of the document, item, button, heading, label, \textit{etc}.), LLMs still require a sufficiently large context to process this information effectively. 
To further understand this, we sample some a11y tree observations from \ours and conducted the statistical analysis, as shown in Figure~\ref{fig:obs_length_distribution}. 
The analysis indicates that a context length of 6000 is needed to accommodate about 90\% of cases for a single observation.
However, relying solely on current observations inherently leads to agents making repeated errors.
Therefore, we include current observations as well as past \texttt{N} rounds of observations and actions in the constructed prompts (see appendix for more details), to explore the impact on agent performance when \texttt{N} is set to 1, 2, 3, and all where we put as much context as we can.
The experimental results (as shown in Figure~\ref{fig:history_effect}) show the performance increase with more history context for SoM.
Future work on constructing models with enhanced capabilities for longer context support and understanding reasoning, improving model efficiency, and designing new agent architectures for efficient memory storage will have a significant impact on digital agents.

However, we also note that the inclusion of additional trajectory history does not enhance performance under the pure screenshot setting.
This suggests that contemporary advanced VLMs might not be as adept at extracting robust contextual information from images as they are from textual data. 
Strengthening this capability to harness information from images constitutes an important avenue for future enhancements.

\paragraph{VLM agents struggle with perturbation of position and size of application windows and irrelevant information}
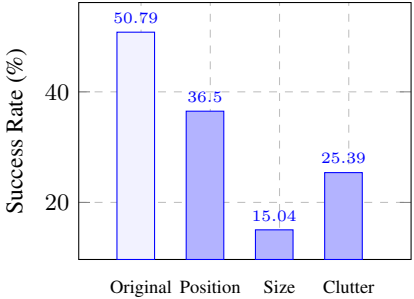
\begin{wrapfigure}{l}{0.4\textwidth}

\begin{tikzpicture}
    \begin{axis}[
            ybar=3.8pt,
            enlargelimits=0.15,
            enlarge x limits=0.3,
            width=6cm, height=5cm,
            bar width=0.5cm,
            grid=both,
            grid style=dashed,
            every axis label={font=\footnotesize},
            tick label style={font=\footnotesize},
            legend style={at={(0.5,1.25), font=\tiny},
            anchor=north,legend columns=-1},
            xlabel style={font=\footnotesize},
            ylabel={Success Rate (\%)},
            ylabel style={font=\footnotesize,xshift=2.5em},
            ylabel near ticks,
            symbolic x coords={Original, Position, Size, Clutter},
            xtick=data,
            xlabel near ticks,
            xtick style={draw=none},
            ylabel style={font=\footnotesize},
            xticklabel style={font=\scriptsize},
            yticklabel style={font=\footnotesize},
            nodes near coords,
            nodes near coords align={vertical},
            nodes near coords style={font=\tiny}
        ]
        \addplot+[bar shift=-2pt, forget plot] coordinates {(Position, 36.50) (Size, 15.04) (Clutter, 25.39)};
        \addplot+[bar shift=-2pt, forget plot, fill=blue!5] coordinates {(Original,50.79)};
    \end{axis}
\end{tikzpicture}
    \vspace{-10pt}
    \caption{Decline in performance due to window perturbations.
    }
    \label{fig:ui_robustness}
    \vspace{-5pt}
\end{wrapfigure}

We continue to adopt the SoM setting and sample a subset of 28 tasks that agents relatively well perform (with a success rate of 50.79\%) in \ours.
At the beginning of each task, we introduce disturbances to the windows by 1) changing the 
\textit{position} of the window; 2) changing the \textit{size} of the window to the minimal; 
3) opening some irrelevant software and maximizing them to \textit{clutter} the screen.
This process generates several times more samples from the subset of tasks to observe their performance. 
We find current agents are not robust in handling all these changes, which leads to a performance drop to over 60\% to even 80\%.
Surprisingly, we find agents can switch the window to a certain degree but fail to maximize the window as an intermediate step and are stuck on other things. 
This suggests that while agents possess some capability to navigate between windows, they lack a comprehensive strategy for managing window states effectively.

\subsection{Performance across Different Operating Systems}
Another key challenge in building universal digital agents is ensuring that these agents can maintain efficient and consistent performance across different operating system environments. 
The differences between OS and their software ecosystems can significantly impact an agent's observation and action spaces, leading to performance uncertainties. 
Here, we explore and analyze the correlation between the success of agents in completing tasks on Windows after migrating from Ubuntu using examples from \ours.

\begin{wraptable}{r}{6cm}
\vspace{-0.1in}
\caption{Comparison of model performance and correlation across operating systems.}  
\centering
\resizebox{1.0\linewidth}{!}{%
\begin{tabular}{@{}lcc@{}}
\toprule
OS & SR (\%) & \small{Correlation Coefficient} \\ \midrule
Ubuntu           & 4.88               & \multirow{2}{*}{0.7}  \\
Windows          & 2.55               &                        \\ \bottomrule
\end{tabular}
}
\vspace{-0.1in}
\label{tab:os_performance}
\end{wraptable}

We enhance the functionality of the \ours environment to support setting up initial experiment states, final evaluations, and obtaining observations such as the a11y tree and screenshots in Windows OS.
Additionally, we have made example-wise fine-tuning modifications to the existing subset in \ours for migration to Windows. 
We conduct evaluations using the GPT-4V screenshot-only method and present the correlation of performance across the two operating systems. 
As shown in Tab.~\ref{tab:os_performance}, the model's performance on Ubuntu and Windows is 4.88\% and 2.55\%, respectively, with a correlation coefficient of 0.7, despite the differences in their observation spaces.
This implies that insights and methodologies developed within the \ours framework can be effectively transferred to Windows environments with a high degree of reliability.

\subsection{Qualitative Analysis}
\label{qualitative_analysis}
In this section we highlight representative examples of success, failure, and surprising outcomes, alongside a comparative study between GPT-4V and Claude-3 agents, to elucidate the unique challenges and insights our environment introduces. 
See App.~\ref{appendix:qualitative_analysis} for more details.

\begin{figure}[htbp]
    \vspace{-5pt}
    \centering
    \includegraphics[width=\linewidth]{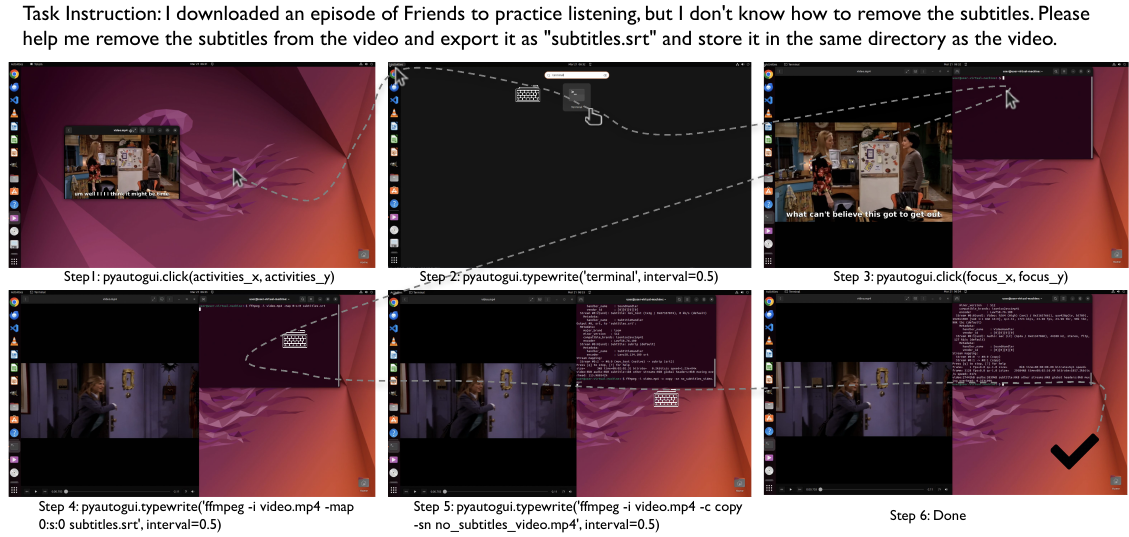}
    \caption{
    The agent successfully understood the complex task instructions, extracted the subtitle file from the video, and generated a pure video without embedded subtitles.
    }
    \vspace{-5pt}
    \label{fig:success_example}
\end{figure}

\paragraph{Success and failure cases}
We find agents, particularly based on GPT-4V, can successfully solve tasks that involve complex problem-solving or creative thinking, showcasing the advanced understanding and processing capabilities of the model already.
One successful task is shown in the first row of Figure~\ref{fig:success_example}. 
The agent is requested to extract subtitle files from the video stream and save them locally. 
The agent first divides the screen into two parts, with the VLC application window on the left and the terminal window open on the right, and uses the ffmpeg command twice. 
The first use removes the subtitles embedded in the original video, and the second use saves the extracted subtitles locally.

\begin{figure}[htbp]
    \centering
    \vspace{-5pt}
    \includegraphics[width=\linewidth]{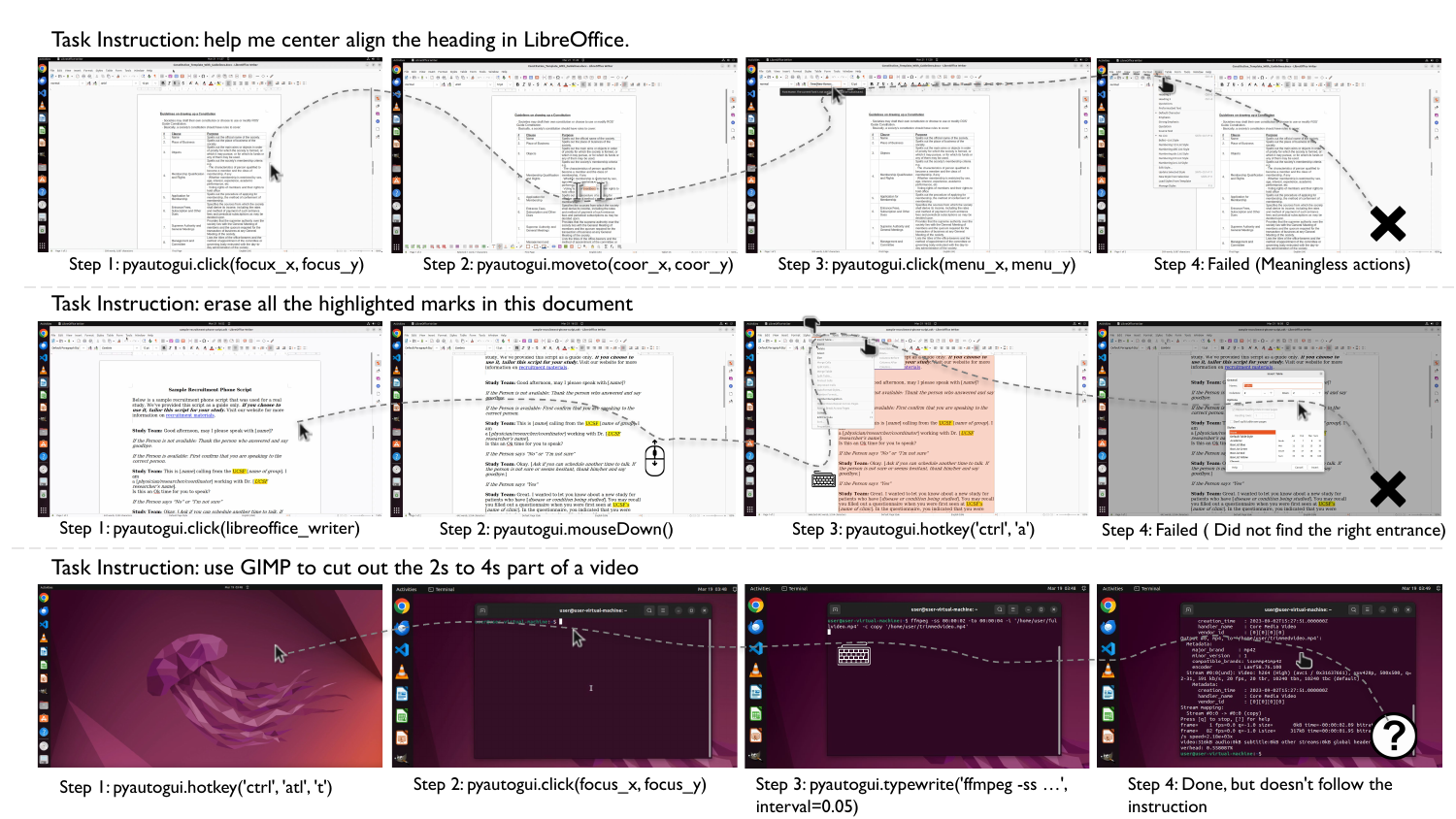}
    \caption{
    Screenshots of the three examples mentioned in the quality analysis. 
    The first line is an example of GPT-4V failing at a very simple task, the second line is one example where agents face more difficulty than humans, and the third line is one example that is more difficult for humans than for agents.
    }
    \vspace{-10pt}
    \label{fig:failure_human_agent}
\end{figure}

Despite the successes, there are notable failures that highlight the limitations of current models. In the task of ``center-aligning the title of the document'' (Fig.~\ref{fig:failure_human_agent} line 1), the agent fails to ground the relatively simple requirement of ``center alignment of texts'', performing many useless actions such as selecting irrelevant words, opening irrelevant menus, \textit{etc}.

Moreover, we find that the agent lacks prior knowledge in using software, performing poorly in many specialized tasks (as shown in Fig.~\ref{fig:model_lack_knowledge}, with GIMP, LibreOffice Calc, and Chrome selected). 
Taking GIMP as an example, for the instruction ``reduce brightness'' the agent does not know which menu in the toolbar is for brightness adjustment and instead randomly tries until exhausting the maximum number of steps.

\paragraph{Common errors by GPT-4V agents}
Among the 550 failed examples from different settings in our sample, more than 75\% exist \textit{mouse click inaccuracies}, which is the most common error. 
The agent fails to click the correct coordinates despite planning detailed and accurate steps in their code comments, indicating strong planning but weak execution capabilities. 
Mouse click inaccuracies lead to two other frequent errors: 
\textit{repetitive clicks} and \textit{environmental noise dilemma}. 
Repetitive clicks occur when the agent repeatedly misclicks, adjusts, and fails, consuming too many steps. 
Environmental noise arises from clicking unintended objects, causing pop-ups, or opening unrelated applications. 
Due to a lack of prior knowledge about most professional software, it falls into a mismatch dilemma between the actions taken and the current state, and don't know how to get back to normal.
Moreover, the agent lacks basic human-like cognition of web pages, such as not closing pop-ups in real-world web pages or being attracted by advertisement content, which affects its original correct judgment. 
Failures also arise from \textit{misinterpretation of instructions} and \textit{visual oversight}, highlighting the need for improvement in language and visual processing. 
See App.~\ref{appendix:common_errors} for the specific execution process.

\paragraph{Discrepancies in task difficulty between agent and human}
We identify notable disparities in the perceived difficulty of tasks between humans and AI agents. Tasks that are intuitively simple for humans often present substantial challenges to agents, and conversely, tasks that humans find demanding can be more straightforward for agents to execute.
You can find more details in Fig.~\ref{fig:human_agent_supplement} and App.~\ref{appendix:human_vs_agent}.

\paragraph{Tasks where humans outperform agents
} These tasks mainly involve text-based and design-related work, such as ``bold the font on this slide and add notes'' or ``erase all the highlighted marks in this document'' (Fig.~\ref{fig:failure_human_agent} Line 2). 
Since the Internet lacks such fine-grained data as the software execution process, the agent also lacks the corresponding training process, so its grounding ability is not good enough.
The lack of understanding of GUI logic also causes poor performance on operations like selecting and scrolling.

\paragraph{Tasks where agents outperform humans} Tasks that the agent considers simple but humans find difficult are concentrated in ``code solvability tasks'', such as ``monitor the system CPU for 30s and output the results'' and ``force close a process''. 
These tasks require little or no GUI interaction and can be completed by executing complex codes and instructions. 
It's worth noting that completing through code sometimes mismatches with human instructions. 
In the task "use GIMP to cut out the 2s to 4s part of a video,(Fig.~\ref{fig:failure_human_agent} Line 3)" the agent used ``ffmpeg'' command to complete the video cropping, ignoring the ``use GIMP'' requirement in the instructions.

Surprisingly, we discovered that agents are as prone to inefficiency in mechanically repetitive tasks, such as copying, pasting, and batch editing of Excel sheets, as humans. 
Humans frequently commit careless errors during execution. 
The shortcomings in agents stem either from the absence of an API or from insufficient training data related to the API, hindering their ability to efficiently process tasks in batches. 
Furthermore, sluggish response times can cause tasks to either time out or surpass the maximum allowed steps.

\paragraph{Comparative analysis: Claude-3 vs. GPT-4V}
Although Claude outperforms GPT-4 in many benchmarks such as GSM8K, HumanEval, \textit{etc}., in our main experiment, we find that Claude has an average lower accuracy rate compared to GPT-4V by 2.84\% to 7.76\%. 
We find that Claude can provide satisfactory high-level solutions, but its grounding ability contains hallucinations in detail.
For instance, Claude would interpret double-clicking a file as selecting it instead of opening it, treat column B in LibreOffice Calc software as column C, and enter text in the VS Code text replacement box without clicking on global replace. 
This shows that Claude can align well with human planning in problem-solving, but lacks excellent grounding ability when it comes to execution. Details can be seen in Fig.~\ref{fig:claude_error} and App.~\ref{appendix:claude3}.

\section{Related Work}
\label{sec:related}

\paragraph{Benchmarks for multimodal agents}
Testing digital interaction agents mainly spans coding environments, web scenarios, and mobile applications. In the coding domain, several works provide frameworks and datasets for evaluating agents across programming languages and software engineering activities~\citep{yang2023intercode, jimenez2023swe, li2024devbench, si2024design2code}. 
For web browsing, platforms have been developed for agents to interact with web interfaces through keyboard and mouse actions, alongside datasets focusing on open-ended web tasks and realistic web navigation~\citep{shi2017world, liu2018reinforcement, yao2022webshop, deng2023mind2web, zhou2023webarena, Koh2024VisualWebArenaEM, drouin2024workarena}.
Mobile device interaction research aims at improving accessibility, with simulators for mobile UI interactions and platforms dedicated to InfoUI tasks~\citep{li2020mapping, sun2022meta, venkatesh2022ugif, toyama2021androidenv, rawles2023android, zhang2023mobile, wen2023empowering, zhang2023appagent, wang2024mobile}. 
Further, environments connecting to real computers and datasets for GUI grounding, albeit without interactive capability, have emerged~\citep{gao2023assistgui, Cheng2024SeeClickHG, niu2024screenagent, Kapoor2024OmniACTAD, tan2024towards}. 
Comprehensive task evaluation across different aspects also sees innovations~\citep{liu2023agentbench, mialon2023gaia}.
Differing from previous endeavors focusing on singular environments or lacking executability, \ours integrates an interactive setup enabling agents to engage with operating systems openly, supported by a diverse array of tasks and precise evaluation scripts within a fully controllable setting, marking it as a competitive benchmarking realism and reliability, as well as an environment for learning and evaluating general-purpose digital agent (See Tab.~\ref{tab:benchmark_comparsion} for comparison).

\paragraph{Vision-language models for multimodal agents}
Many existing works on GUI interaction utilize some form of structured data (such as HTML, accessibility trees, view hierarchies) as a grounding source~\citep{deng2023mind2web, gur2023real, li2020mapping, nakano2021webgpt, zhao2022tie, sridhar2023hierarchical, zhang2024large, zhou2023webarena}. 
However, source code often tends to be verbose, non-intuitive, and filled with noise. 
In many cases, it is even inaccessible or unavailable for use, making multi-modality or even vision-only perception a must.
To take screenshots as input, there are already specialized, optimized multi-modal models available that are suited for tasks on web~\citep{Baechler2024ScreenAIAV, furuta2023multimodal, humphreys2022data, lee2023pix2struct, shaw2023pixels} and mobile devices~\citep{hong2023cogagent, zhang2023you}. 
Additionally, general-purpose foundation models~\citep{bai2023qwen, li2023silkie, liu2023visual, zhu2023minigpt} also demonstrate significant potential for multi-modal digital agents.
The development of prompt-based methods~\citep{gao2023assistgui, he2024webvoyager, yan2023gpt, zheng2024gpt}, as well as visual reasoning paradigms, have also further facilitated the performance of digital agents in web pages, mobile apps, and desktop. 
To investigate how well do current models and methods perform in digital agent tasks, our paper evaluates the results of text-only, vision-only, and multi-modal input as well as across multiple methods, demonstrating that existing multi-modal models are far from capable computer agents.
Specifically, there is ample room for improvement in long-horizon planning, screenshot details perception, pixel coordinate locating, and world knowledge.

\section{Conclusion and Future Work}\label{sec:conclusion}

In conclusion, the introduction of \ours marks a significant step forward in the development of autonomous digital agents, addressing critical gaps in existing interactive learning environments. 
By providing a rich, realistic setting that spans multiple operating systems, interfaces, and applications, \ours not only broadens the scope of tasks digital agents can perform but also enhances their potential for real-world application. 
Despite the promise shown by advancements in vision-language models, evaluations within \ours reveal notable challenges in agents' abilities, particularly in GUI understanding and operational knowledge, pointing to essential areas for future research and development.

We identify several potential directions for community development and progress toward general-purpose agents for computer operation:
\paragraph{Enhancing VLM capabilities for efficient and robust GUI interactions}
For foundation model development, we need to boost the efficiency of our models, enabling them to process much longer contexts and perform inference computations efficiently, akin to the robotics community~\citep{brohan2022rt, brohan2023rt} to better handle real-world cases.
Enhancements in VLMs' GUI grounding capabilities that is robust to application windows changes and  are also sought, focusing on the accurate understanding and generation of precise actions aligned with given instructions. 
Moreover, amplifying VLMs' ability to comprehend context in the form of images is a pivotal goal, since it is crucial to enable history encoding using images so that we can build memory and reflection upon that.
These improvements may require more efforts in the upstream pre-training stage, downstream fine-tuning stage, and even in the model structure itself, as pointed out in previous work~\citep{deng2023mind2web, hong2023cogagent, lu2024weblinx}.

\paragraph{Advancing agent methodologies for exploration, memory, and reflection}
The next-level approach encompasses designing more effective agent architectures that
augment the agents' abilities to explore autonomously and synthesize their findings. 
The agents face challenges in leveraging lengthy raw observation and action records. It's fascinating to explore novel methods for encoding this history, incorporating efficient memory and reflection solutions to condense contextual information and aid the agent in extracting key information. Additionally, integrating knowledge grounding into (V)LLM agents through memory mechanisms is a promising avenue as well.
Moreover, practice GUI assistants also require features of personalization and customization. These features rely on techniques such as user profiling and retaining memories from long-term user-assistant interactions.
Additionally, crafting protocols specifically for digital agents operating within GUI and CLI interfaces aims at facilitating efficient actions is also an essential thing for the feasibility of general-purpose digital agents in the mid-short term.

\paragraph{Addressing the safety challenges of agents in realistic environments}
The safety of agents is a critical issue if applying a built agent in fully realistic environments, the developed universal digital agent could potentially be used to bypass CAPTCHA systems in the future, as noted in \citep{searles2023empirical}. 
However, due to the currently limited capabilities of agents, we have not observed any harmful and damaging behaviors during our experiments, an automatic agent has the opportunity to damage patent rights, abuse accounts, attempt to exploit software vulnerabilities to create viruses, or engage in attacks. 
Currently, we adopt virtual machines to make it difficult for developing digital agents to cause irreversible damage to our host machines. 
However, there still lacks a reliable metric to assess the safety of an agent developed in an isolated environment. 
The current evaluation functions mainly focus on the results closely regarding the task instructions, assess only the correctness of task completion, and pay little attention to potential unnecessary damaging actions of agents.
Owing to the complexity of a complete computer environment, we didn't work out an efficient way to detect the latent side effects of the agent.
Consequently, how to assess and control potential behaviors in open and real environments through environmental constraints and agent training is an important further direction of research.

\paragraph{Expanding and refining data and environments for agent development}
In terms of datasets and environments, we can broaden the scope to cover more specialized domains, including real-sector needs in healthcare, education, industry, transportation, and personalized requirements. 
Efforts can be made to ensure our environment's seamless deployment across various hardware and software settings. 
The variance of a11y tree quality across different applications is also noticed. Although
the problem is not remarkable in the applications currently included, there is no guarantee
of that the application developers obey the a11y convention and offer clear and meaningful 
descriptions for GUI elements. More intelligent approaches to filter redundant a11y tree elements
and to handle latently missing elements deserve careful investigation as well.
We also highlight the necessity of a painless data collection method, allowing for the effortless acquisition of computer operation data and its transformation into agent capabilities.

\section*{Acknowledgements}
We thank Sida Wang, Peter Shaw, Alane Suhr, Luke Zettlemoyer, Chen Henry Wu, Pengcheng Yin, Shunyu Yao, Xing Han Lu, Siva Reddy, Ruoxi Sun, Zhiyuan Zeng, Chengyou Jia, and Lei Li for their helpful feedback on this work.

\bibliographystyle{plainnat}
\bibliography{bibliography}

\newpage
\appendix
\tcbset{colframe = blue!50!black, 
colback = white, 
toptitle=1mm, 
bottomtitle=1mm, 
fonttitle=\fontsize{10pt}{12pt}\selectfont,  breakable, enhanced jigsaw, finish broken={\tcbset{colframe=white, bottom=0mm, bottomrule=0mm}},before upper={\tcbset{colframe=white, toprule=0mm}},after={\tcbset{colframe=blue!40!black, toprule=1mm, bottomrule=1mm}},}

\section{Details of \ours Environment}

\subsection{Environment Infrastructure}
\label{app:env_infra}
As compared to core commonly used techniques like Docker~\footnote{\url{https://www.docker.com/}}, virtual machines can operate their own kernel and system, enabling compatibility with a wide variety of operating systems (such as Windows, macOS, Linux, etc.) across different CPU hardware types (x64, ARM, etc.), and supports training and evaluation in a multiprocess manner on both headless servers and personal computers.

\subsection{Observation Space}
\label{app:observation_space}
We implement three kinds of observation: 
\textbf{complete screenshot}, \textbf{accessibility tree} and \textbf{terminal output}.
We also implement a video recorder of the environment but don't put it into our modeling due to the agent's ability limitations.
\ours supports observation refactoring and extending if needed, such as getting data from certain opening applications that we want to focus on.

\subsubsection{Screenshot}
To align with the perception of a human user, we capture a screenshot of the entire computer screen. 
Including the mouse cursor also proves helpful in certain cases where mouse information is crucial. 
For screen resolution, we default to 1920$\times$1080, as it is the most commonly used screen resolution according to Internet Users Screen Resolution Realtime Statistics for 2023\footnote{\url{https://www.screenresolution.org/year-2023/}}. 
This resolution also offers a 16:9 aspect ratio. 
\ours also supports modifying the resolution of virtual machines to avoid potential memorization of absolute pixel values and to assist studies on topics like generalization across different resolutions.


\subsubsection{Accessibility Tree}
An accessibility tree (or a11y tree, same logic to kubernetes and k8s), refers to an intricate structure generated by the browser or OS accessibility APIs that renders a representative model of the web content, providing a means of interaction for assistive technologies. 
Each node within the accessibility tree hosts important information about a UI element. 
This could range from the nature of the object (a button, checkbox, or paragraph of text), its current state (checked or unchecked, for checkboxes), and even its spatial orientation on the screen.

\begin{figure}[ht]
    \centering
    \includegraphics[width=0.8\textwidth]{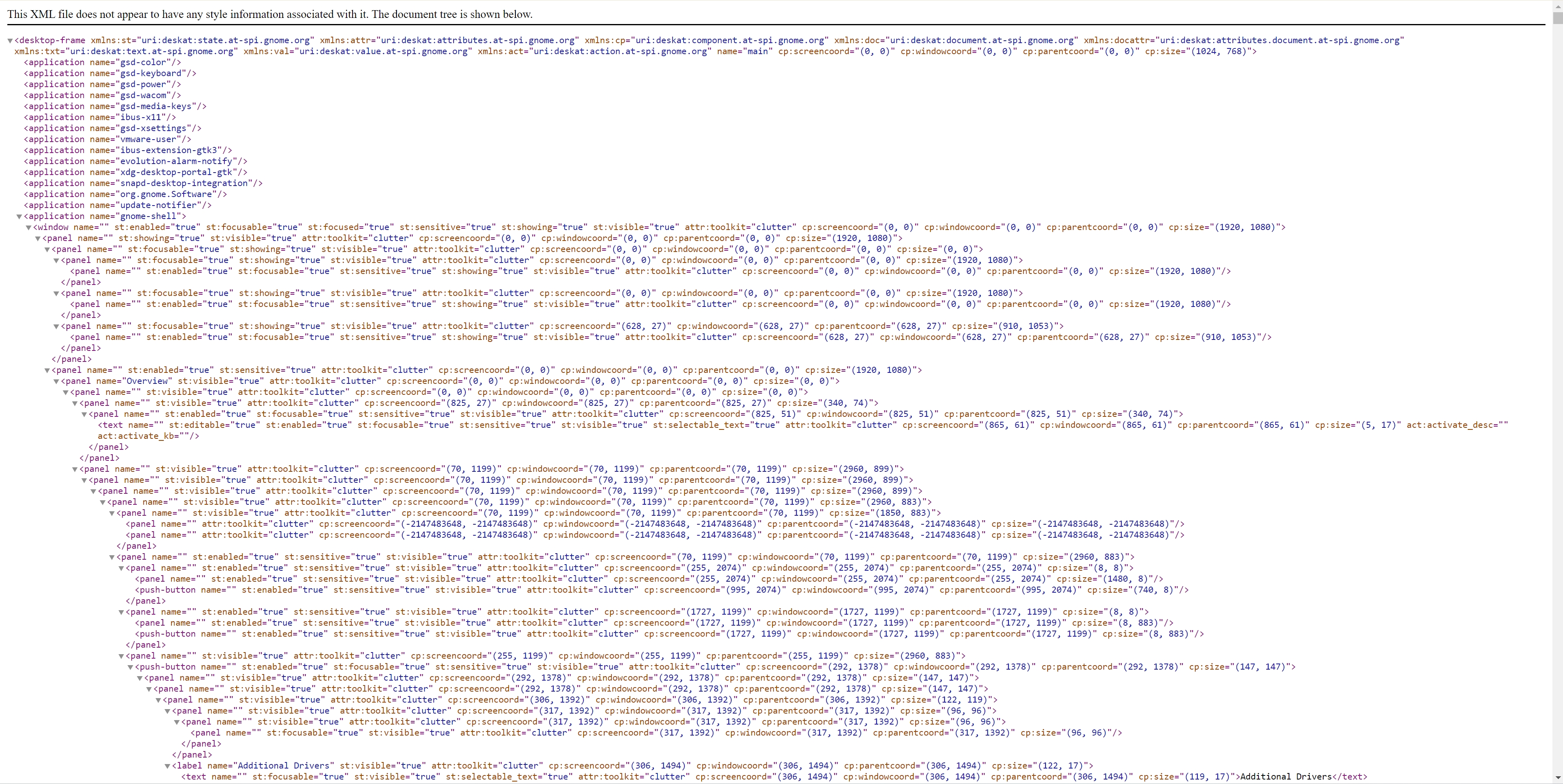}
    \caption{Raw XML formatted accessibility tree visualization.}
    \label{fig:accessibility_tree_xml}
\end{figure}

Different operating systems employ varied accessibility APIs and tools to construct and manipulate the accessibility tree. 
These include Microsoft Active Accessibility (MSAA) and User Interface Automation (UIA) for Windows, NSAccessibility Protocol and macOS Accessibility Inspector for macOS, and 
Assistive Technology Service Provider Interface (ATSPI)~\footnote{\url{https://docs.gtk.org/atspi2/}} for GNOME dekstop used on Ubuntu.
We adopt \texttt{pyatspi} to get the accessibility tree on Ubuntu and \verb|pywinauto| on Windows.
We further convert it into XML format for message passing.
Partial pieces of the XML formatted accessibility tree are shown in Figure~\ref{fig:accessibility_tree_xml}.

\begin{figure}[htbp]
    \centering
    \includegraphics[width=0.8\textwidth]{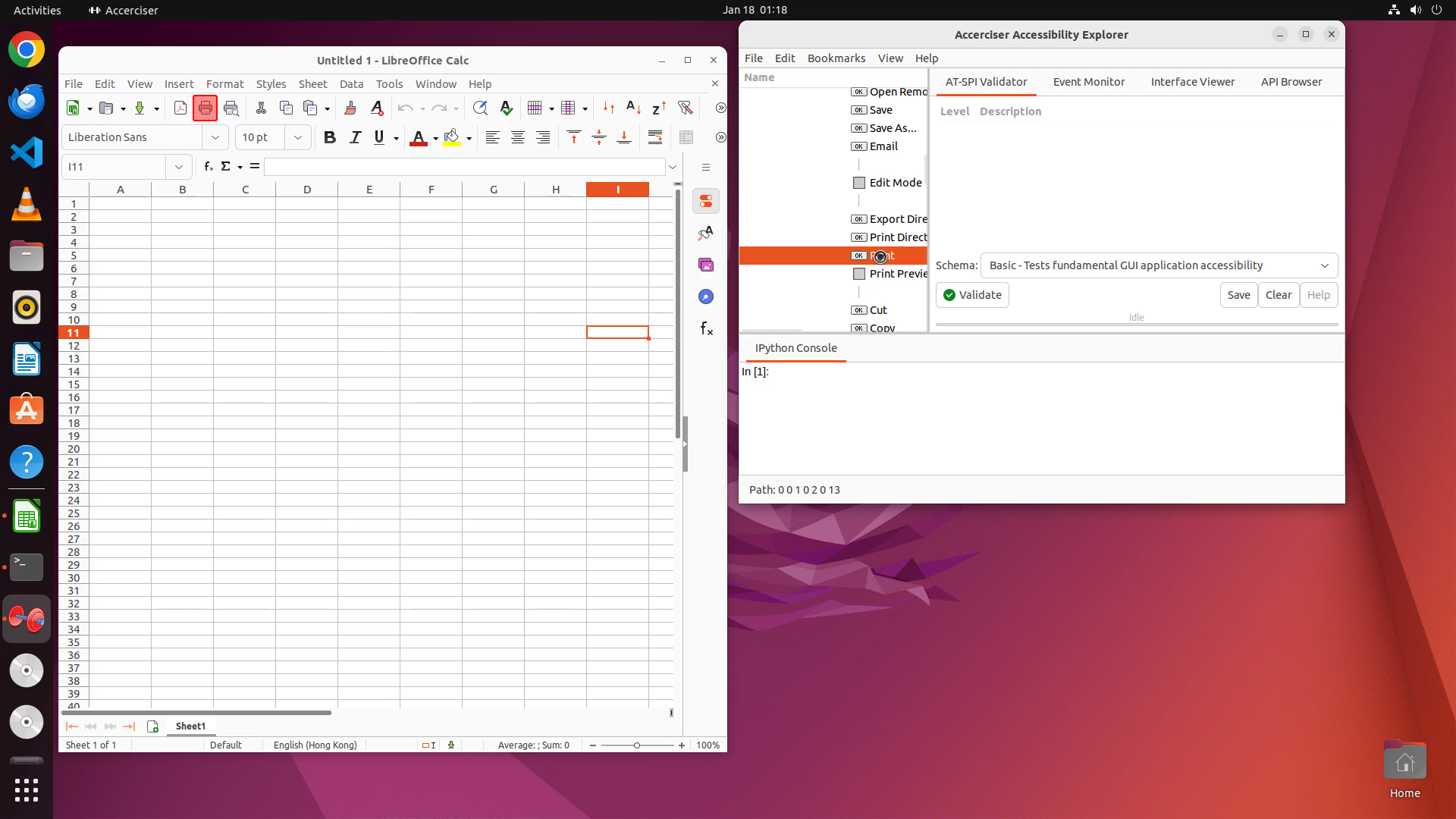}
    \caption{Use \texttt{Accerciser} to visualize components in accessibility tree. This example shows the corresponding relationship between the ``Print'' button (which is a printer icon on the bar of the LibreOffice app as shown on the left) and the ``Print'' tree node (which is the ``Print'' of the Accerciser Accessibility Explorer as shown on the right).}
    \label{fig:accerciser_example}
\end{figure}

Tools such as \texttt{Accerciser} can help visualize the corresponding relationship of tree nodes and GUI components in the accessibility tree as shown in Figure~\ref{fig:accerciser_example}.

\subsection{Action Space}
\label{app:action_space}
We implement two kinds of action space: 
\texttt{pyautogui} and \texttt{computer\_13}.
We mainly use \texttt{pyautogui} action space, since it saves tokens for describing action space definition in prompting, compared with self-designed actions.

\subsubsection{\texttt{PYAUTOGUI}}
\texttt{pyautogui} is an open-source, cross-platform Python module utilized for programmatically controlling the mouse and keyboard. 
It can control simple movements, clicks, as well as keyboard inputs, and can provide screen capture features or locate where the screen-specific images reside that can be useful for GUI automation. 
Compared with other Python packages such as xdotool~\footnote{\url{https://github.com/jordansissel/xdotool}}, mouse~\footnote{\url{https://github.com/boppreh/mouse}} and keyboard~\footnote{\url{https://github.com/boppreh/keyboard}}, \texttt{pyautogui} has better compatibility across different OSes as found in our early attempts, so that we adapt that as our core component of computer controlling as well as an official valid action space.

Herein, we will demonstrate some use cases of \texttt{pyautogui} and illustrate how it can be wielded as an action space.

\paragraph{Mouse Controlling Functions}
\small
\begin{Verbatim}
import pyautogui

# move mouse to XY coordinates over num_second seconds
pyautogui.moveTo(x, y, duration=num_seconds)

# drag mouse to XY
pyautogui.dragTo(x, y, duration=num_seconds)

pyautogui.click(x=moveToX, y=moveToY, clicks=num_of_clicks, button='left')
pyautogui.rightClick(x=moveToX, y=moveToY)
pyautogui.middleClick(x=moveToX, y=moveToY)
pyautogui.doubleClick(x=moveToX, y=moveToY)
pyautogui.tripleClick(x=moveToX, y=moveToY)
pyautogui.scroll(amount_to_scroll, x=moveToX, y=moveToY)
pyautogui.mouseDown(x=moveToX, y=moveToY, button='left')
pyautogui.mouseUp(x=moveToX, y=moveToY, button='left')
\end{Verbatim}

\paragraph{Keyboard Controlling Functions}

\small
\begin{Verbatim}
import pyautogui

# useful for entering text, newline is Enter
pyautogui.typewrite('Hello world!\n', interval=secs_between_keys)

pyautogui.typewrite(['a', 'b', 'c', 'left', 'backspace', 'enter', 'f1'], interval=secs_between_keys)
pyautogui.hotkey('ctrl', 'c')  # ctrl-c to copy
pyautogui.hotkey('ctrl', 'v')  # ctrl-v to paste
pyautogui.keyDown(key_name)
pyautogui.keyUp(key_name)
\end{Verbatim}

\paragraph{\texttt{pyautogui} as an Action Space}

Given the various controls it provides, \texttt{pyautogui} can readily be used as an action space in building automation software or testing interfaces with minor adjustments.
More formally, an action is within the action space when it meets the syntax of \texttt{pyautogui} or is one of three special actions \texttt{WAIT}, \texttt{FAIL}, and \texttt{DONE}.
This might include actions like clicking at a certain location, entering text or key sequences, or even resting for a span (Pause). Each action could be mapped to an identifying label or number, forming a discrete action space. 
For example:
\small
\begin{Verbatim}
import pyautogui

def perform_action(action):
  if action == 0:
    pyautogui.moveTo(100, 100)
  elif action == 1:
    pyautogui.write('Hello world!', interval=0.25)
  else:
    pyautogui.pause(1)
\end{Verbatim}
In this scheme, the "perform\_action" function constitutes the action space, where each unique action is associated with a unique integer (its action ID). 
The function interprets these action IDs and performs the corresponding action, forming a rudimentary discrete action space.

One interesting finding is that language models generate screenshot locate functions like:
\begin{Verbatim}
pyautogui.locateOnScreen('Apple.png')
\end{Verbatim}
When there is insufficient grounding evidence (such as when no screenshot is inputted, the accessibility tree lacks a specific element, or the multimodal model cannot comprehend the user interface), employing this function to retrieve the correct icon image could present an interesting method.

\subsubsection{\texttt{COMPUTER\_13}}
To facilitate potential reinforcement learning applications, we have created a variant of \texttt{pyautogui}, which we named \texttt{computer\_13}. 
In this variant, we wrap \texttt{pyautogui} into a finite action class with parameterized enumeration, such that it features 13 action types, excluding three special ones for task process control. 
Utilizing this structured approach allows more effective reinforcement learning by providing a distinct and finite set of actions to be learned and optimized.
As summarized in Table~\ref{table:action_space}, each action type has certain parameters, detailed in the collection, confirming the type, range, and whether each parameter is optional for that action.

\begin{table}[ht]
\centering
\caption{Action types and parameters defined in action space \texttt{computer\_13}, a variance we created for the potential reinforcement learning research based on our environment.}
\label{table:action_space}
\begin{tabular}{p{2.4cm}|p{2cm}|p{8.5cm}}
\textbf{Action Type} & \textbf{Parameters} & \textbf{Note} \\
\hline
\texttt{MOVE\_TO} & \textit{x, y} & \textit{Move the cursor to the specified position} \\
\texttt{CLICK} & \textit{button, \newline x, y, \newline num\_clicks} & \textit{Click the left button if the button not specified, otherwise click the specified button; click at the current position if x and y are not specified, otherwise click at the specified position} \\
\texttt{MOUSE\_DOWN} & \textit{button} & \textit{Press the left button if the button not specified, otherwise press the specified button} \\
\texttt{MOUSE\_UP} & \textit{button} & \textit{Release the left button if the button not specified, otherwise release the specified button} \\
\texttt{RIGHT\_CLICK} & \textit{x, y} & \textit{Right click at the current position if x and y are not specified, otherwise right click at the specified position} \\
\texttt{DOUBLE\_CLICK} & \textit{x, y} & \textit{Double click at the current position if x and y are not specified, otherwise double click at the specified position} \\
\texttt{DRAG\_TO} & \textit{x, y} & \textit{Drag the cursor to the specified position with the left button pressed} \\
\texttt{SCROLL} & \textit{dx, dy} & \textit{Scroll the mouse wheel up or down} \\
\texttt{TYPING} & \textit{text} & \textit{Type the specified text} \\
\texttt{PRESS} & \textit{key} & \textit{Press the specified key and release it} \\
\texttt{KEY\_DOWN} & \textit{key} & \textit{Press the specified key} \\
\texttt{KEY\_UP} & \textit{key} & \textit{Release the specified key} \\
\texttt{HOTKEY} & \textit{keys} & \textit{Press the specified key combination} \\
\hline
\texttt{WAIT} & - & \textit{Wait until the next action} \\
\texttt{FAIL} & - & \textit{Decide the task cannot be performed} \\
\texttt{DONE} & - & \textit{Decide the task is done} \\
\end{tabular}
\end{table}

\section{Details of \ours Benchmark}
\subsection{Operating System Selection}
\label{app:os_selection}
As of 2023, the most popular desktop operating systems are Windows (69.5\%), macOS (20.4\%), ChromeOS (3.2\%), and Linux (3.1\%) \footnote{\url{https://www.statista.com/statistics/218089/global-market-share-of-windows-7/}}. 
While Windows and macOS dominate the market share and boast the richest software ecosystems, their closed-source nature raises potential copyright concerns for direct usage. 
ChromeOS, being a web-based operating system, heavily depends on a Google account for its functionalities, rendering it less suitable for a public benchmark. 

In contrast, Linux desktops offer a wealth of open-source software for most daily tasks, supported by an active community for both basic and advanced use. 
Essential applications such as Office Suite, browsers, email clients, multimedia apps, and thousands of games and applications are either pre-installed or readily available through the software center of the distribution. 
Consequently, we select Ubuntu, the most representative Linux desktop OS, as the foundation for the main part of our benchmark intended for public use.
Additionally, we have developed components to facilitate agent interaction on a Windows virtual machine and have created a relatively small set of examples focusing on the Microsoft Office suite, including Excel, PowerPoint, and Word. 
This serves as a counterpart to the LibreOffice suite available on Ubuntu. 
These components can be utilized in-house or officially with the purchase of a license.
Regarding macOS, theoretically, it is illegal to install macOS on non-Apple devices, which leads us to refrain from developing our benchmark on this platform to avoid copyright issues.

\subsection{Software Selection}
\label{app:software_selection}
Due to the high cost of obtaining operation and evaluation script annotation data, we have chosen a representative set of software for the examples of Ubuntu part.
We adopt standards that consider: 
1) Availability - the software must be available on Ubuntu 22.04; 
2) Open-source - the software should be open-sourced with an appropriate license to prevent copyright issues; 
3) Popularity - the software should take a high download number and frequency of recommendations in blogs and tutorials; 
4) Strong user community and good support resources - it is preferable to have an active and robust user community as well as official documents, which can serve as ample resources for task collection and agent learning; 
5) Diversity of categories - the software should be diverse to adequately represent and cover a wide range of real-world cases. 

As a result, we have shortlisted the software into two categories: general usage and professional usage. 
For general usage, we have VLC for media playback, Chrome for web browsing, and Thunderbird for email management.
For professional usage, we have VS Code as a coding IDE, and LibreOffice (Calc, Writer, and Impress) for handling spreadsheets, documents, and presentations respectively, along with GIMP for image editing. 
This brings our total to eight different types of software.

\subsection{Task Example Sources}
\label{sub:task_exam_srcs}

We detail the task example sources in Table~\ref{tab:task_exam_srcs}.

\begin{longtable}{cll}
    \caption{Task Example Resources} \label{tab:task_exam_srcs} \\
\textbf{App} & \textbf{Resources} & \textbf{Link} \\
\hline
\endfirsthead

\multicolumn{2}{c}%
{{\bfseries \tablename\ \thetable{} -- continued from previous page}} \\
\hline
\textbf{App} & \textbf{Resources} \\
\hline
\endhead

\multicolumn{2}{r}{{\textit{Continued on next page}}} \\ 
\endfoot

\hline
\endlastfoot

\multirow{5}{*}{OS} &    Ubuntu Documentations & \url{https://help.ubuntu.com/} \\
&   Ask Ubuntu & \url{https://askubuntu.com/} \\
&   Super User & \url{https://superuser.com/} \\
&   Stack Overflow & \url{https://stackoverflow.com} \\
&   YouTube & \url{https://www.youtube.com/} \\
\hline
\multirow{10}{*}{Calc} &   LibreOffice Help & \url{https://help.libreoffice.org/} \\
&   Microsoft Tech Community & \url{https://techcommunity.microsoft.com/} \\
&   libreofficehelp.com & \url{https://www.libreofficehelp.com/} \\
&   Reddit r/LibreOfficeCal & \url{https://www.reddit.com/r/LibreOfficeCalc/} \\
&   Reddit r/Excel & \url{https://www.reddit.com/r/Excel/} \\
&   Super User & \url{https://superuser.com/} \\
&   Medium & \url{https://medium.com/} \\
&   Quora & \url{https://www.quora.com/} \\
&   YouTube & \url{https://www.youtube.com/} \\
&   Ryan and Debi \& Toren Personal Site & \url{https://www.ryananddebi.com/} \\
\hline
\multirow{9}{*}{Writer} &   LibreOffice Help & \url{https://help.libreoffice.org/} \\
&   LibreOffice Forum & \url{https://ask.libreoffice.org/} \\
&   libreofficehelp.com & \url{https://www.libreofficehelp.com/} \\
&   Super User & \url{https://superuser.com/} \\
&   Stack Overflow & \url{https://stackoverflow.com} \\
&   Ask Ubuntu & \url{https://askubuntu.com/} \\
&   Quora & \url{https://www.quora.com/} \\
&   YouTube & \url{https://www.youtube.com/} \\
&   SearchStar Personal Site & \url{https://seekstar.github.io/} \\
\hline
\multirow{8}{*}{Impress} & LibreOffice Help & \url{https://help.libreoffice.org/} \\
&   libreofficehelp.com & \url{https://www.libreofficehelp.com/} \\
&   Reddit r/LibreOffice & \url{https://www.reddit.com/r/LibreOffice/} \\
& Super User & \url{https://superuser.com/} \\
&   Stack Overflow & \url{https://stackoverflow.com} \\
& Technical Tips & \url{https://technical-tips.com/} \\
&   Just Click Here & \url{https://justclickhere.co.uk/} \\
&   TikTok & \url{https://www.tiktok.com/} \\
\hline
\multirow{9}{*}{VLC} &  VLC Documentation & \url{https://docs.videolan.me} \\
&   VLCHelp.com & \url{https://www.vlchelp.com/} \\
&   VideoLAN's Wiki & \url{https://wiki.videolan.org/} \\
&   Ubuntu Documentations & \url{https://help.ubuntu.com/} \\
&   Reddit r/Fedora & \url{https://www.reddit.com/r/Fedora/}\\
&   Super User & \url{https://superuser.com/} \\
&   Medium & \url{https://medium.com/} \\
&   YouTube & \url{https://www.youtube.com/} \\
&   Dedoimedo & \url{https://www.dedoimedo.com/index.html} \\
\hline
\multirow{8}{*}{Thunderbird} &  Thunderbird Support & \url{https://support.mozilla.org/en-US/products/thunderbird}\\
&   Reddit r/Thunderbird & \url{https://www.reddit.com/r/Thunderbird/} \\
&   Reddit r/Automation & \url{https://www.reddit.com/r/automation/} \\
&   Super User & \url{https://superuser.com/} \\
&   WikiHow & \url{https://www.wikihow.com/} \\
&   Quora & \url{https://www.quora.com/} \\
&   BitRecover & \url{https://www.bitrecover.com/} \\
&   AdSigner & \url{https://www.adsigner.com/} \\
\hline
Chrome &    Google Chrome Help & \url{https://support.google.com/chrome} \\
&   Reddit r/Chrome & \url{https://www.reddit.com/r/Chrome/} \\
&   Super User & \url{https://superuser.com/} \\
&   WikiHow & \url{https://www.wikihow.com/} \\
&   in5steps.com & \url{https://in5stepstutorials.com/} \\
&   How-To Geek & \url{https://www.howtogeek.com/} \\
&   Medium & \url{https://medium.com/} \\
&   Quora & \url{https://www.quora.com/} \\
&   YouTube & \url{https://www.youtube.com/} \\
&   Laptop Mag & \url{https://www.laptopmag.com} \\
\hline
\multirow{5}{*}{VS Code} &   Super User & \url{https://superuser.com/} \\
&   Stack Overflow & \url{https://stackoverflow.com} \\
& Quora & \url{https://www.quora.com/} \\
&   YouTube & \url{https://www.youtube.com/} \\
&   Campbell Muscle Lab GitHub & \url{https://campbell-muscle-lab.github.io/} \\
\hline
\multirow{6}{*}{GIMP} &   Reddit r/GIMP & \url{https://www.reddit.com/r/GIMP/} \\
&   Super User & \url{https://superuser.com/} \\
&   Stack Overflow & \url{https://stackoverflow.com} \\
& Quora & \url{https://www.quora.com/} \\
&   Make-Use-Of & \url{https://www.makeuseof.com/} \\
&   YouTube & \url{https://www.youtube.com/} \\
\hline
\multirow{18}{*}{Workflow} & UniPath Marketplace & \url{https://marketplace.uipath.com/} \\
&   sync.blue & \url{https://www.sync.blue/} \\
&   Device Tests & \url{https://devicetests.com/} \\
&   Make Tech Easier & \url{https://www.maketecheasier.com/} \\
&   Unix \& Linux Stack Exchange & \url{https://unix.stackexchange.com/} \\
&   Geeks for Geeks & \url{https://www.geeksforgeeks.org/} \\
&   I Love Free Software & \url{https://www.ilovefreesoftware.com/} \\
&   The Geek Diary & \url{https://www.thegeekdiary.com/} \\
&   Zyxware & \url{https://www.zyxware.com/} \\
&   GNOME Discourse & \url{https://discourse.gnome.org/} \\
&   It's FOSS & \url{https://itsfoss.com/} \\
&   Super User & \url{https://superuser.com/} \\
&   Stack Overflow & \url{https://stackoverflow.com} \\
&   LibreOffice Forum & \url{https://ask.libreoffice.org/} \\
&   ImpressExtractNotes & \url{https://github.com/danielrcollins1/ImpressExtractNotes} \\
&    Medium & \url{https://medium.com/} \\
&   YouTube & \url{https://www.youtube.com/} \\
&   Kelvin Smith Library & \url{https://case.edu/library/} \\
\end{longtable}

\subsection{Task Examples Collection}
\label{app:examples_collection}

Here we show the detailed statistics of \ours benchmark, including the main set on Ubuntu (\numexamples in total) and the analytic set on Windows (\numexampleswindows in total).
\begin{table}[htbp]
\centering
\caption{
Detailed statistics of \ours benchmark suite about examples number, average instruction tokens, infeasible instructions and integrated instructions.
}
\label{tab:d-env-stats}
\resizebox{\textwidth}{!}{
\begin{tabular}{@{}lccccccccccc@{}}
\toprule
 & OS & Calc & Impress & Writer & VLC & Thunderbird & Chrome & VSCode & GIMP & \textbf{Workflow} & \textbf{Overall} \\
\midrule
Examples & 24 & 47 & 47 & 23 & 17 & 15 & 46 & 23 & 26 & 101 & 369 \\
Avg. Inst. Tokens & 22.38 & 33.30 & 25.19 & 35.30 & 35.82 & 34.07 & 22.07 & 20.78 & 16.23 & 51.24 & 33.36 \\
\#Infeasible & 5 & 1 & 0 & 1 & 3 & 1 & 3 & 5 & 10 & 1 & 30 \\
\#Integrated & 7 & 19 & 30 & 0 & 0 & 0 & 26 & 0 & 0 & 2 & 84 \\
\bottomrule
\end{tabular}
}
\end{table}

\begin{table}[htbp]
\centering
\caption{
Detailed statistics of Windows analytic set benchmark suite. This set contains no infeasible tasks and integrated tasks.
}
\label{tab:d-env-stats-windows}
\resizebox{0.5\textwidth}{!}{
\begin{tabular}{@{}lccccc@{}}
\toprule
 & Excel & Word & PPT & \textbf{Workflow} & \textbf{Overall} \\
\midrule
Examples & 11 & 9 & 7 & 16 & 43 \\
Avg. Inst. Tokens & 19.45 & 21.44 & 21.86 & 47.57 & 32.48\\
\bottomrule
\end{tabular}
}
\end{table}

\begin{figure}
    \centering
    \includegraphics[width=0.32\linewidth]{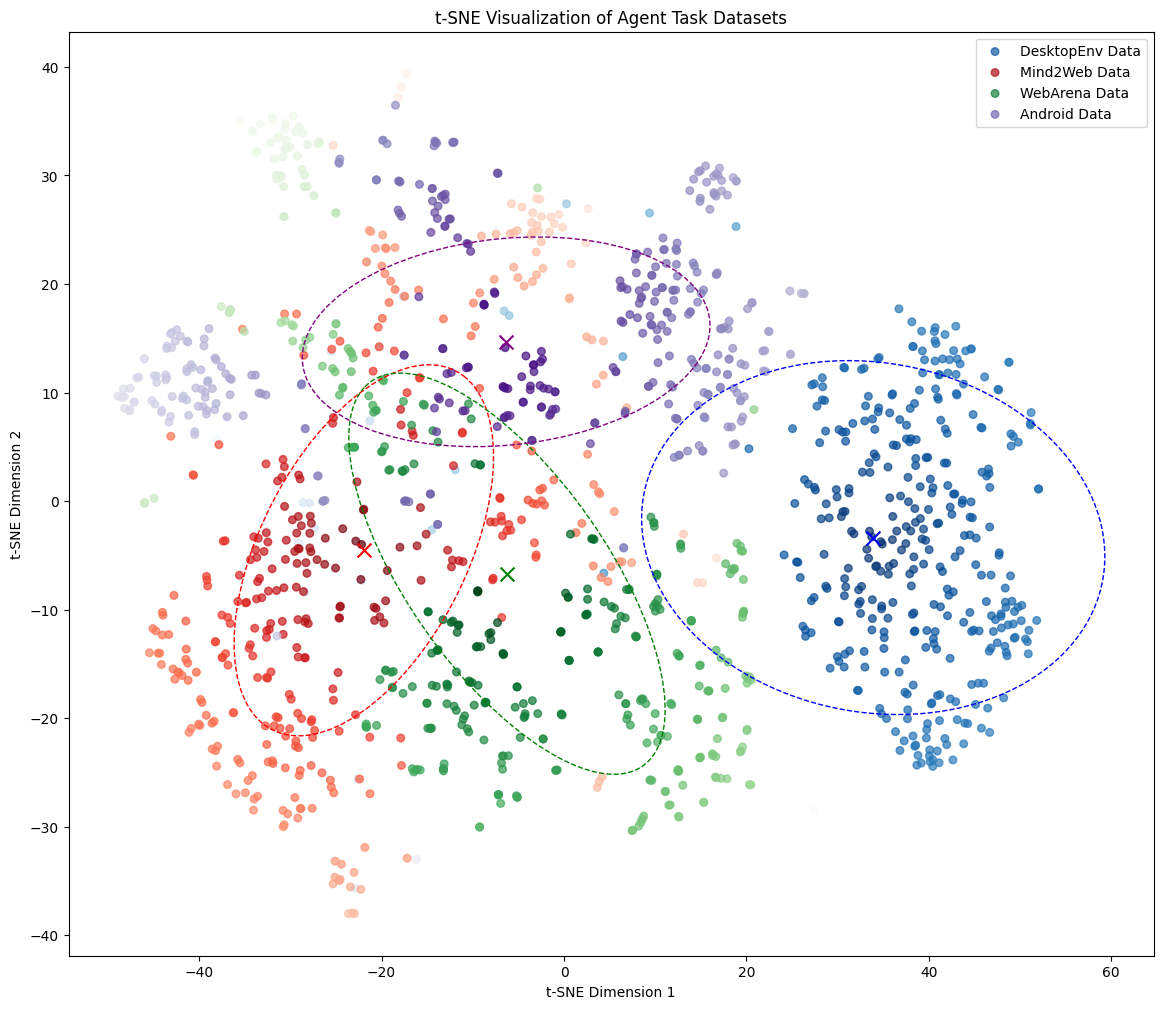}
    \includegraphics[width=0.32\linewidth]{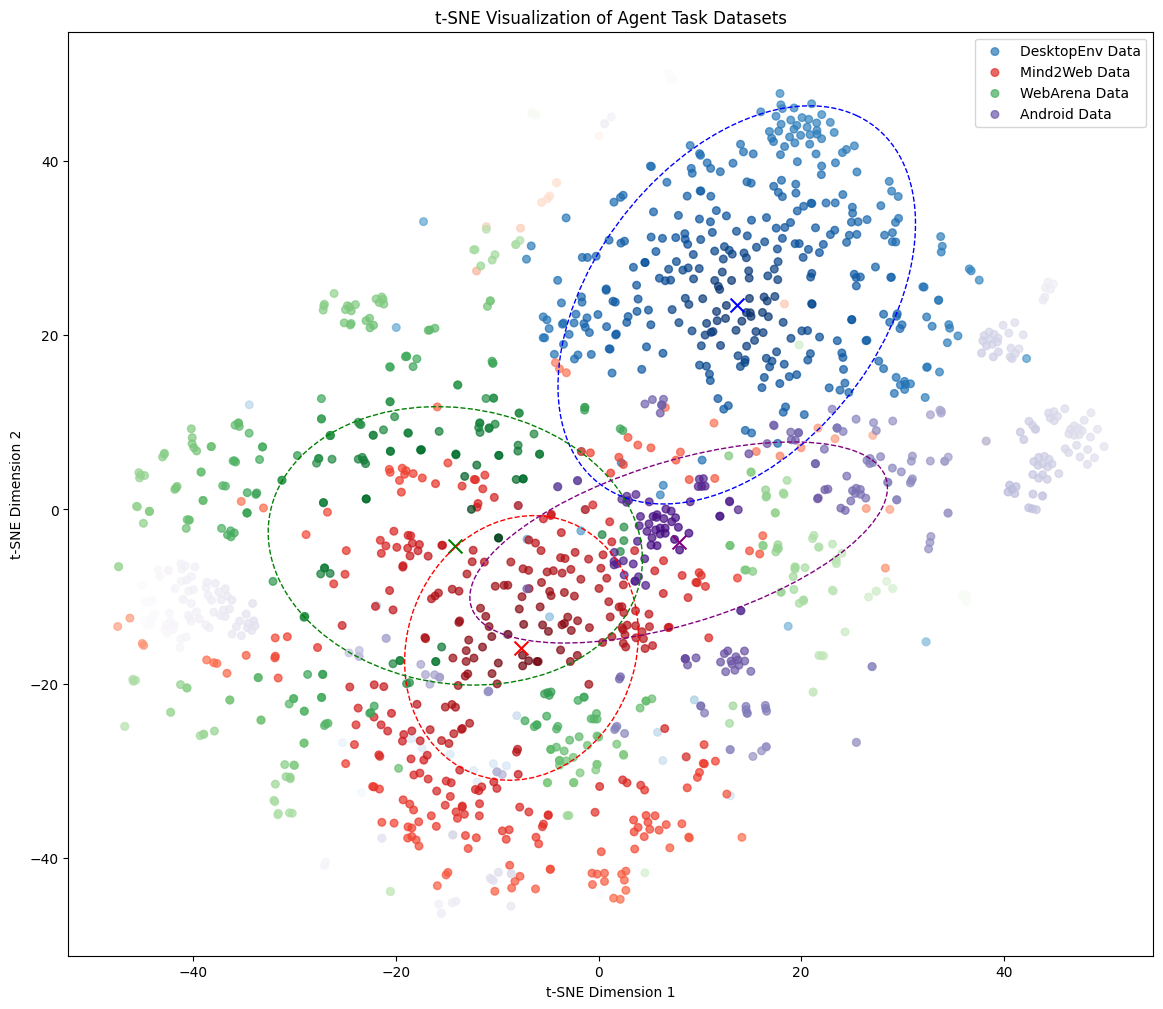}
    \includegraphics[width=0.32\linewidth]{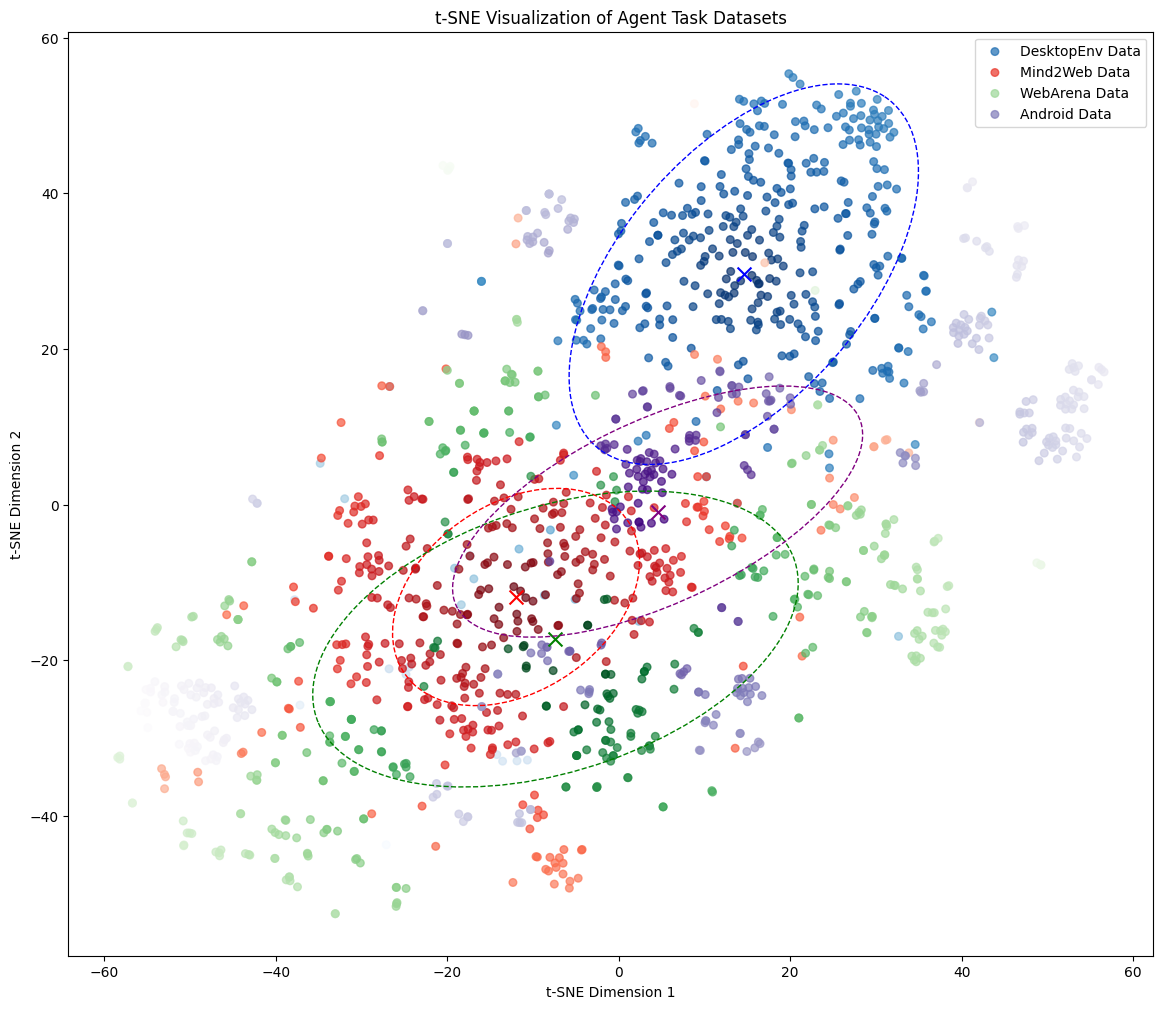}
    \caption{Comparison of instructions distribution. All datasets are sampled to 300 to make a fair comparison. The hyper-parameters of t-SNE are randomly sampled for each plot.}
    \label{fig:instructions_distribution_visualization}
\end{figure}

We also visualize the intent distribution (We obtain sentence embeddings for instructions using OpenAI's embedding model, and then apply t-SNE to reduce the dimensionality to two dimensions for visualization.) and compare it with other benchmarks which also focus on the digital agent.
We randomly sample 300 examples from each dataset and randomly choose three different hyperparameters for t-SNE.
Visualization results are shown in Figure~\ref{fig:instructions_distribution_visualization}.
From the figure, we can observe that the semantic distribution of the instructions alone has reached the most comprehensive level. 
Additionally, our environment remains controllable and executable, offering a more reliable evaluation. 
It is also noticeable that the clustering centers of the other three are closely positioned, whereas the points in our distribution approaches are inconsistent with theirs, indicating that we can serve as a unique choice for a more comprehensive assessment of the capabilities of future intelligent agents.

\subsection{Initial State Setup Details}
\label{app:initial_state_setup_details}
The setup of the initial state contains three stages:
1) Start emulator. 
The specified virtual machine is activated and automatically reverted to the corresponding snapshot, which records the initial system settings of the machine.
2) Prepare files (Optional). The file or software that specifies the initial state of the task to be executed is downloaded to the virtual machine and opened. 
The system is configured to first download the files to the host through a direct link and then upload them to the VM via a LAN connection. 
Specifically, some initial files are set up for OS-related tasks by manipulating the file system directly from the command line.
3) Execute reprocessing commands (Optional).
For tasks that require additional preprocessing, task-specific operations are executed after the completion of the first two phases. 
For example, taking the currently open LibreOffice Impress file to page five, clicking in the center of the screen to return to the main interface, \textit{etc}.
We provide convenient APIs to configure initial conditions and world settings, standardizing our tasks to make this process user-friendly and easily extendable.

\subsection{Evaluation Configuration Details}
\label{app:evaluation_configuration_details}
In this section, we will show details of preparations for the evaluation of the selected apps (LibreOffice -- Calc, Writer and Impress, Thunderbird, VLC Media Player, Chrome, VS Code, GIMP) and OS (Ubuntu and Windows).

\subsubsection{Ubuntu}

\paragraph{LibreOffice: Calc, Writer, and Impress}
LibreOffice is a popular open-source fully-featured office suite for Linux desktops. Our benchmark is built upon version 7.3.7.2, the 
version pre-installed in Ubuntu 22.04. Calc, Writer, and Impress are selected to build tasks on them. As the majority of tasks are to 
conduct a little revision to a file, we evaluate these tasks mainly by checking the final result file (in xlsx, docx, or pptx format).
The check can be done by comparing 
the result file with a golden reference, or inspecting some particular attributes or settings of the file, \textit{e.g.,}, page style,
freezing, and locale. Usually, the xlsx, docx, and pptx files are mainly accessed through 
\verb|openpyxl|\footnote{\url{https://openpyxl.readthedocs.io/en/stable/}},
\verb|python-docx|\footnote{\url{https://github.com/python-openxml/python-docx}}, and
\verb|python-pptx|\footnote{\url{https://github.com/scanny/python-pptx}}. For some properties not supported by the current libraries,
we also look them up directly via parsing the Office Open XML 
format\footnote{\url{https://learn.microsoft.com/en-us/office/open-xml/about-the-open-xml-sdk}}.

\paragraph{Thunderbird}
Thunderbird is a popular open-source fully-featured email client for Linux desktops. Version 115.6.0 of Thunderbird is pre-installed
in Ubuntu 22.04. We crafted an account profile to set up a feasible initial state. Evaluation for Thunderbird configurations
is mainly performed by reading various configurations or data files in the profile folder. An open-source reverse engineering
tool Firefox Decrypt\footnote{\url{https://github.com/unode/firefox_decrypt}} is adopted to decrypt the stored account 
information for evaluation. The involved account information is just for examples and contains no information about the real person.
Besides, there are tasks instructing to help to compose a new email. In these cases, the accessibility tree is leveraged to inspect 
the contents in the composing window before really sending it.

\paragraph{VLC Media Player}
VLC Media Player is a popular open-source cross-platform multimedia player and framework that plays most multimedia files. The evaluation for VLC Media Player is multifold, ranging from utilizing VLC HTTP interface\footnote{\url{https://wiki.videolan.org/Control_VLC_via_a_browser/}}, reading the VLC configuration file, comparing final result files, and leveraging accessibility tree to inspect the desired content.

\paragraph{Chrome}
Google Chrome is one of the most popular and powerful cross-platform web browsers developed on Google's free and open-source software project Chromium. 
The evaluation of Chrome is mainly based on the utilization of Playwright\footnote{\url{https://playwright.dev/}}, a browser automation library to control Chromium, Firefox, and WebKit with a single API. To connect Playwright running 
on host machine with Chrome running on virtual machine, port transferring tool 
socat\footnote{\url{http://www.dest-unreach.org/socat/},\url{https://linux.die.net/man/1/socat}} 
is leveraged. Additional information such as the HTML source codes of websites is also leveraged in the evaluation of some tasks.

\paragraph{VS Code}
VS Code is a popular open-source multi-functional cross-platform editor for source-code editing. 
The evaluation of VS Code tasks is primarily divided into two different categories. One subset of tasks is predominantly oriented towards file manipulation. 
In the context of these tasks, a comparative analysis is conducted between the resultant file and an anticipated reference gold file. 
Another subset of tasks is centered around how to utilize the intrinsic functionalities of the VS Code software itself, such as modifying color themes, initiating workspace sessions, and modifying settings. 
In these instances, it becomes important to extract relevant internal information and configurations from the VS Code environment.

In the course of this research, we principally leverage the capabilities offered by the VS Code Extension API\footnote{\url{https://code.visualstudio.com/api}} and information in the settings JSON file\footnote{\url{https://code.visualstudio.com/docs/getstarted/settings\#_settingsjson}} to obtain the requisite internal signal for the evaluation process. Our methodology involves the development of a custom VS Code extension, its installation within the VS Code software deployed on our virtual machine, and the subsequent invocation of the extension's command each time an evaluation is required, as well as checking whether the settings JSON has the correct value for a specific key.

\paragraph{GIMP}
GIMP is an open-source raster graphics editor used for image manipulation, editing, free-form drawing, format transcoding, and more specialized tasks. 
The evaluation for GIMP tasks is also mainly divided into two different categories, just like the VS Code evaluation. 
One type of task is mainly oriented to file operations.
In these tasks, the resulting files are compared and analyzed with the expected reference golden files, mainly relying on some powerful image processing libraries such as \verb|pillow|\footnote{\url{https://pypi.org/project/pillow/}}. 
Another category of tasks revolves around taking advantage of the inherent capabilities of the GIMP software itself. 
In these instances, we primarily read GIMP's configuration files to obtain internal information to evaluate the tasks.

\subsection{Windows}

\paragraph{Microsoft Office: Excel, Word, and PowerPoint} Microsoft Office is the most
popular office suite on Windows desktops. These three components share the same functions
with the corresponding LibreOffice components by and large. They are used to edit xlsx, docx,
and pptx files, respectively. Thus, the evaluation for LibreOffice tasks can be reused
for Microsoft Office tasks.

\paragraph{Thunderbird} Thunderbird is a cross-platform email client. Only the 
structure of profile folder on Windows is sightly
different from that on Linux. We thus revised the account profile and reuse it to 
set up the same initial state on Windows.

\paragraph{Chrome} Chrome is a cross-platform web browser. To evaluate tasks on 
Chrome, only the port transferring tool needs to be replaced with 
Ncat\footnote{\url{http://www.dest-unreach.org/socat/}}. Other configurations and the
evaluations can be shared with Linux-version tasks.

\subsection{More Task Examples}
\label{app:representitive_examples}
In this section, we curate a collection of examples from various app sets, each characterized by distinct operational logic and requiring different capabilities. 
These examples are carefully chosen to illustrate the diverse challenges and requirements encountered when interacting with different types of applications.

\begin{longtable}{m{1.5cm}m{4cm}m{6cm}m{2cm}}
\caption{More Example Showcase from Each Subset of Domains.} \label{tab:representative_cases} \\
\textbf{Related App(s)} & \textbf{Instruction(s)} & \textbf{Screenshot} & \textbf{Abilities Needed} \\
\hline
\endfirsthead

\multicolumn{4}{c}%
{{\bfseries \tablename\ \thetable{} -- continued from previous page}} \\
\hline
\textbf{Related App(s)} & \textbf{Task Instruction} & \textbf{Screenshot of Initial State} & \textbf{Abilities Needed} \\
\hline
\endhead

\multicolumn{4}{r}{{\textit{Continued on next page}}} \\ 
\endfoot

\endlastfoot

OS & \textit{I want to install Spotify on my current system. Could you please help me?} & \includegraphics[width=6cm, height=3.38cm]{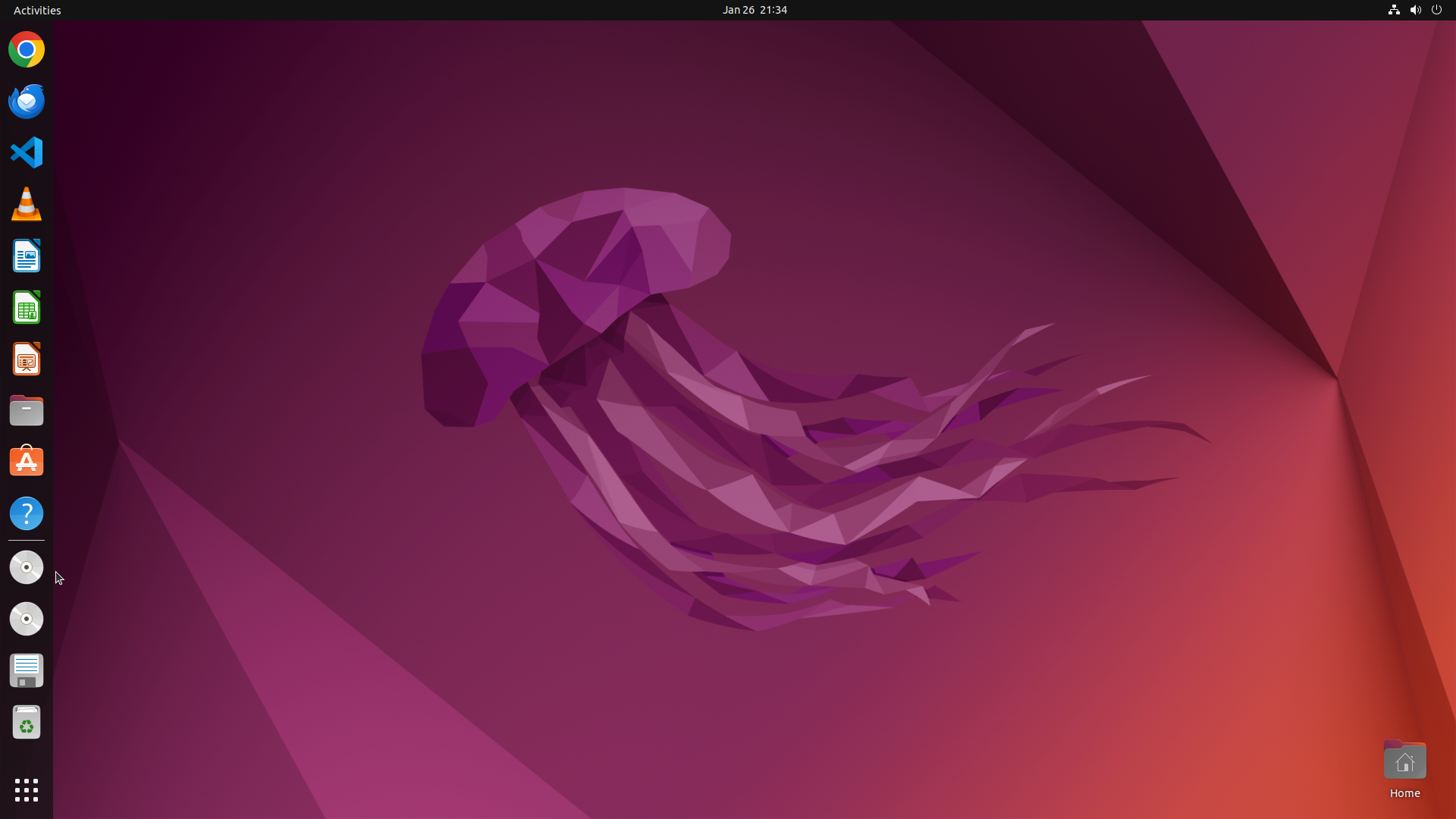} &  \texttt{knowledge of OS; omit distractions} \\
\hline
Calc &  \textit{Check the names in column ``Names with duplicates'' and put the unique ones in column ``Unique Names''. Keep the original order.} & \includegraphics[width=6cm, height=3.38cm]{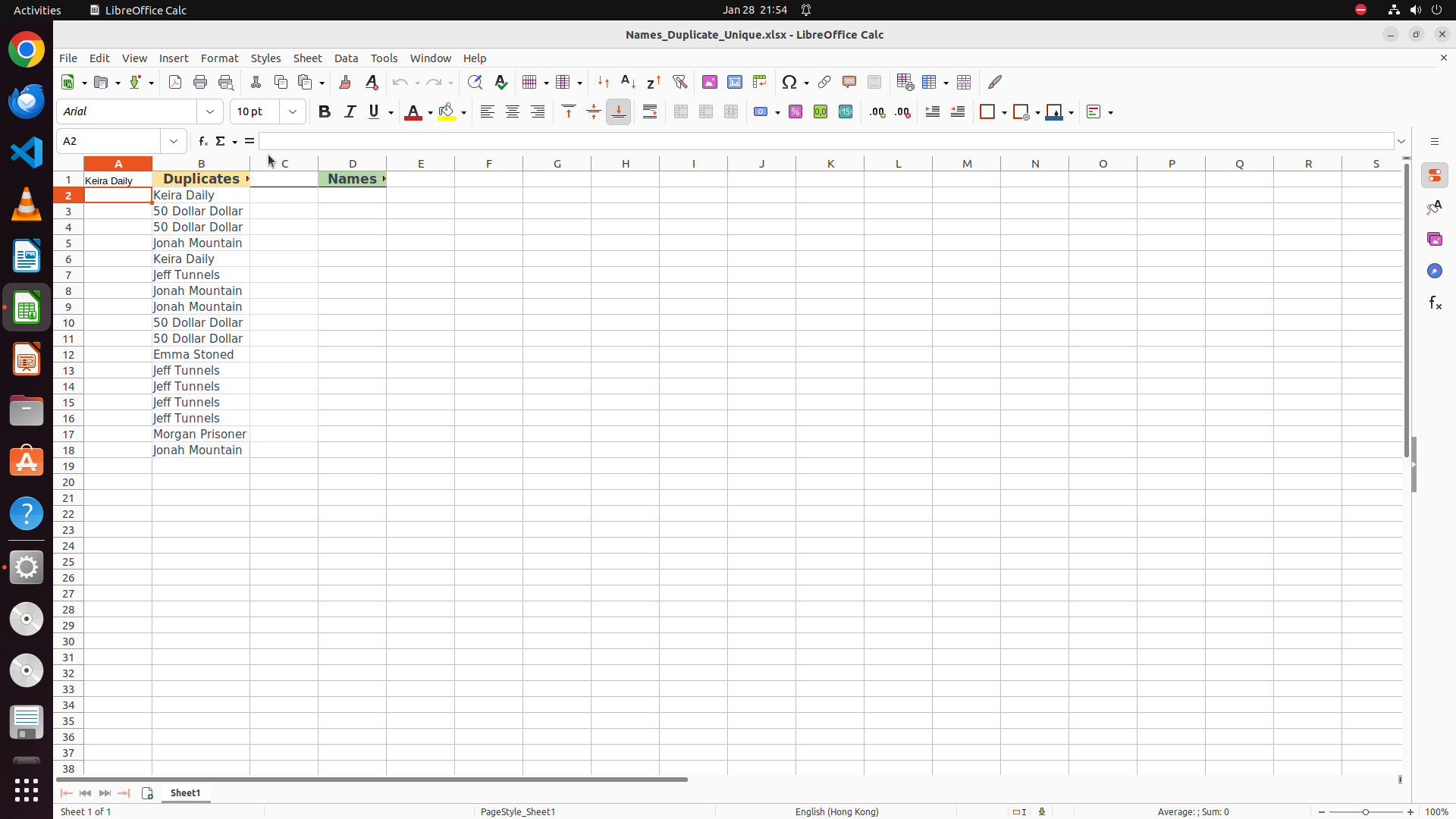} & \texttt{massive elements; knowledge tricks or reasoning over long actions} \\
\hline
Calc & \textit{I have a lookup table for the officers of each branch. Please, here is another table in which I need to fill with the officer names according the headoffice (i.e., the branch name). Help me to complete this.} & \includegraphics[width=6cm, height=3.38cm]{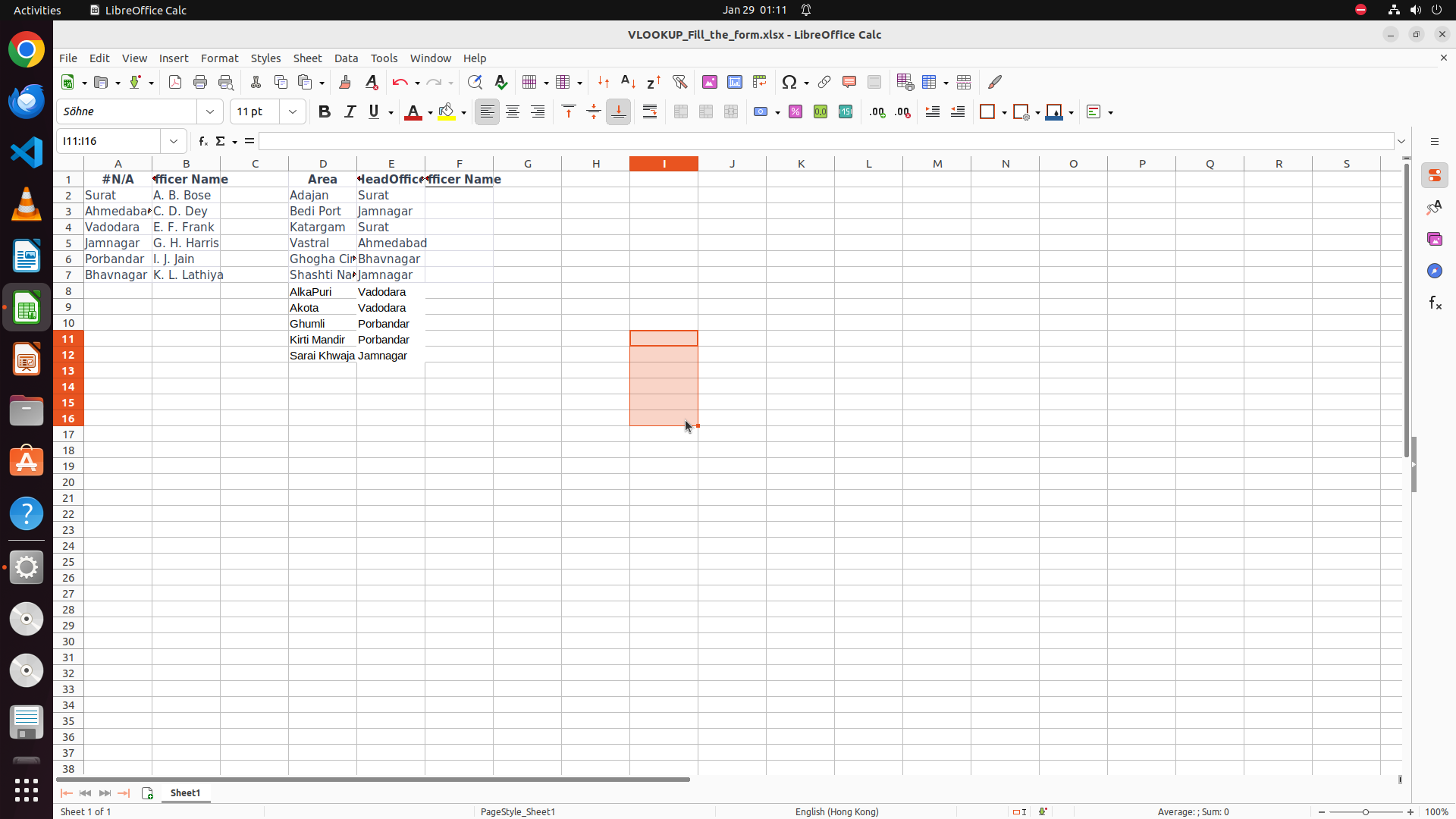} & \texttt{massive elements; knowledge of formulas and functions} \\
\hline
Calc & \textit{Given a partial calendar, please highlight all the weekends (Saturday \& Sunday) by setting the cell background as red (\#ff0000).}  & \includegraphics[width=6cm, height=3.38cm]{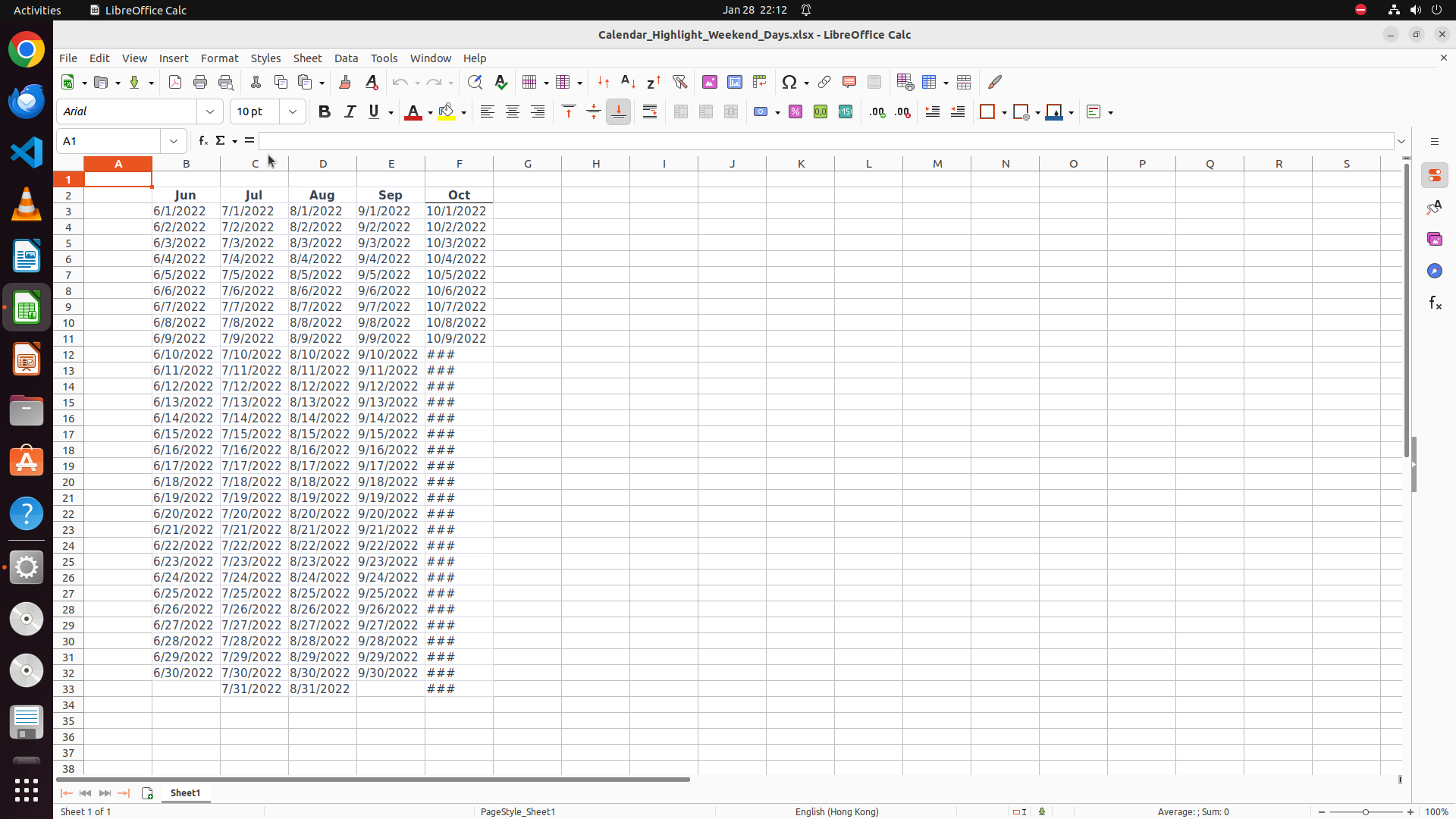} & \texttt{massive elements; commonsense reasoning; software tricks} \\
\hline
Impress & \textit{I closed the slide pannel on the left and idk how to get it back please help} & \includegraphics[width=6cm, height=3.38cm]{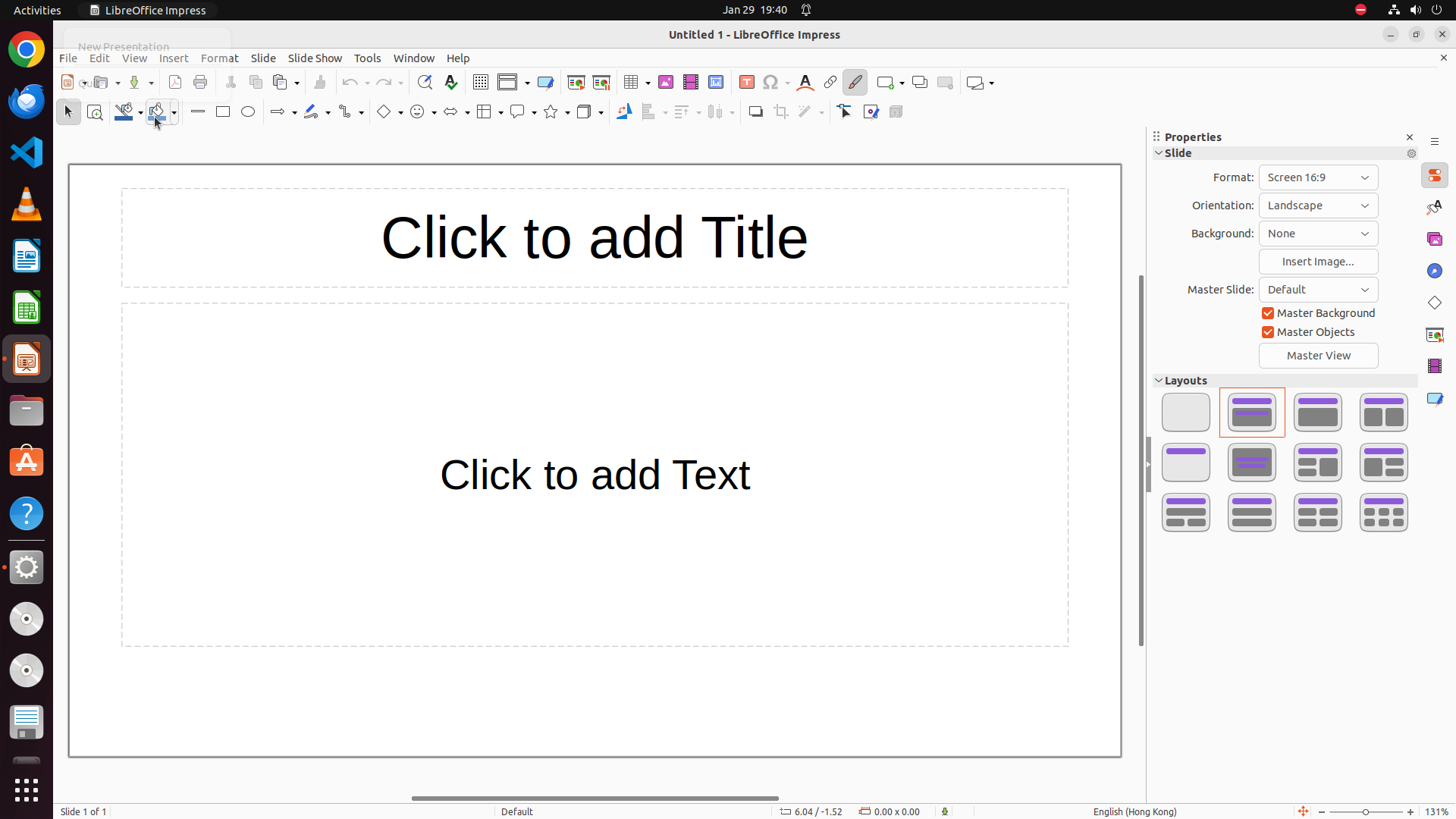} &  \texttt{software knowledge; imagine about UI layouts; overcome typos in instruction} \\
\hline
Impress & \textit{On it Whenever I launch a LibreOffice Impress, it uses both screens, one for current slide and next slide and another for actual presentation. What I want is to use only one monitor which shows presentation. I dont want the screen with Current slide and Next slide so that it can be used for other purposes. How should I achieve this?} & \includegraphics[width=6cm, height=3.38cm]{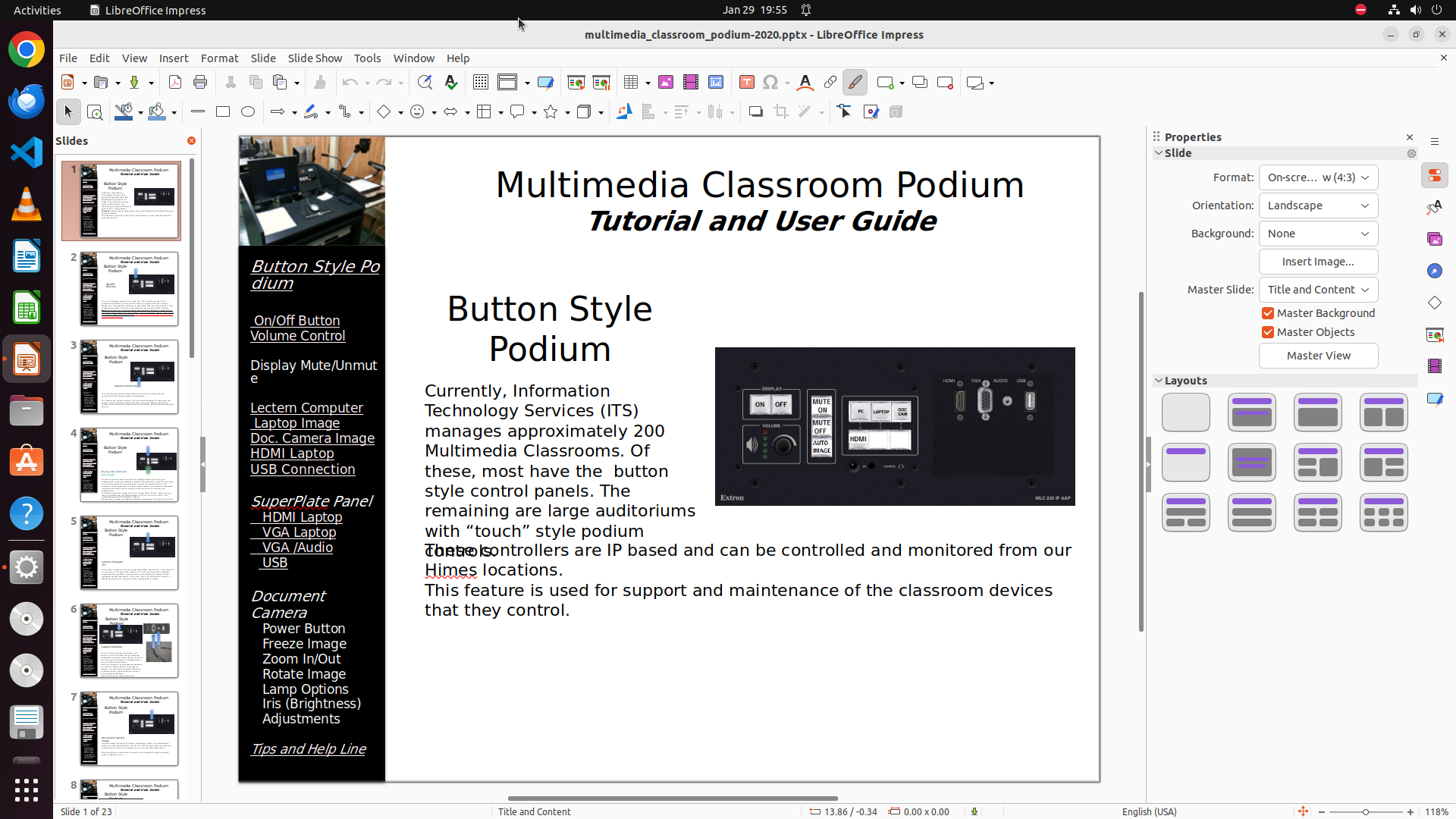} &  \texttt{reason from unprofessional phenomenon expression} \\
\hline
Writer & \textit{Copy the screenshot 1.png from the desktop to where my cursor is located} & \includegraphics[width=6cm, height=3.38cm]{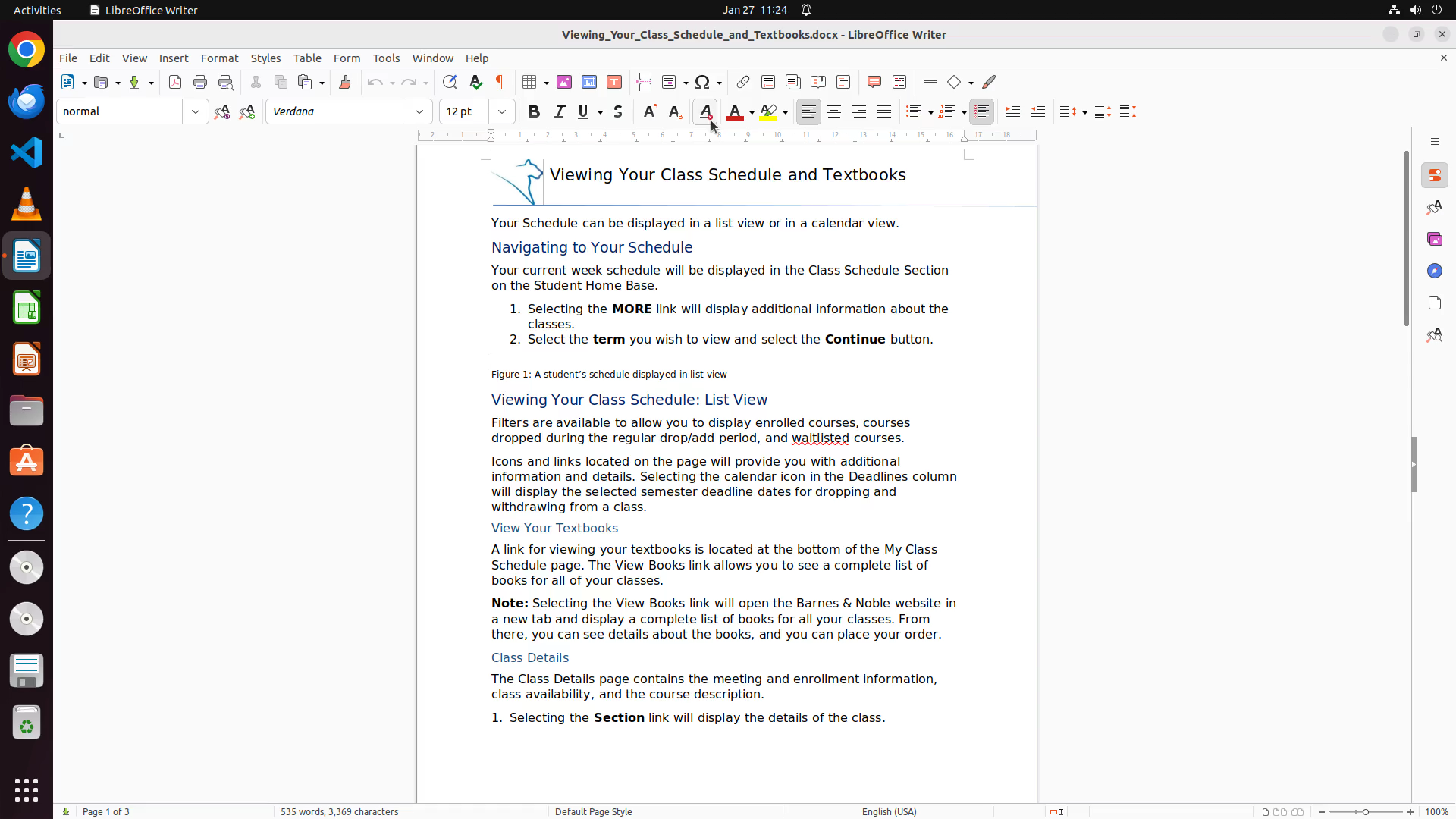} & \texttt{locate the position of cursor; switch from desktop and app} \\
\hline
Chrome &  \textit{Can you help me clean up my computer by getting rid of all the tracking things that Amazon might have saved? I want to make sure my browsing is private and those sites don't remember me.} & \includegraphics[width=6cm, height=3.38cm]{images/cases/chrome_clean_cookies.png} &  \texttt{understanding the unprofessional expression} \\
\hline
VLC & \textit{I am reading lecture note in PDF while a music video is running in VLC media player. But I find I need to switch to the player every time I need to pause/start.Could you help me change the setting to allow pausing the video using keyboard shortcut without minimizing the PDF reader? I want to focus on the lecture note and don't be disturbed by the app switching.} & \includegraphics[width=6cm, height=3.8cm]{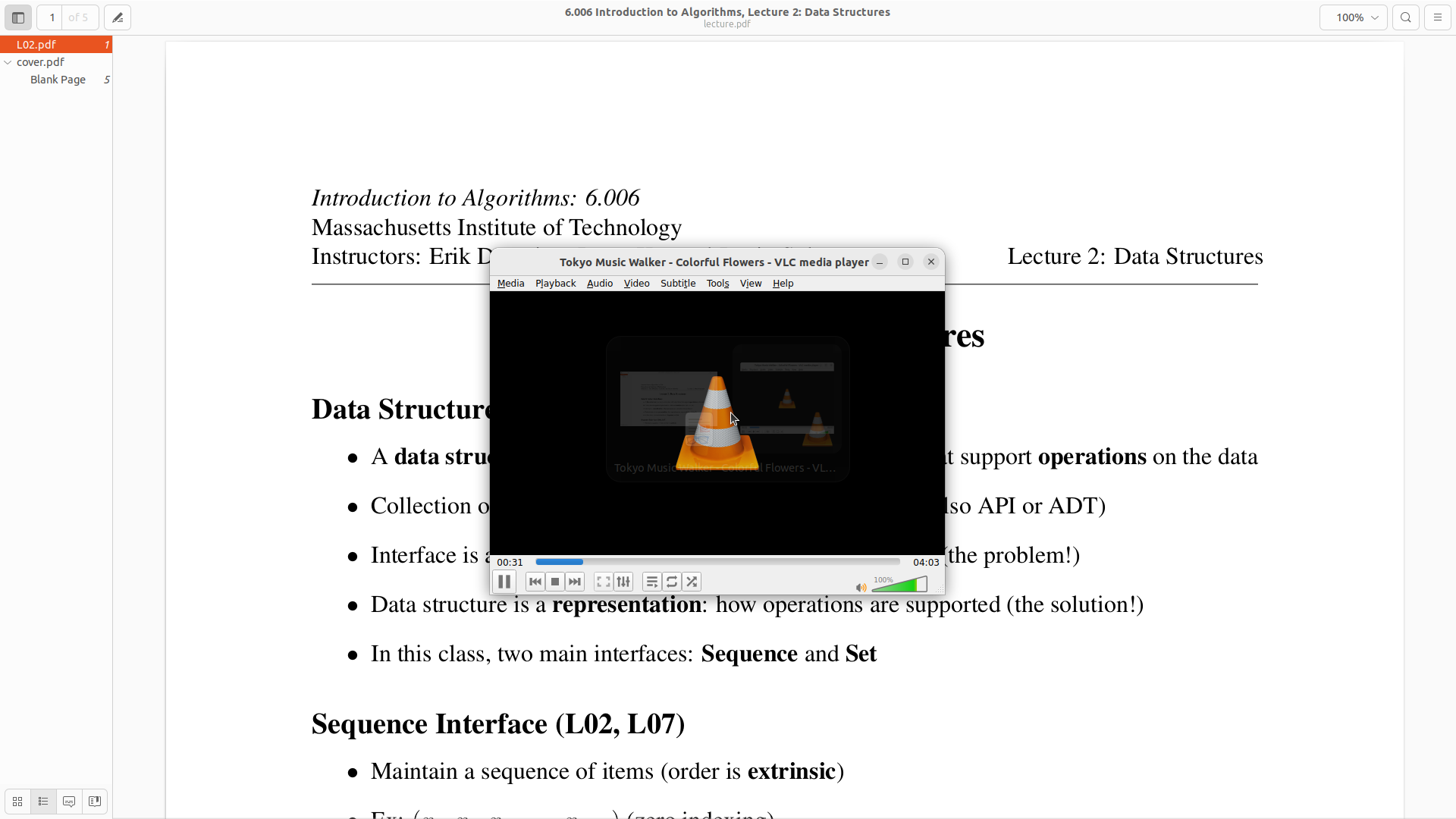} & \texttt{understanding the reference from unprofessional expression; software knowledge} \\
\hline
VLC & \textit{Hey, could you turn this video the right way up for me? And once it's flipped around, could you save it for me with the name `1984\_Apple.mp4' on the main screen where all my files are?} & \includegraphics[width=6cm, height=3.8cm]{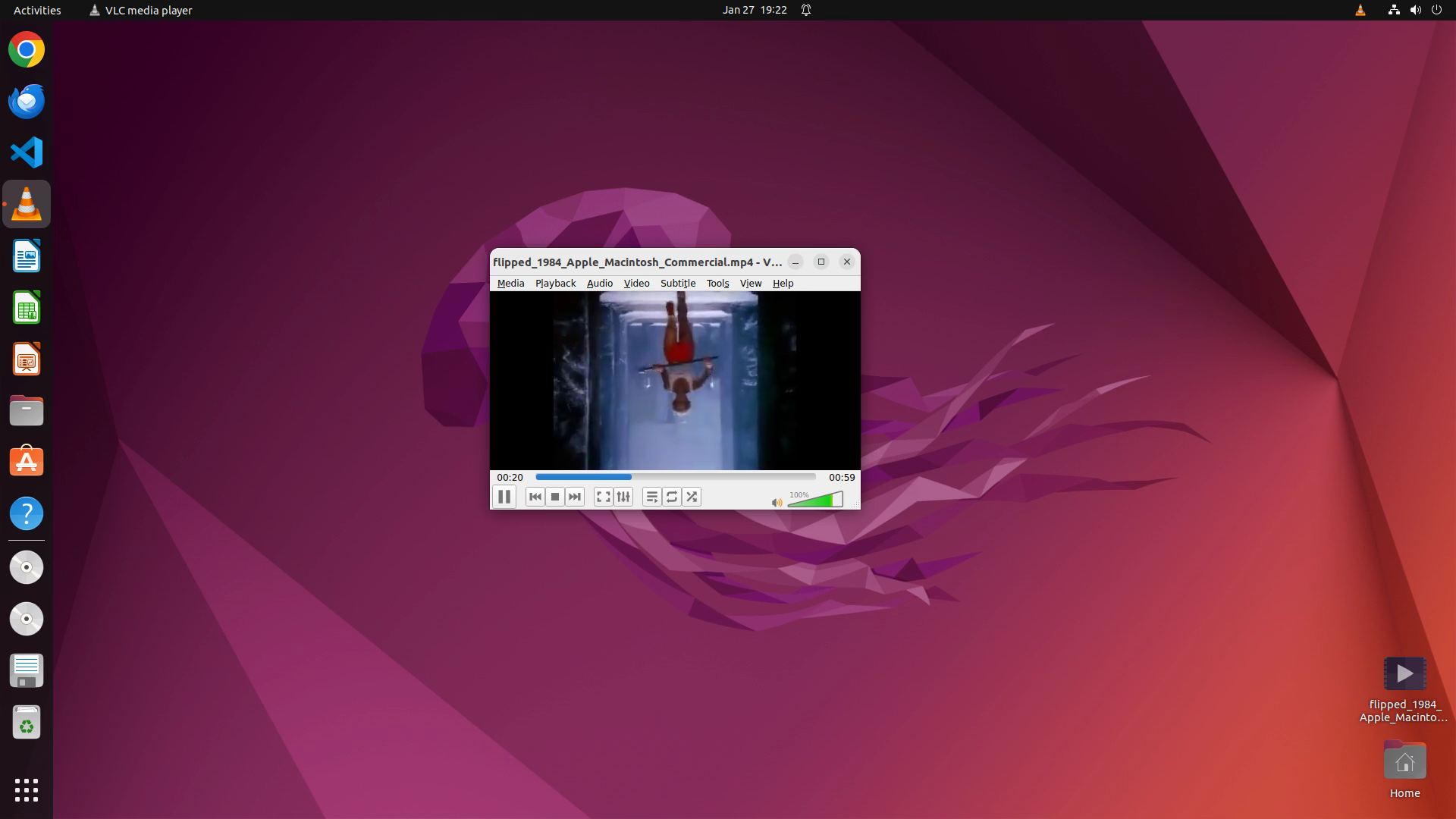} &  \texttt{software knowledge; spatial judgment ability} \\
\hline
Thunderbird & \textit{Create a local folder called "Promotions" and create a filter to auto move the inbox emails whose subject contains “discount” to the new folder} & \includegraphics[width=6cm, height=3.38cm]{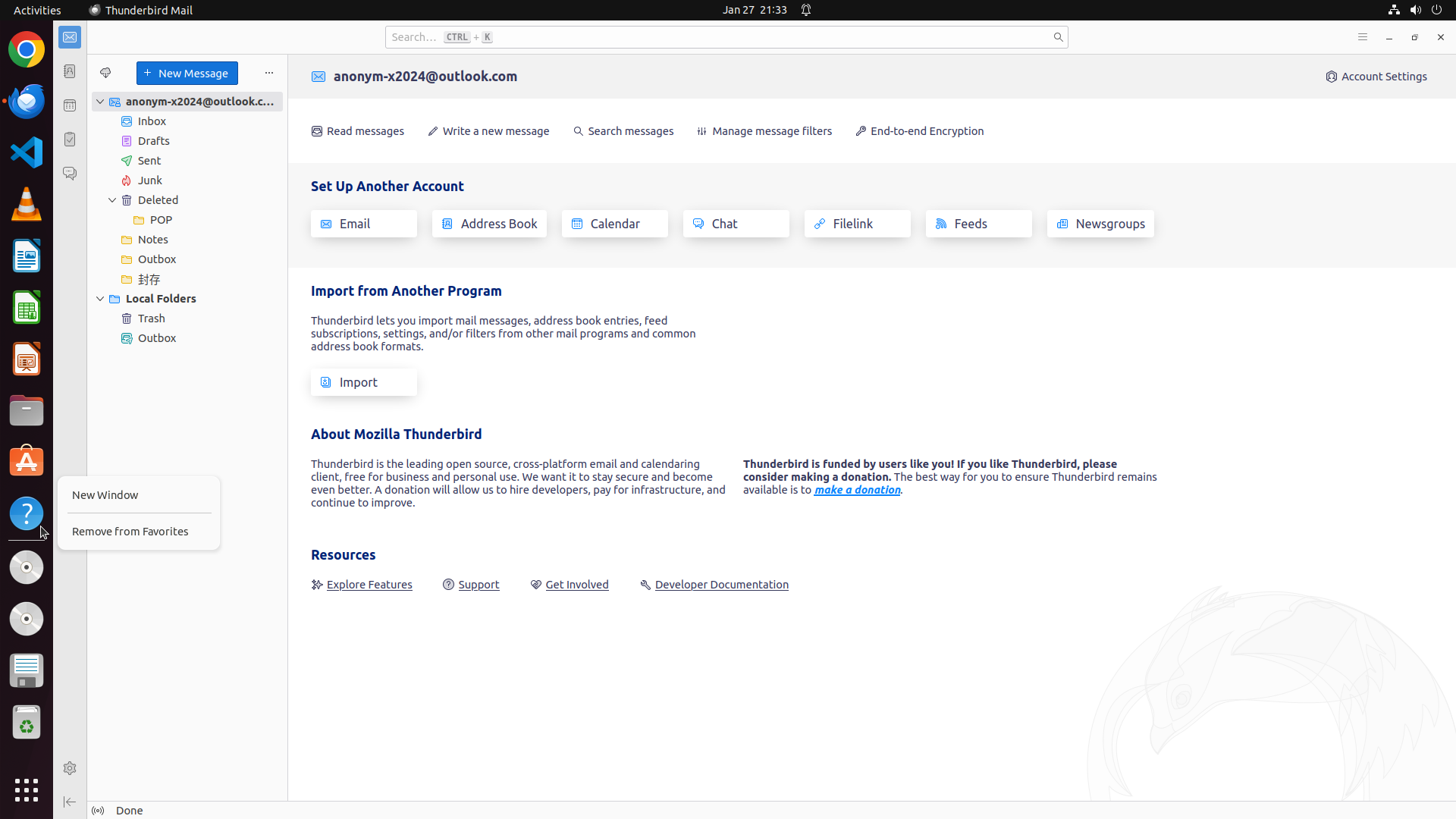} & \texttt{software knowledge} \\
\hline
Thunderbird & \textit{Attach the my AWS bill to the email. The bill is stored at ~/aws-bill.pdf. Don't close it or send it. I haven't finish all the contents.} & \includegraphics[width=6cm, height=3.38cm]{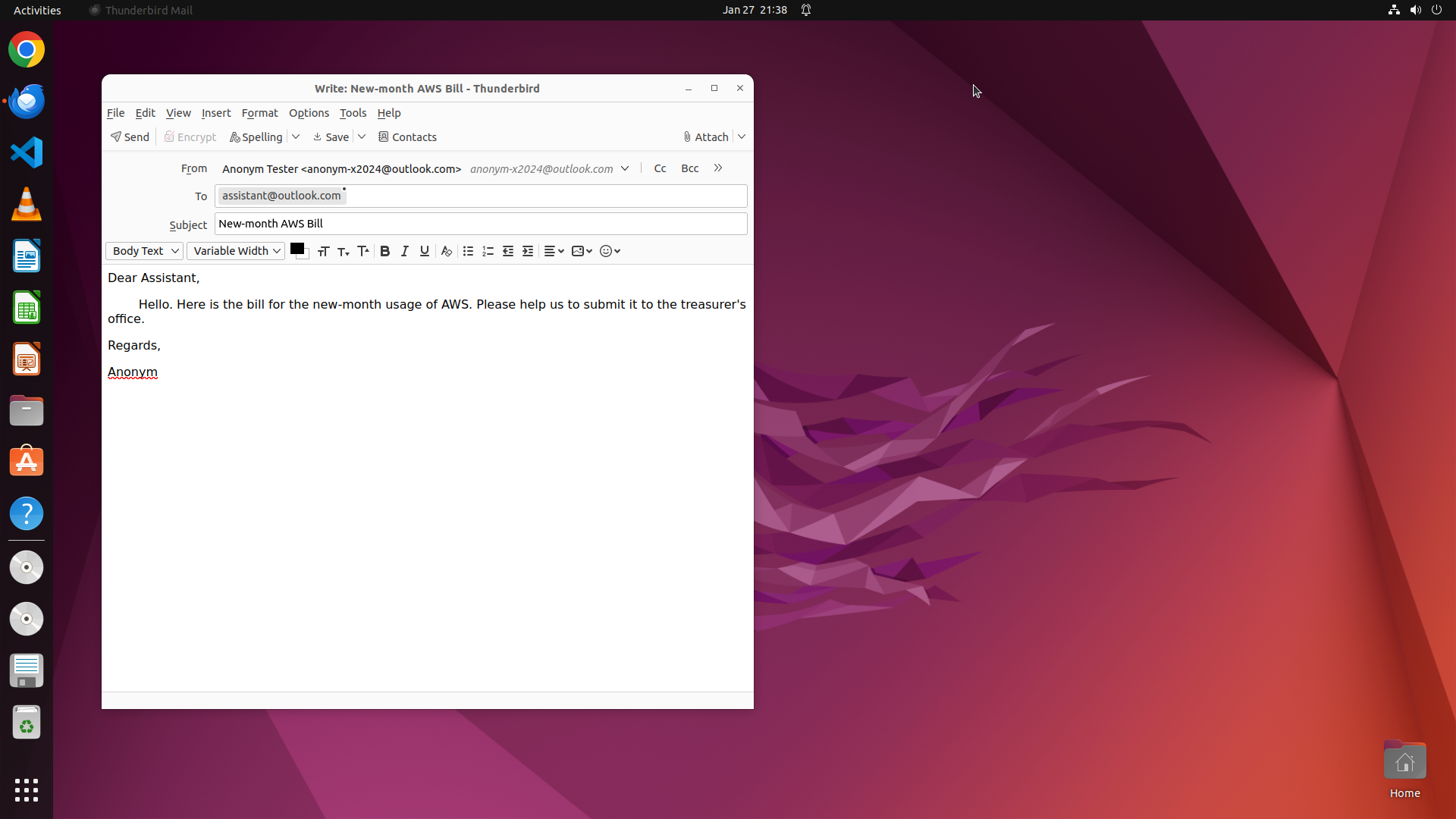} & \texttt{file management; extra requirement} \\
\hline
Thunderbird & \textit{I've got a bunch of email accounts in Thunderbird, and it's a hassle to check them one by one. Can you show me how to set up a unified inbox so I can see all my emails in one place?} & \includegraphics[width=6cm, height=3.38cm]{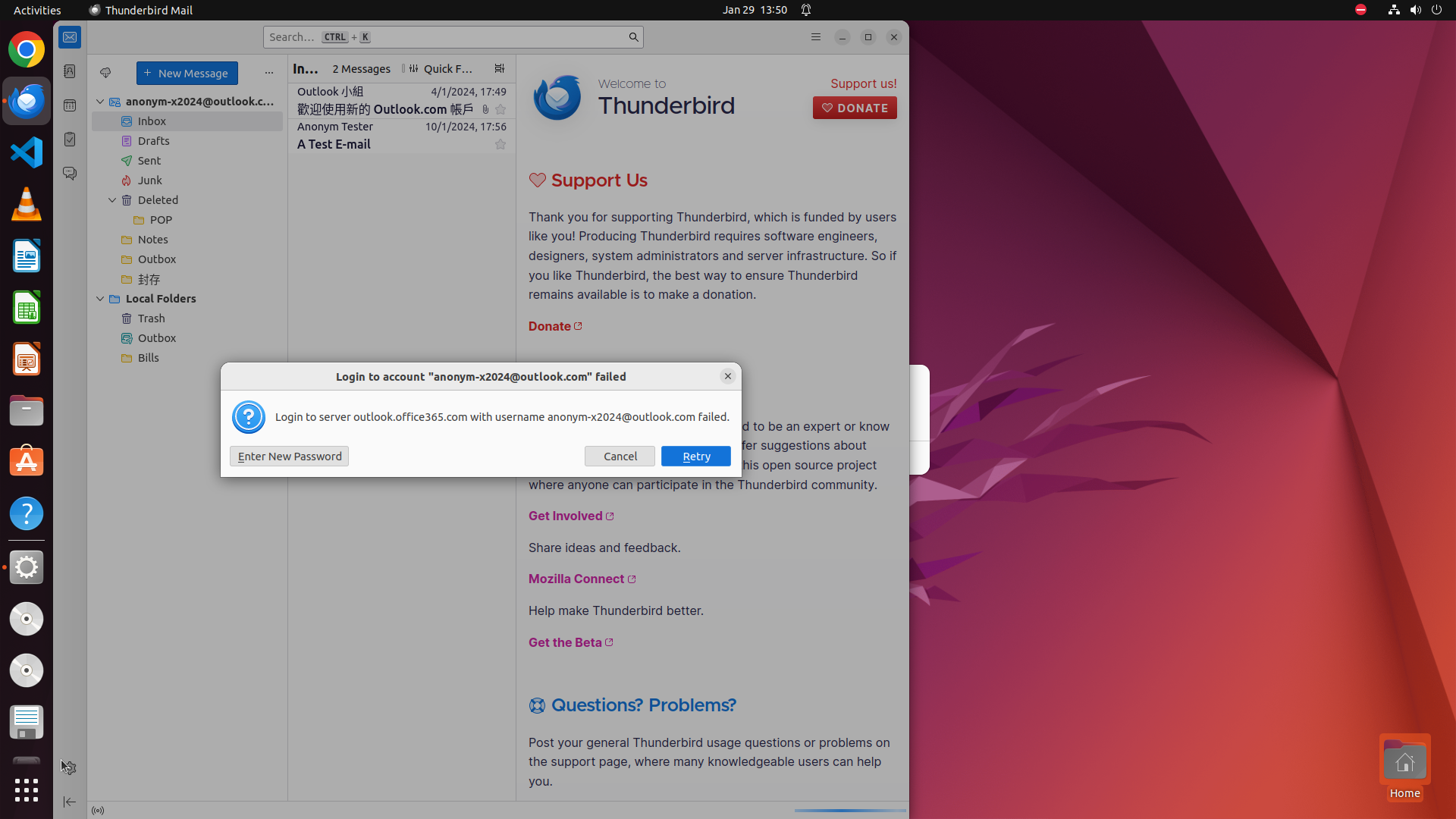} & \texttt{deep-hided feature, need to be explored even by human users; pop-up window} \\
\hline
VS Code & \textit{Please modify VS Code's settings to disable error reporting for Python missing imports.} & \includegraphics[width=6cm, height=3.38cm]{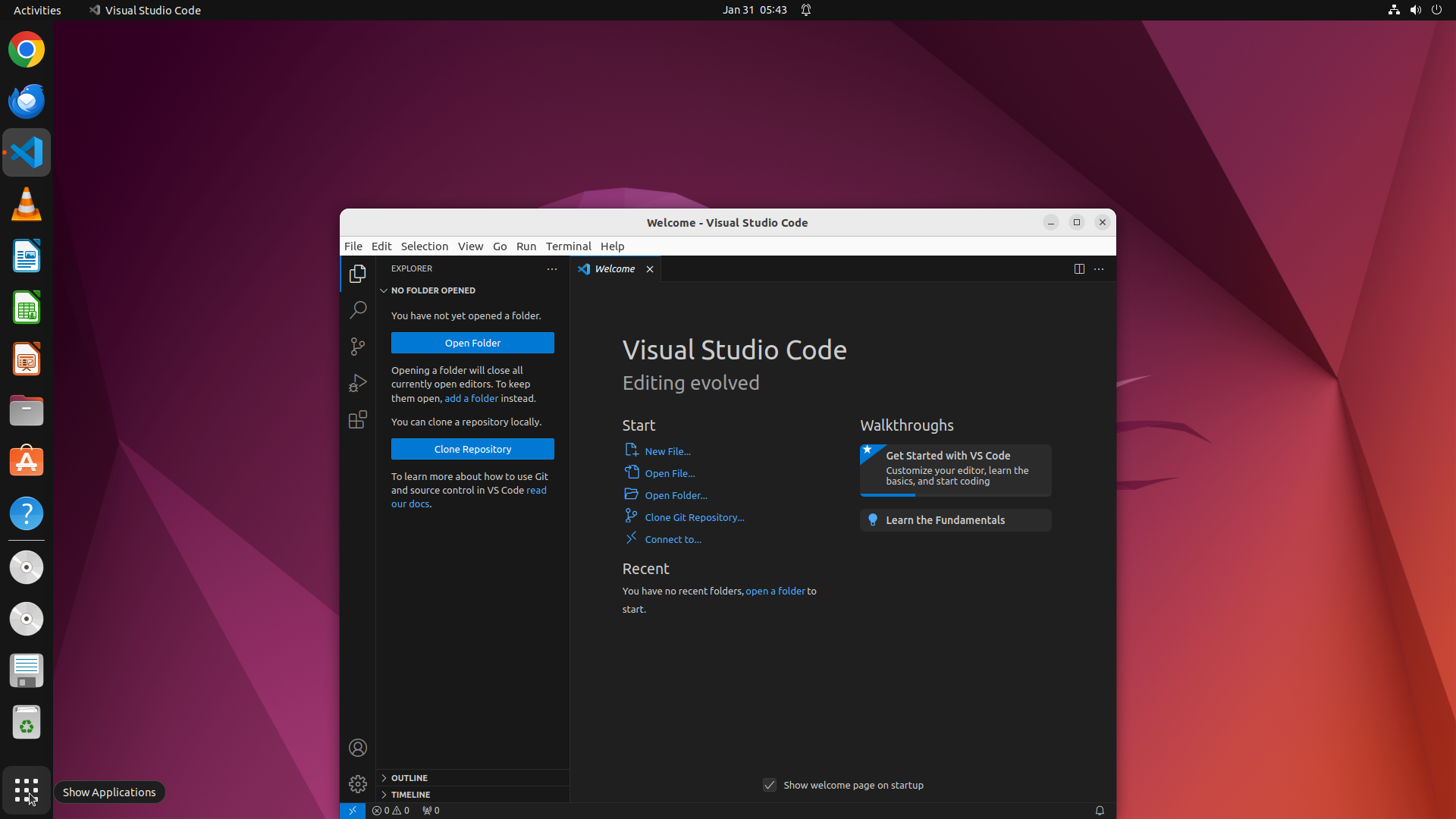} & \texttt{software knowledge to deal with settings; reasoning to understand the cause and solution of the error} \\
\hline
VS Code & \textit{Please help me install the autoDocstring extension in VS Code.} & \includegraphics[width=6cm, height=3.38cm]{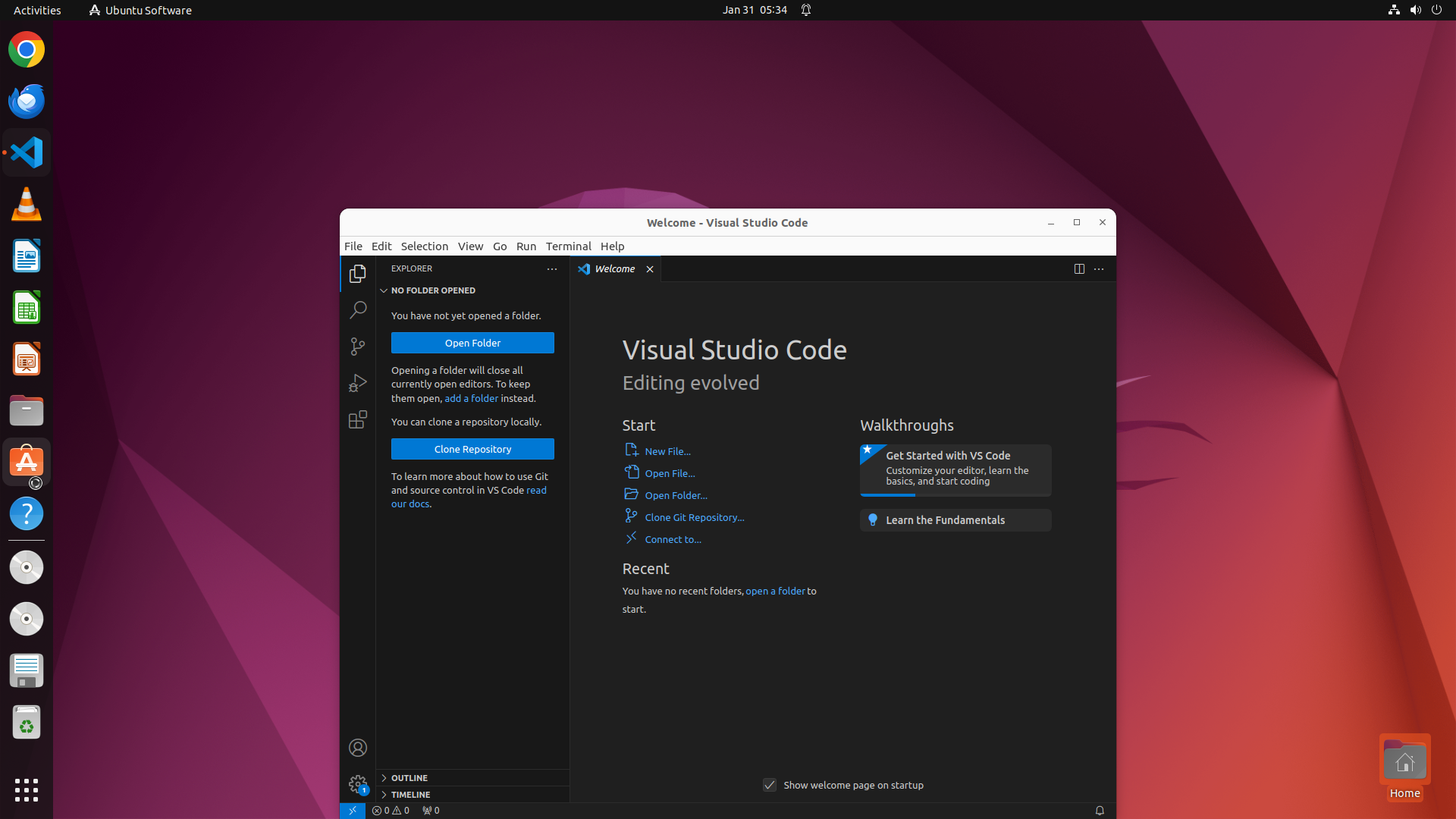} & \texttt{software knowledge to deal with Extensions; reasoning to search and install the extension successfully} \\
\hline
GIMP & \textit{Could you make the background of this image transparent for me?} & \includegraphics[width=6cm, height=3.38cm]{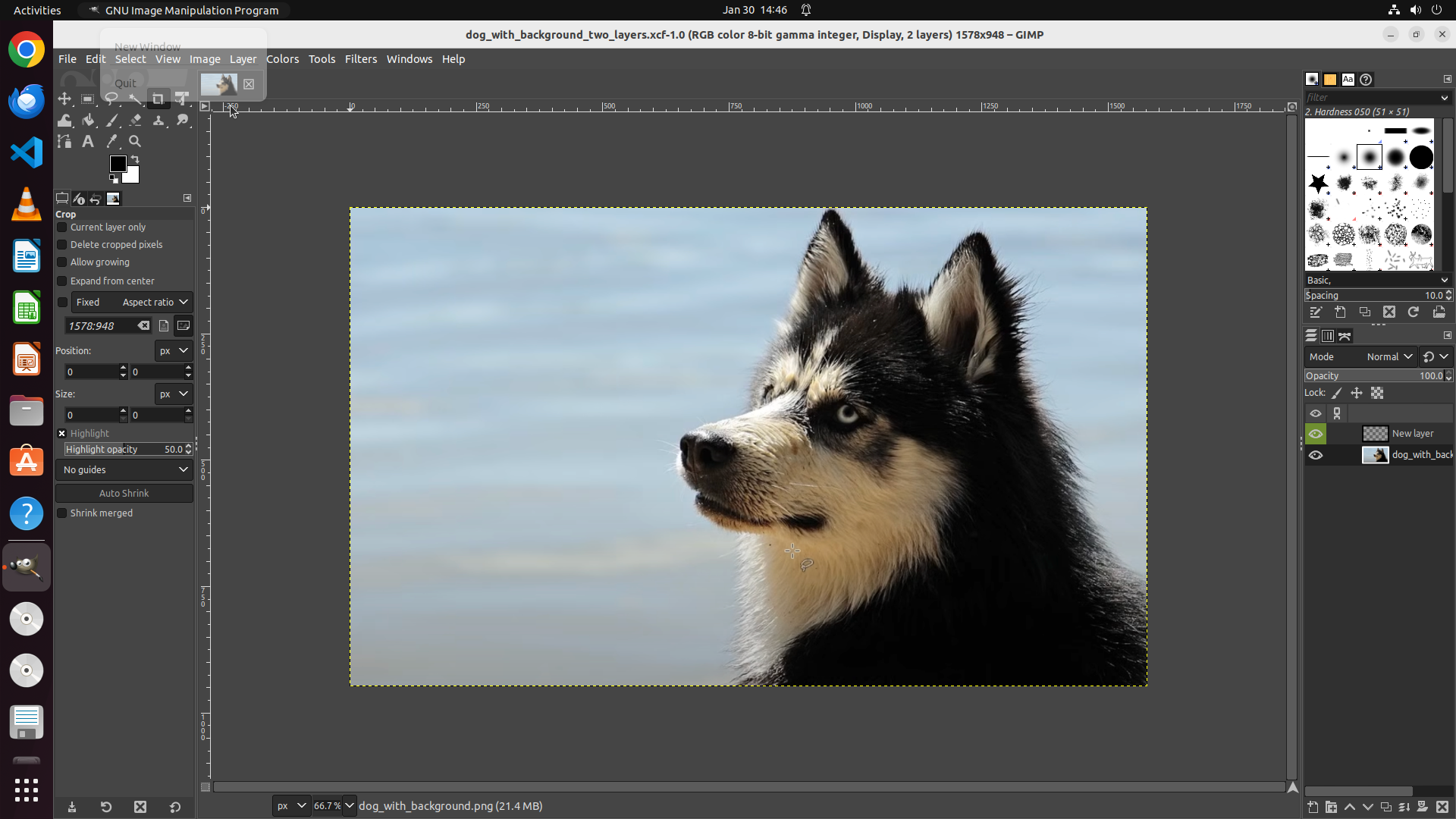} & \texttt{precise and intricate operations} \\
\hline
GIMP & \textit{Help me choose the yellow triangle and position it at the center of my picture.} & \includegraphics[width=6cm, height=3.38cm]{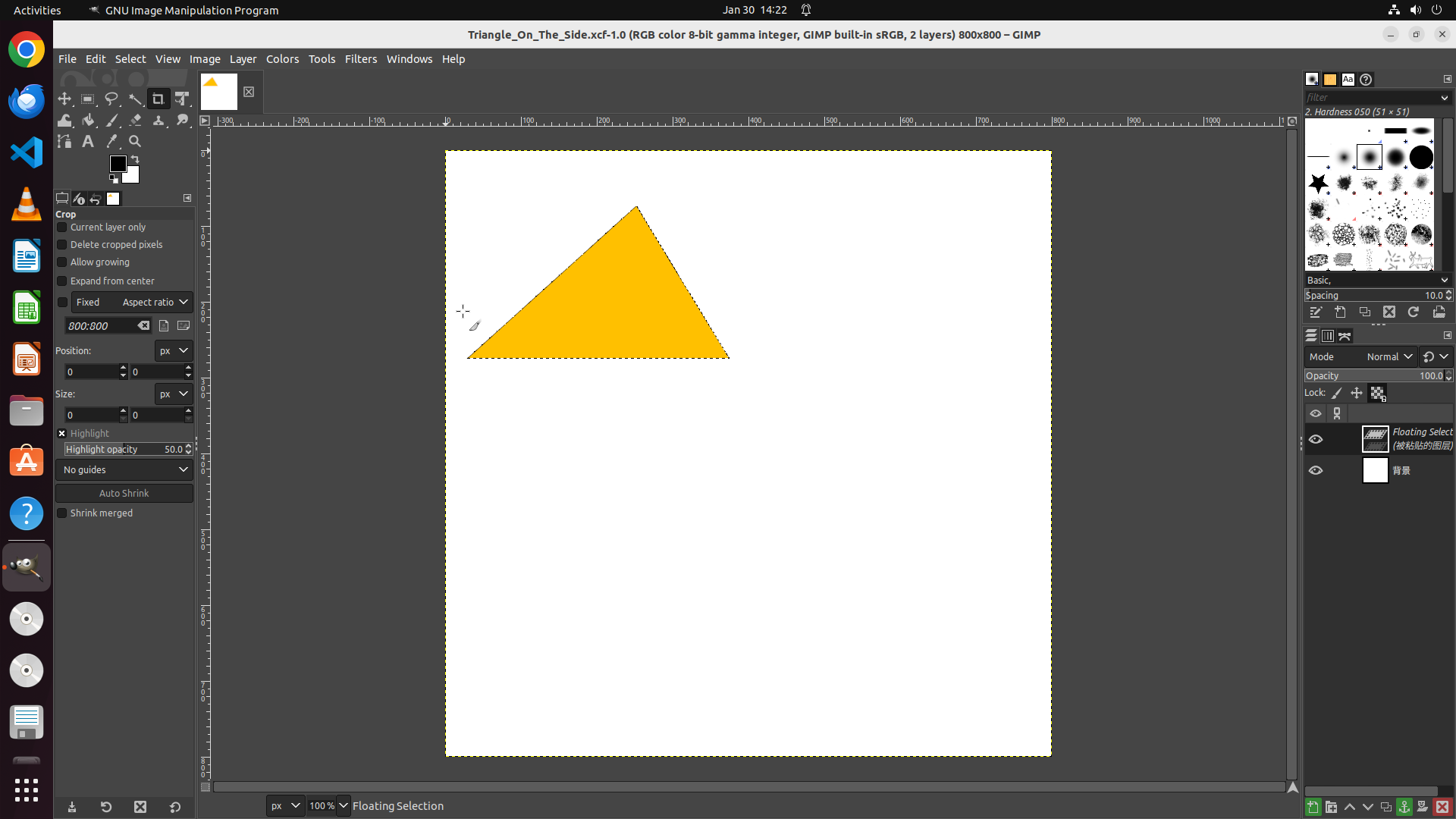} &  \texttt{spatial perception and reasoning, as well as precise control of actions} \\
\hline
Multiple {\fontsize{7pt}{10pt}\selectfont (VLC+GIMP)} & \textit{Could you help me create an Animated GIF from a video file using VLC and GIMP from the source of video ``src.mp4'', 5-second clip beginning at 00:03?} & \includegraphics[width=6cm, height=3.38cm]{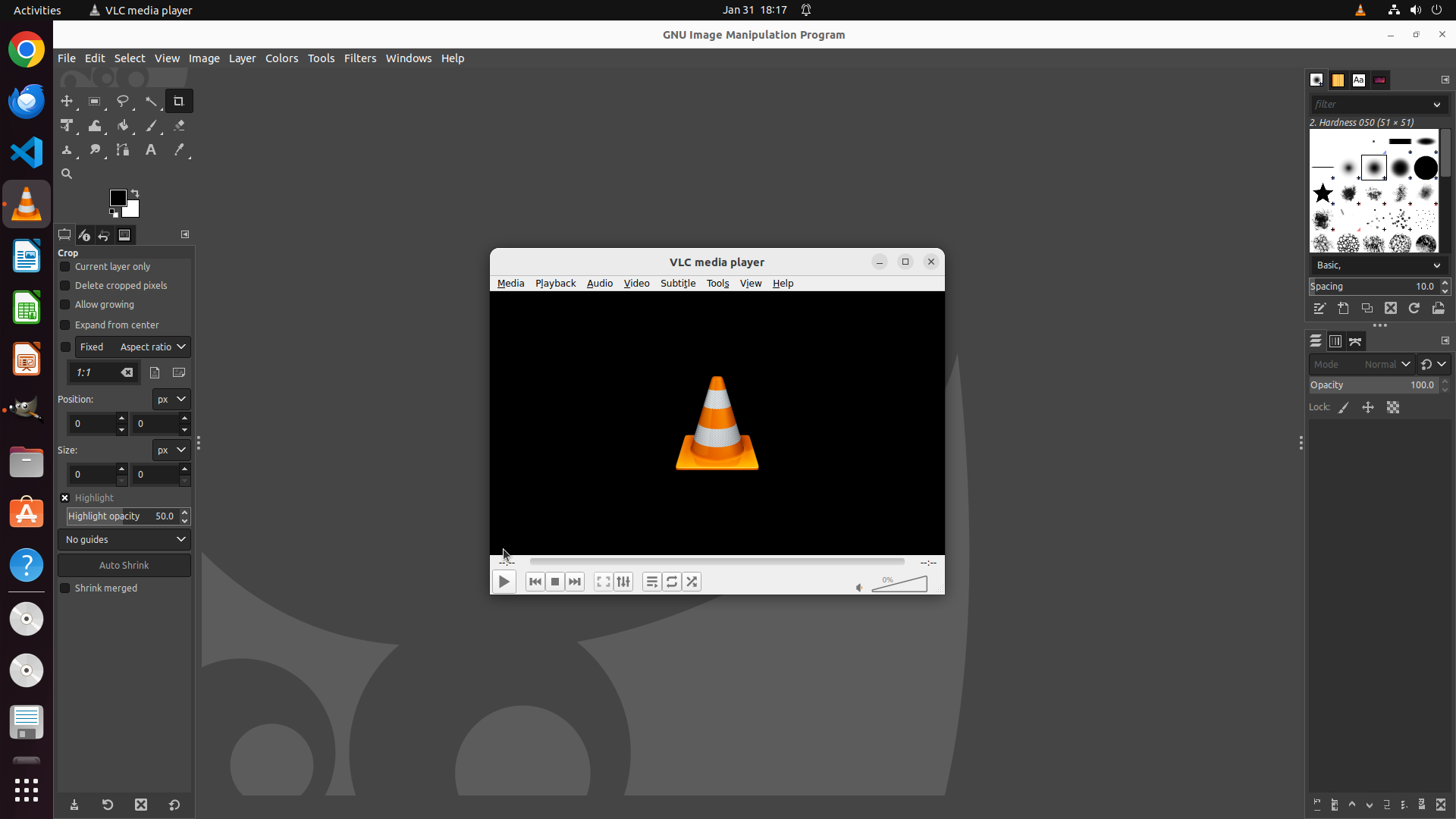} & \texttt{software knowledge to undergo sophisticated processes and ability to process multi-step procedure successfully} \\
\hline
Multiple \thead{\fontsize{7pt}{10pt}\selectfont (ThunderBird+\\ \fontsize{7pt}{10pt}\selectfont Writer+Chrome)} & \textit{Help me export charts, graph or other images from docx files received in email ``Lecture Document'' in Notes folder and upload these png files to the figures/ folder in Google Drive for later use (use numbers to name them).} & \includegraphics[width=6cm, height=3.38cm]{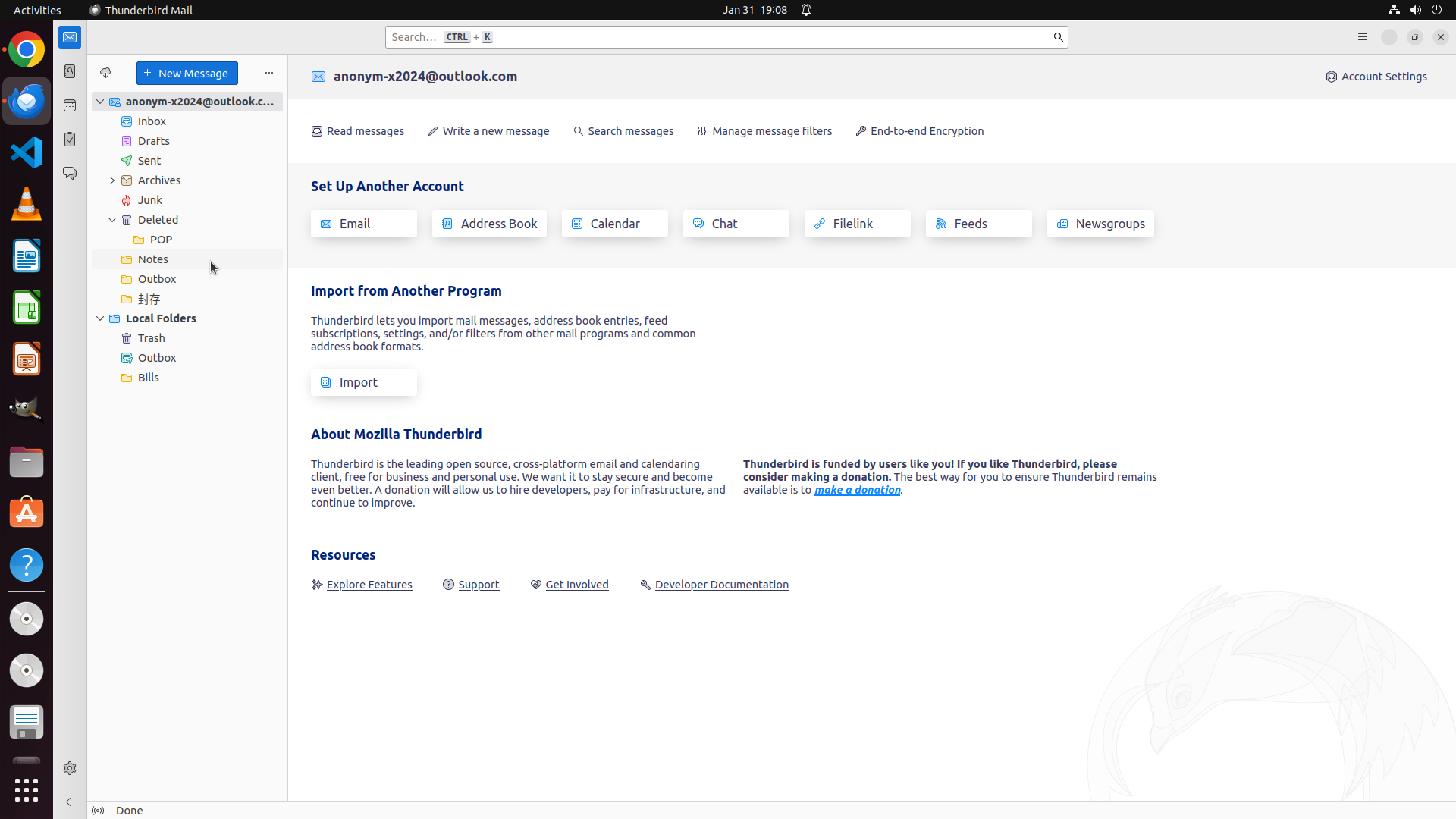} & \texttt{ability to selectively export charts, graphs and images from docx file; software knowledge for google drive file upload} \\
\hline
Multiple {\fontsize{7pt}{10pt}\selectfont (Chrome+Calc)} & \textit{Could you help me extract data in the table from a new invoice uploaded to my Google Drive, then export it to a Libreoffice calc .xlsx file in the desktop?} & \includegraphics[width=6cm, height=3.38cm]{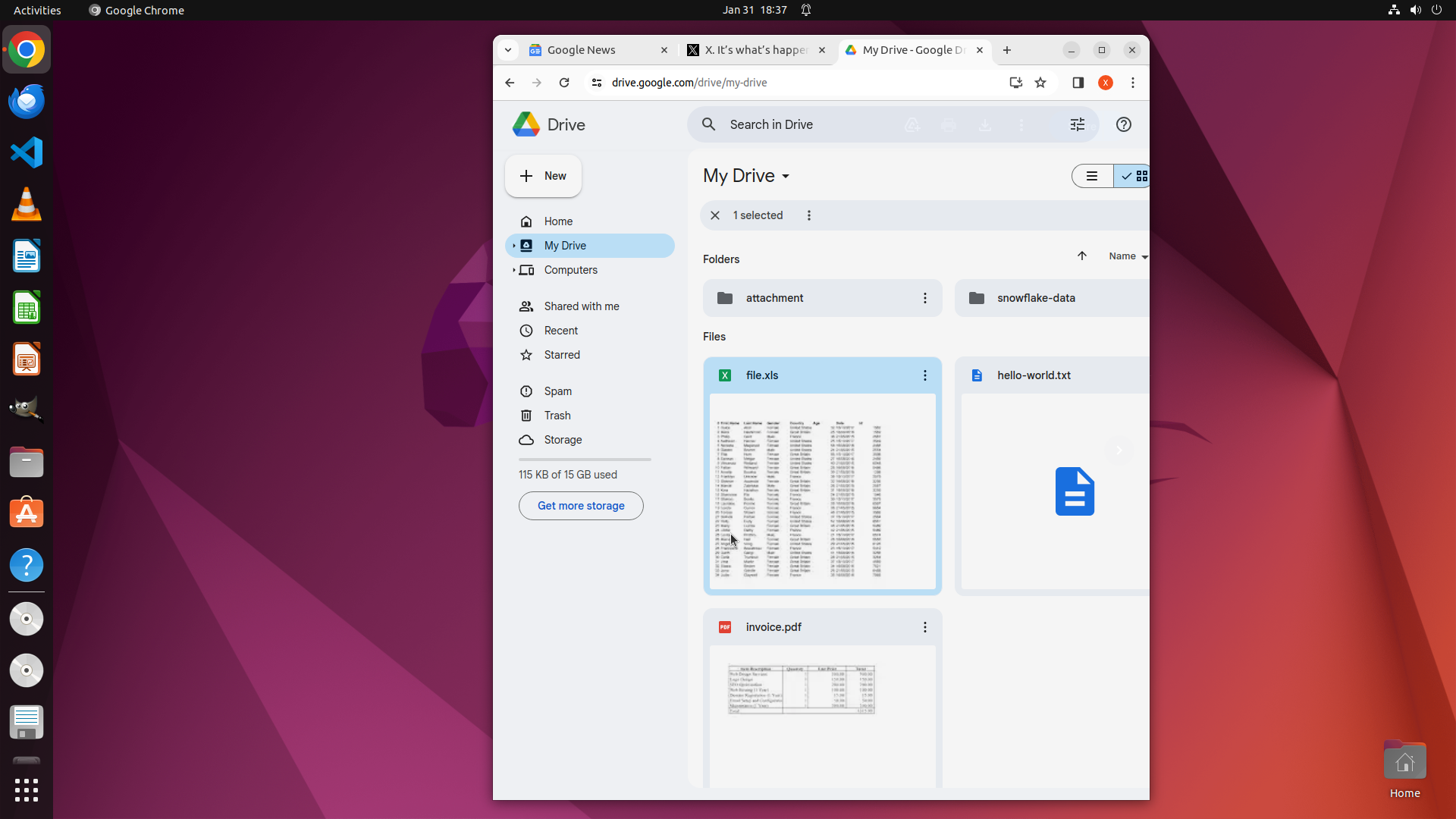} & \texttt{ability to do table data extraction; export data to .xlsx file} \\
\hline
\end{longtable}

\section{Details of Baseline Methods}
\subsection{Hyper-Parameter of the Baseline Agents}
\label{app:prompting_details}
We utilize the versions of \texttt{gpt-3.5-turbo-16k}, \texttt{gpt-4-0125-preview}, and \texttt{gpt-4-vision-preview}, respectively for GPT results, need to be noted that result could be changed from time since it is close-sourced.
We also employ the \texttt{gemini-pro} and \texttt{gemini-pro-vision} versions for the Gemini models 
For all language models, we set the temperature parameter to 1.0, and top\_p to 0.9, and the maximum number of tokens for generation is set to 1500. 
We set the maximum steps of interaction to 15 and the maximum time limits to 30 minutes for all tasks since the agent could lead to a stuck environment under some unexpected cases.

\subsection{Prompt Details}

\subsubsection{Prompt for A11y Tree, Screenshot and their Combination Setting}

\begin{tcolorbox}
\begin{Verbatim}[breaklines=true]
You are an agent which follow my instruction and perform desktop computer tasks as instructed.
You have good knowledge of computer and good internet connection and assume your code will run on a computer for controlling the mouse and keyboard.
For each step, you will get an observation of an image, which is the screenshot of the computer screen and you will predict the action of the computer based on the image.

You are required to use `pyautogui` to perform the action grounded to the observation, but DONOT use the `pyautogui.locateCenterOnScreen` function to locate the element you want to operate with since we have no image of the element you want to operate with. DONOT USE `pyautogui.screenshot()` to make screenshot.
Return one line or multiple lines of python code to perform the action each time, be time efficient. When predicting multiple lines of code, make some small sleep like `time.sleep(0.5);` interval so that the machine could take; Each time you need to predict a complete code, no variables or function can be shared from history
You need to to specify the coordinates of by yourself based on your observation of current observation, but you should be careful to ensure that the coordinates are correct.
You ONLY need to return the code inside a code block, like this:
```python
# your code here
```
Specially, it is also allowed to return the following special code:
When you think you have to wait for some time, return ```WAIT```;
When you think the task can not be done, return ```FAIL```, don't easily say ```FAIL```, try your best to do the task;
When you think the task is done, return ```DONE```.

My computer's password is 'password', feel free to use it when you need sudo rights.
First give the current screenshot and previous things we did a short reflection, then RETURN ME THE CODE OR SPECIAL CODE I ASKED FOR. NEVER EVER RETURN ME ANYTHING ELSE.
\end{Verbatim}
\end{tcolorbox}

For a11y tree setting and a11y tree + screenshot setting, the prompts are basically the same, just replace the screenshot words with a11y tree words.

\subsubsection{Prompt for SoM Setting}
\begin{tcolorbox}
\begin{Verbatim}[breaklines=true]
You are an agent which follow my instruction and perform desktop computer tasks as instructed.
You have good knowledge of computer and good internet connection and assume your code will run on a computer for controlling the mouse and keyboard.
For each step, you will get an observation of the desktop by 1) a screenshot with interact-able elements marked with numerical tags; and 2) accessibility tree, which is based on AT-SPI library. And you will predict the action of the computer based on the image and text information.

You are required to use `pyautogui` to perform the action grounded to the observation, but DONOT use the `pyautogui.locateCenterOnScreen` function to locate the element you want to operate with since we have no image of the element you want to operate with. DONOT USE `pyautogui.screenshot()` to make screenshot.
You can replace x, y in the code with the tag of the element you want to operate with. such as:
```python
pyautogui.moveTo(tag_3)
pyautogui.click(tag_2)
pyautogui.dragTo(tag_1, button='left')
```
When you think you can directly output precise x and y coordinates or there is no tag on which you want to interact, you can also use them directly. 
But you should be careful to ensure that the coordinates are correct.
Return one line or multiple lines of python code to perform the action each time, be time efficient. When predicting multiple lines of code, make some small sleep like `time.sleep(0.5);` interval so that the machine could take; Each time you need to predict a complete code, no variables or function can be shared from history
You need to to specify the coordinates of by yourself based on your observation of current observation, but you should be careful to ensure that the coordinates are correct.
You ONLY need to return the code inside a code block, like this:
```python
# your code here
```
Specially, it is also allowed to return the following special code:
When you think you have to wait for some time, return ```WAIT```;
When you think the task can not be done, return ```FAIL```, don't easily say ```FAIL```, try your best to do the task;
When you think the task is done, return ```DONE```.

My computer's password is 'password', feel free to use it when you need sudo rights.
First give the current screenshot and previous things we did a short reflection, then RETURN ME THE CODE OR SPECIAL CODE I ASKED FOR. NEVER EVER RETURN ME ANYTHING ELSE.
\end{Verbatim}
\end{tcolorbox}

\subsection{Accessibility Tree Filtering}
\label{app:a11y_tree_handling}
Since the original tree is large (usually over 1 million tokens in XML format), we 
filter the accessibility tree nodes by their tags, visibilities, availabilities,
\textit{etc}. The concrete rules are illustrated in 
the following Table~\ref{table:a11y_tree_handling}.

\begin{longtable}{p{0.3\linewidth}p{0.3\linewidth}p{0.3\linewidth}}
\caption{Criteria for keeping a11y tree nodes on Ubuntu and Windows platforms}
\label{table:a11y_tree_handling}\\
\toprule
\textbf{Condition} & \textbf{Ubuntu} & \textbf{Windows} \\
\midrule
\endfirsthead
\toprule
\textbf{Condition} & \textbf{Ubuntu} & \textbf{Windows} \\
\midrule
\endhead
\bottomrule
\endfoot
Node Tags & document$*$, $*$item, $*$button, $*$heading, $*$label, $*$scrollbar, $*$searchbox, $*$textbox, $*$link, $*$tabelement, $*$textfield, $*$textarea, $*$menu,
alert, canvas, check-box, combo-box, entry, icon, image, paragraph, scroll-bar, section, slider, static, table-cell, terminal, text, netuiribbontab, start, trayclockwclass, traydummysearchcontrol, uiimage, uiproperty, uiribboncommandbar & Same as Ubuntu \\
\midrule
Showing & True & Not Applicable \\
\midrule
Visible & True & True \\
\midrule
Enabled or Editable or Expandable or Checkable & True & True \\
\midrule
Has Name or Text or Image & True & True \\
\midrule
Position &  $>= 0$ &    $>= 0$ \\
\midrule
Size &  $> 0$ &    $> 0$ \\
\end{longtable}

\subsection{Set-of-Mark Implementation Details}
\label{app:som_details}
\begin{figure}[htbp]
    \centering
    \includegraphics[width=0.32\linewidth]{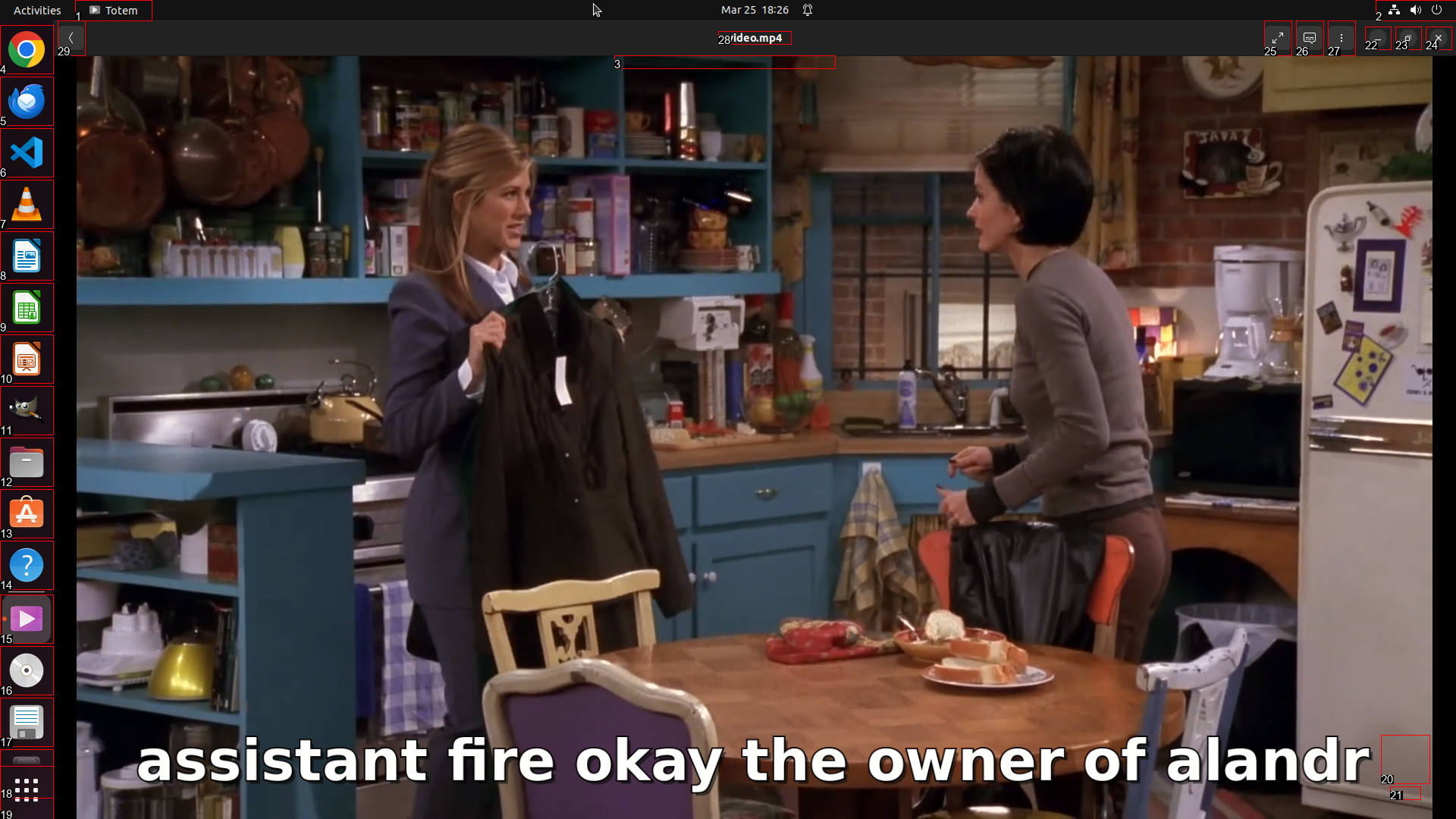}
    \includegraphics[width=0.32\linewidth]{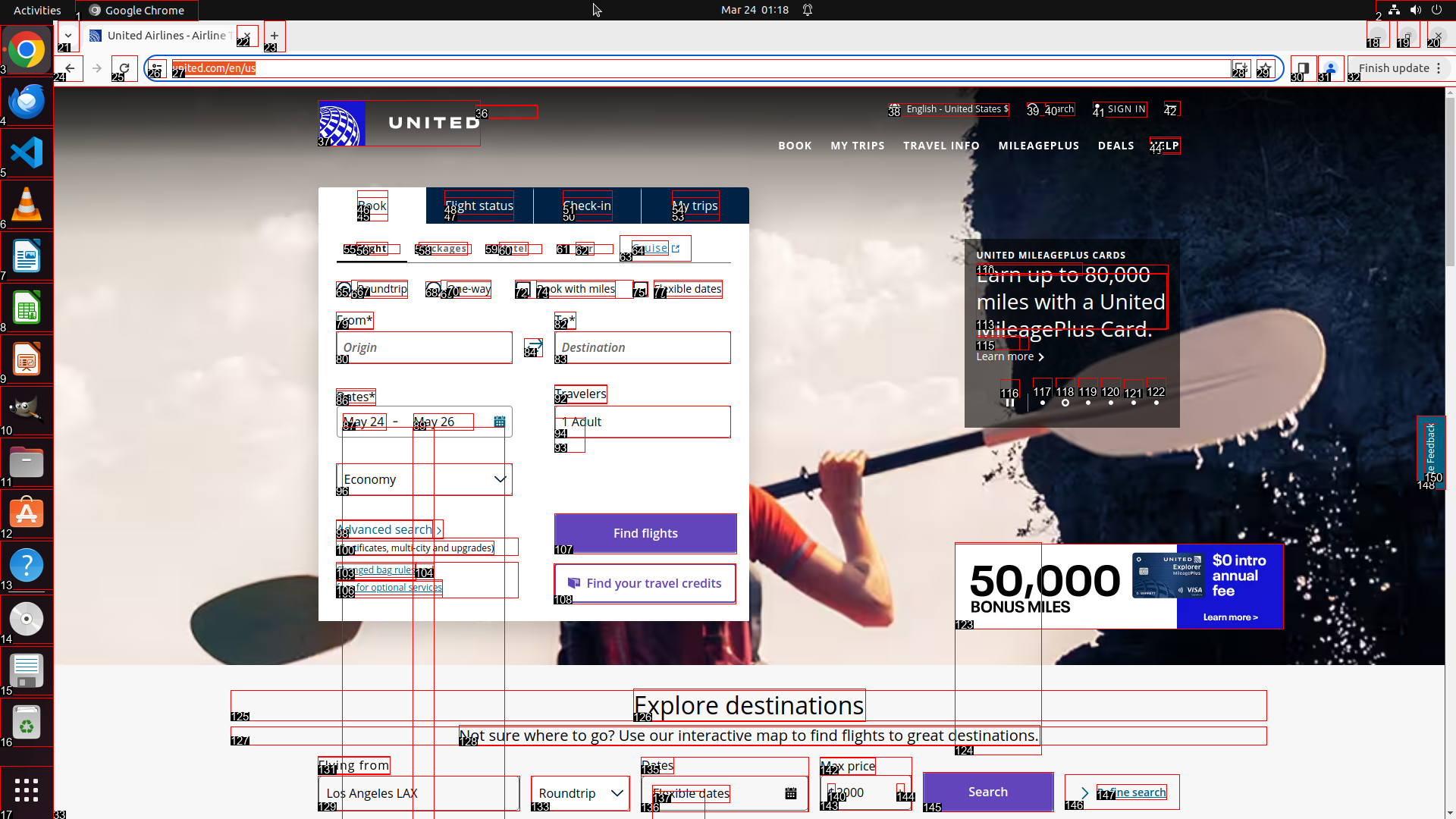}
    \includegraphics[width=0.32\linewidth]{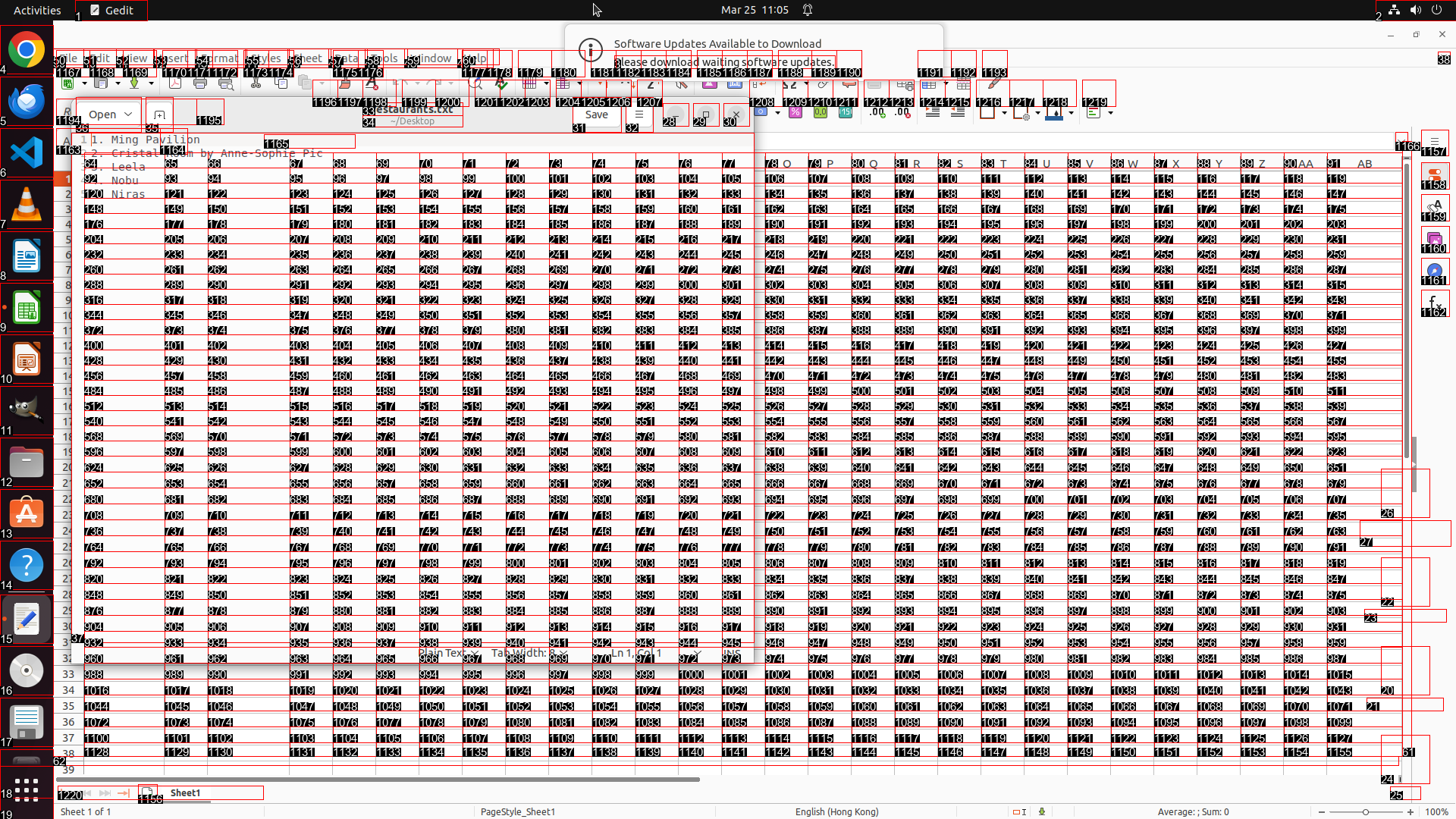}
    \includegraphics[width=0.32\linewidth]{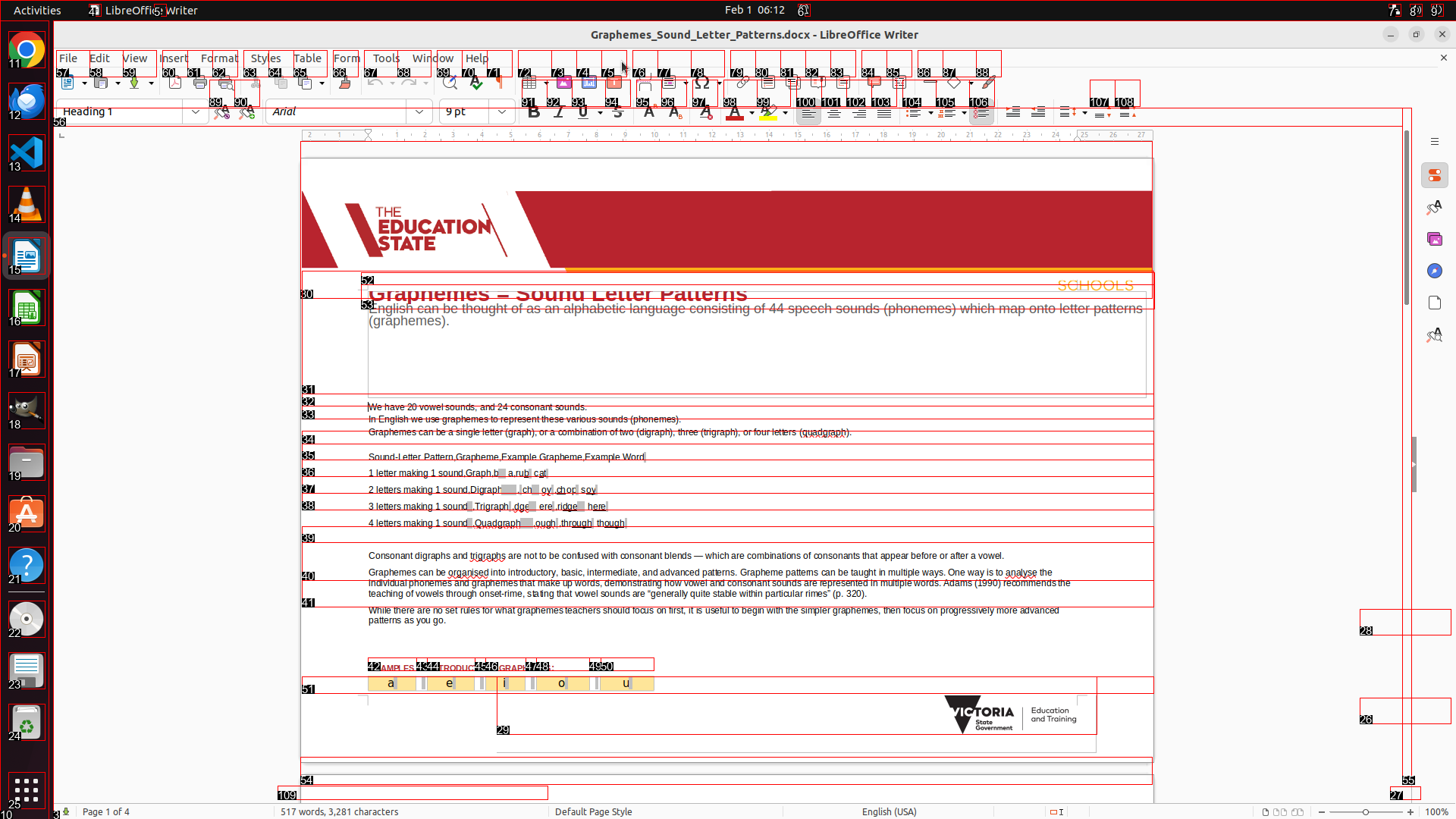}
    \includegraphics[width=0.32\linewidth]{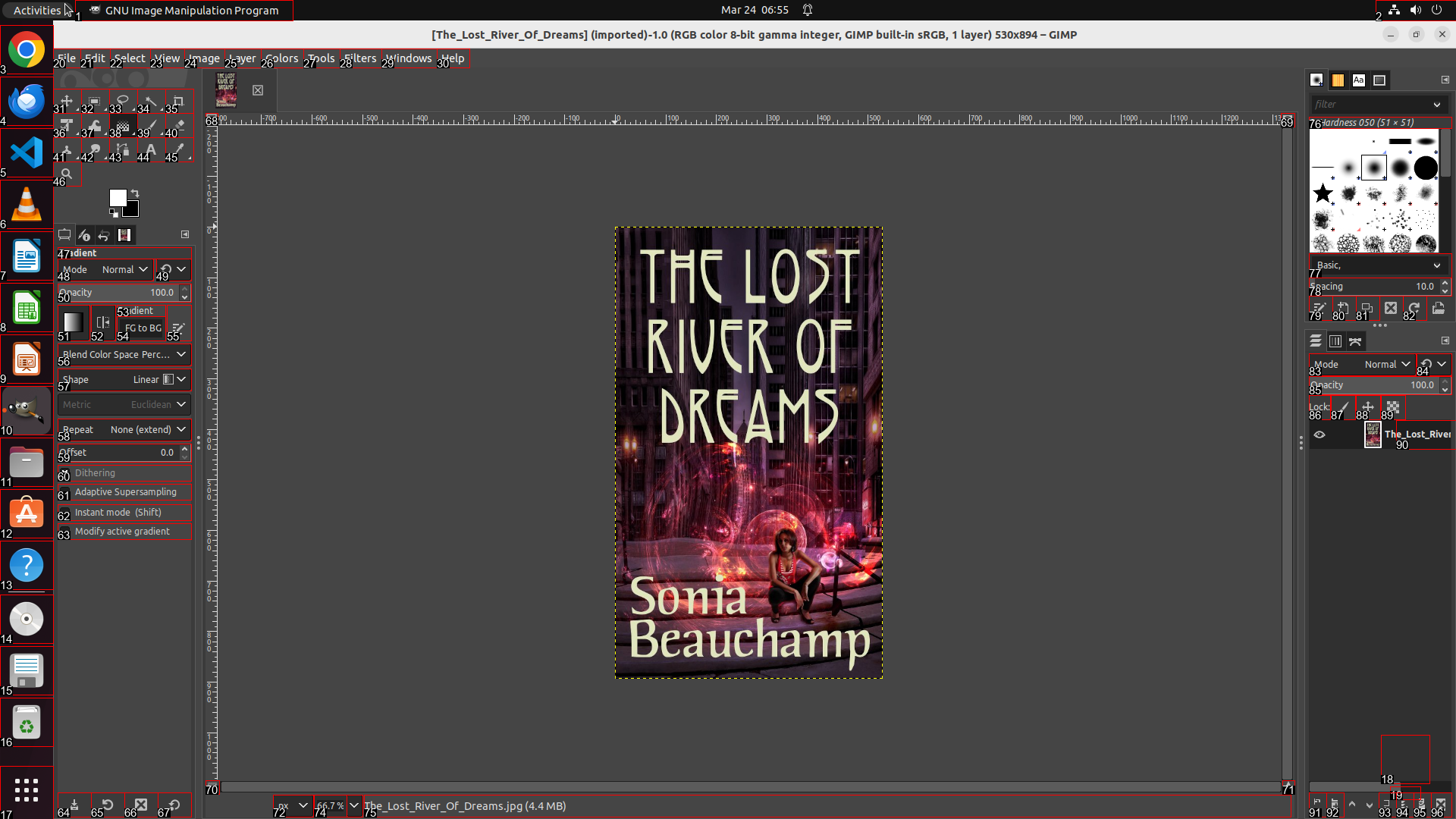}
    \includegraphics[width=0.32\linewidth]{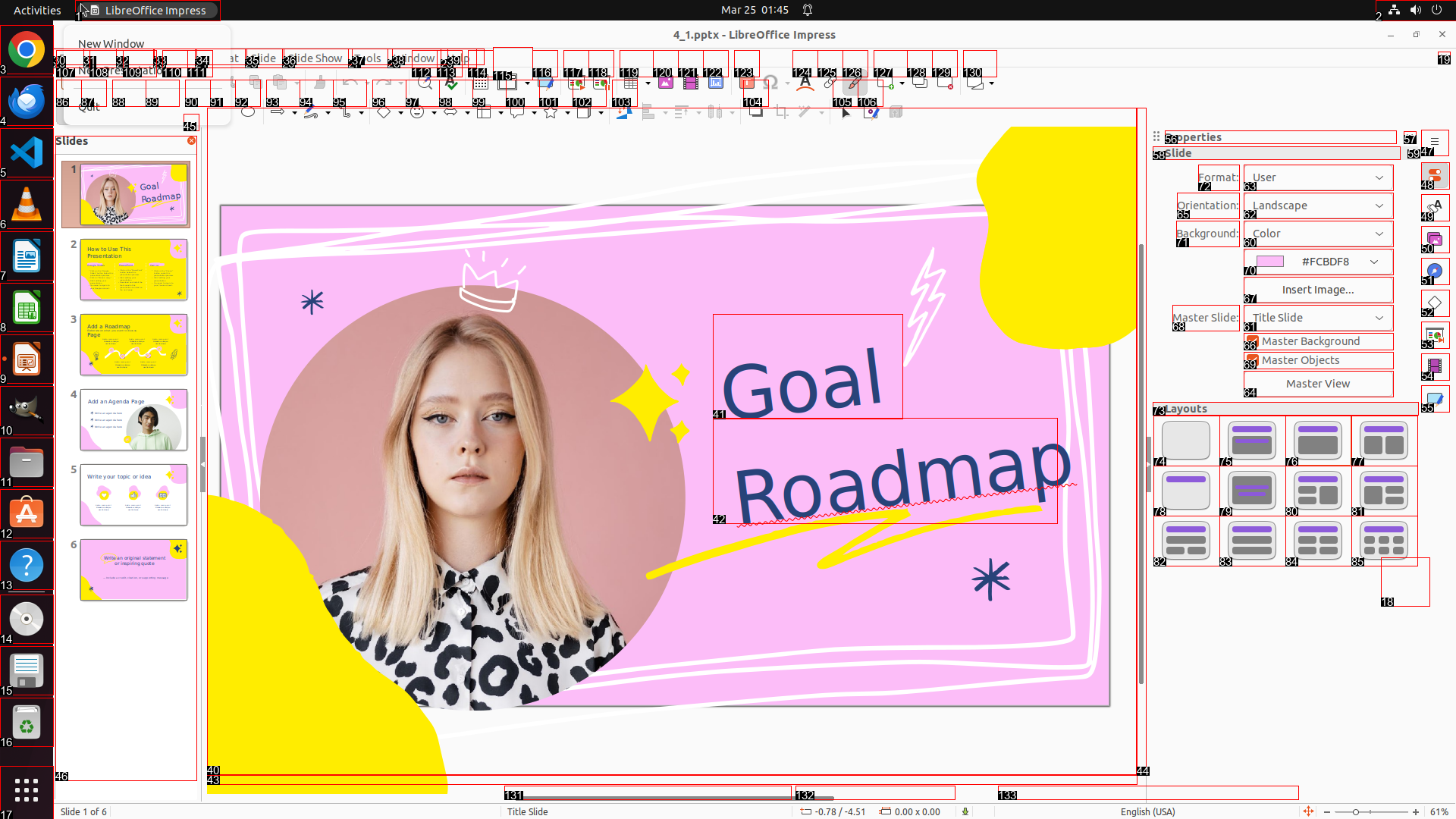}
    \caption{
    Showcase of example screenshots marked by SoM across different app GUI.
    }
    \label{fig:som_exampels}
\end{figure}

Our methodology involves an initial transformation of the original observational image acquired from our environment into the standardized \textit{SoM} format and be putted into VLMs together with a table of the marks with metadata information such as tags and names.
This format consists of bounding boxes that bound the sub-images of interest, each associated with a corresponding integer mark. 
Notably, our approach incorporates the utilization of the accessibility tree to identify the bounding boxes associated with all clickable buttons within the current image observation, instead of using segmentation models like the original \textit{SoM} paper.
Some examples of screenshots after applying SoM are shown in Figure~\ref{fig:som_exampels}.
We can observe the emergence of some low-quality, unmodelable tasks, and even misleading bounding boxes, depending on the level of support from the software ecosystem.
This could be another reason for the poor performance of SoM. Future improvements could be made in this regard.

\subsection{Full Results of Baseline Methods}
\label{app:full_results}
Here we show the break-down results of baseline methods from different LLMs and VLMs for follow-up reference.

\begin{figure}[htbp]
    \centering
    \includegraphics[width=0.8\linewidth]{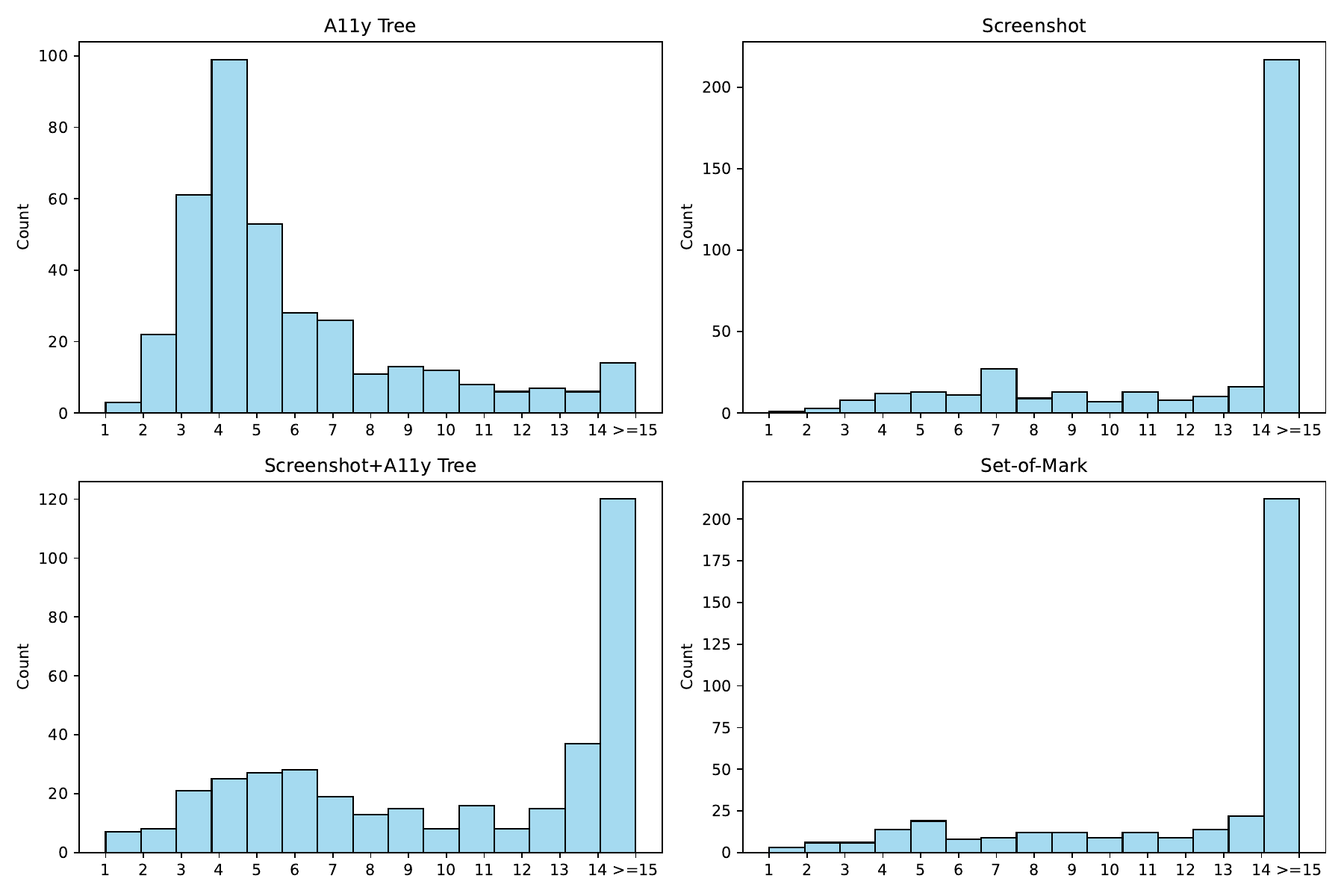}
    \caption{
    Distribution of steps taken by the GPT-4V based agents across four different settings.
    }
    \label{fig:steps_distribution}
\end{figure}

We have also compiled the distribution of steps taken by the GPT-4V model under our four settings: Accessibility Tree (A11y Tree), Screenshot, Screenshot combined with Accessibility Tree (Screenshot+A11y Tree), and Set-of-Mark. 
This data (as shown in Fig.~\ref{fig:steps_distribution}) provides potential clues for future work. 
Overall, there are observable differences in how many steps the agent chooses to execute and when it decides to terminate under different settings. 
More detailed control and analysis of these behaviors can be explored in subsequent research efforts.

\begin{table*}[ht]
\centering
\caption{
Detailed success rates of baseline LLM and VLM agents on \ours, divided by apps (domains): OS, LibreOffice Calc, LibreOffice Impress, LibreOffice Writer, Chrome, VLC Player, Thunderbird, VS Code, GIMP and Workflow which is comprehensive with multiple apps, for gaining insights from interfaces and operation logics.
}
\label{tab:exp_baseline_full}
\small
\scalebox{1}{
\renewcommand{\arraystretch}{1.1} 
\begin{tabularx}{\linewidth}{>{\centering\arraybackslash}p{1cm}>{\centering\arraybackslash}p{2cm}>{\centering\arraybackslash}X>{\centering\arraybackslash}X>{\centering\arraybackslash}X>{\centering\arraybackslash}X>{\centering\arraybackslash}X>{\centering\arraybackslash}X>{\centering\arraybackslash}X>{\centering\arraybackslash}X>{\centering\arraybackslash}X>{\centering\arraybackslash}X}
\toprule
\multirow{2}{*}{\textbf{Inputs}} & \multirow{2}{*}{\textbf{Model}} & \multicolumn{9}{c}{\textbf{Success Rate (↑)}} \\ 
\cline{3-12}
~ & ~ & OS & Calc &Impress & Writer & VLC & TB & Chrome & VSC & GIMP & Workflow \\
\midrule
A11y & Mixtral-8x7B  & 12.50 & 0.00 & 0.39 & 4.34  & 10.22 &6.67 & 2.17 & 8.69 & 3.85 & 0.10 \\
& GPT-3.5   & 4.17 & 2.13 & 6.77 & 4.35 & 6.53 & 0.00 & 2.17 & 0.00 & 0.00 & 1.62 \\
& Gemini-Pro   & 4.17 & 0.00 & 2.13 & 4.35 & 12.41 & 0.00 & 2.17 & 0.00 & 7.69 & 0.63 \\
& GPT-4   & 20.83 & 0.00 & 6.77 & 4.35 & 23.53 & 26.67 & 26.09 & 30.43 & 23.08 & 2.97 \\
& Gemini-Pro-1.5   & 12.50 & 2.13 & 2.13 & 4.35 & 6.53 & 0.00 & 10.87 & 8.70 & 0.00 & 3.60 \\
& Llama-3-70B & 4.17 & 0.00 & 0.39 & 8.70 & 6.53 & 0.00 & 2.17 & 0.00 & 0.00 & 0.63 \\
& GPT-4o   & 20.83 & 6.38 & 6.77 & 8.69 & 12.41 & 20.00 & 17.39 & 21.74 & 11.54 & 7.56 \\
& Qwen-Max   & 29.17 & 0.00 & 2.52 & 13.04 & 8.95 & 0.00 & 10.87 & 8.70 & 11.54 & 2.61 \\
\midrule
Screen & CogAgent    & 4.17 & 0.00 & 0.00 & 4.34 & 6.53 &0.00 & 2.17 & 0.00 & 0.00 & 0.00 \\
& Gemini-ProV  & 8.33 & 0.00 & 6.77 & 4.35 & 12.41 & 0.00 & 6.52 & 8.70 & 23.08 & 2.08 \\
& GPT-4V    & 12.50 & 0.00 & 2.52 & 4.35 & 18.34 & 0.00 & 6.52 & 0.00 & 7.69 & 6.04 \\
& Claude-3-Opus  & 4.17 & 0.00 & 2.52 & 4.34 & 6.53 &0.00 & 2.17 & 0.00 & 3.84 & 2.61 \\
& Gemini-Pro-1.5   & 12.50 & 0.00 & 13.16 & 8.70 & 6.53 & 0.00 & 2.17 & 0.00 & 11.54 & 3.60 \\
& GPT-4o   & 8.33 & 0.00 & 6.77 & 4.35 & 16.10 & 0.00 & 4.35 & 4.35 & 3.85 & 5.58 \\
\midrule
Screen & CogAgent  & 4.17 & 2.17 & 0.00 & 4.35 & 6.53 &0.00 & 2.17 & 0.00 & 0.00 & 0.10 \\
+ A11y & Gemini-ProV  & 4.17 & 2.13 & 6.77 & 4.35 & 18.30 & 0.00 & 4.35 & 0.00 & 0.00 & 1.52 \\
& GPT-4V  & 16.67 & 0.00 & 6.77 & 21.73 & 24.18 & 33.33 & 21.74 & 21.74 & 15.38 & 4.59 \\
& Claude-3-Opus  & 12.50 & 2.13 & 4.65 & 4.34 & 18.30 &0.00 & 2.17 & 8.69 & 7.69 & 0.99 \\
& Gemini-Pro-1.5   & 12.50 & 0.00 & 4.65 & 8.70 & 12.41 & 0.00 & 8.70 & 4.35 & 11.54 & 1.56 \\
& GPT-4o   & 41.67 & 4.26 & 6.81 & 8.70 & 9.50 & 6.67 & 15.22 & 30.43 & 0.00 & 7.46 \\
\midrule
SoM & CogAgent  & 4.17 & 2.17 & 0.00 & 4.34 & 6.53 & 0.00 & 2.17 & 0.00 & 0.00 & 0.00 \\
& Gemini-ProV  & 4.17 & 0.00 & 0.39 & 4.34 & 6.53 & 0.00 & 0.00 & 0.00 & 0.00 & 0.63 \\
& GPT-4V  & 8.33 & 8.51 & 6.38 & 13.04 & 10.66 & 33.33 & 23.91 & 17.39 & 11.54 & 6.57 \\
& Claude-3-Opus  & 12.50 & 2.13 & 0.39 & 8.70 & 6.53 &13.33 & 17.39 & 0.00 & 11.54 & 4.49 \\
& Gemini-Pro-1.5   & 16.67 & 0.00 & 10.64 & 4.35 & 24.18 & 13.33 & 8.70 & 4.35 & 15.38 & 3.60 \\
& GPT-4o   & 20.83 &0.00 & 6.77 & 4.35 & 6.53 & 0.00 & 4.35 & 4.35 & 0.00 & 3.60 \\
\midrule
\rowcolor{gray!20}
\multicolumn{2}{c}{Human Performance}  & 75.00 & 61.70 & 80.85 & 73.91 & 70.59 & 46.67 & 78.26 & 73.91 & 73.08 & 73.27 \\
\bottomrule
\end{tabularx}
}
\end{table*}

\section{Examples of Qualitative Analysis}
\label{appendix:qualitative_analysis}

Here we give the specific code execution process for the examples in the qualitative analysis in \S\ref{qualitative_analysis} and add more error cause analysis, trying to get insights for further improvement.

\subsection{Success and Failure Cases}

\begin{tcolorbox}[title={Success Task: \textit{I downloaded an episode of Friends to practice listening, but I don't know how to remove the subtitles. Please help me remove the subtitles from the video and export it as "subtitles.srt" and store it in the same directory as the video.}}]

\small
\begin{Verbatim}[commandchars=\\\{\}]
\textcolor{purple}{# Action 1}
import pyautogui
import time
\textcolor{blue}{# Move the mouse to the top bar of the media player window to access the menu}
pyautogui.moveTo(800, 10)  
\textcolor{blue}{# Assuming the menu is at the top of the screen}
pyautogui.click()
time.sleep(0.5)

\textcolor{purple}{# Action 2}
\textcolor{blue}{# Move the mouse to the Activities corner to open the GNOME Activities Overview}
pyautogui.moveTo(10, 10)  
\textcolor{blue}{# Assuming the Activities corner is at the top-left}
pyautogui.click()
\textcolor{blue}{# Type `terminal' to search for the Terminal application}
pyautogui.write(`terminal')
time.sleep(0.5)
\textcolor{blue}{# Press Enter to open the Terminal application}
pyautogui.press(`enter')

\textcolor{purple}{# Action 3}
\textcolor{blue}{# Focus the Terminal window and type the command to extract subtitles}
pyautogui.click(640, 360)  
\textcolor{blue}{# Clicking in the center of the screen to focus the Terminal}
time.sleep(0.5)
pyautogui.write(`ffmpeg -i video.mp4 -map 0:s:0 subtitles.srt')
pyautogui.press(`enter')

\textcolor{purple}{# Action 4}
\textcolor{blue}{# Focus the Terminal window and type the command}
\textcolor{blue}{ to create a new video without subtitles}
pyautogui.click(1100, 350)  
\textcolor{blue}{# Clicking inside the Terminal window to focus it}
pyautogui.write(`ffmpeg -i video.mp4 -c copy -sn no_subtitles_video.mp4')

\textcolor{purple}{Done!}
\end{Verbatim}
\end{tcolorbox}

\begin{tcolorbox}[title={Failure Task: \textit{Help me center align the heading in LibreOffice.}}]

\small
\begin{Verbatim}[commandchars=\\\{\}]
\textcolor{purple}{# Action 1}
import pyautogui
import time
\textcolor{blue}{# Move the mouse to the LibreOffice Writer document icon and double-click}
pyautogui.moveTo(1100, 720, duration=1)
pyautogui.doubleClick()

\textcolor{purple}{# Action 2}
\textcolor{blue}{# Move the mouse to the beginning of the heading text}
\textcolor{blue}{and click and drag to select the heading}
pyautogui.moveTo(200, 230, duration=1)
pyautogui.mouseDown()
pyautogui.moveTo(800, 230, duration=1)
pyautogui.mouseUp()
\textcolor{blue}{# Move the mouse to the center align button in the toolbar and click}
pyautogui.moveTo(400, 150, duration=1)
pyautogui.click()

\textcolor{purple}{# Action 3}
\textcolor{blue}{# Assuming the text is still selected, }
\textcolor{blue}{move the mouse to the center align button in the toolbar and click}
pyautogui.moveTo(400, 150, duration=1)
pyautogui.click()

\textcolor{purple}{# Action 4}
\textcolor{blue}{# Move the mouse to the center align button in the toolbar and click}
pyautogui.moveTo(340, 80, duration=1)
pyautogui.click()

\textcolor{purple}{Failed.}
\end{Verbatim}
\end{tcolorbox}

\begin{figure}[htbp]
    \centering
    \includegraphics[width=\linewidth]{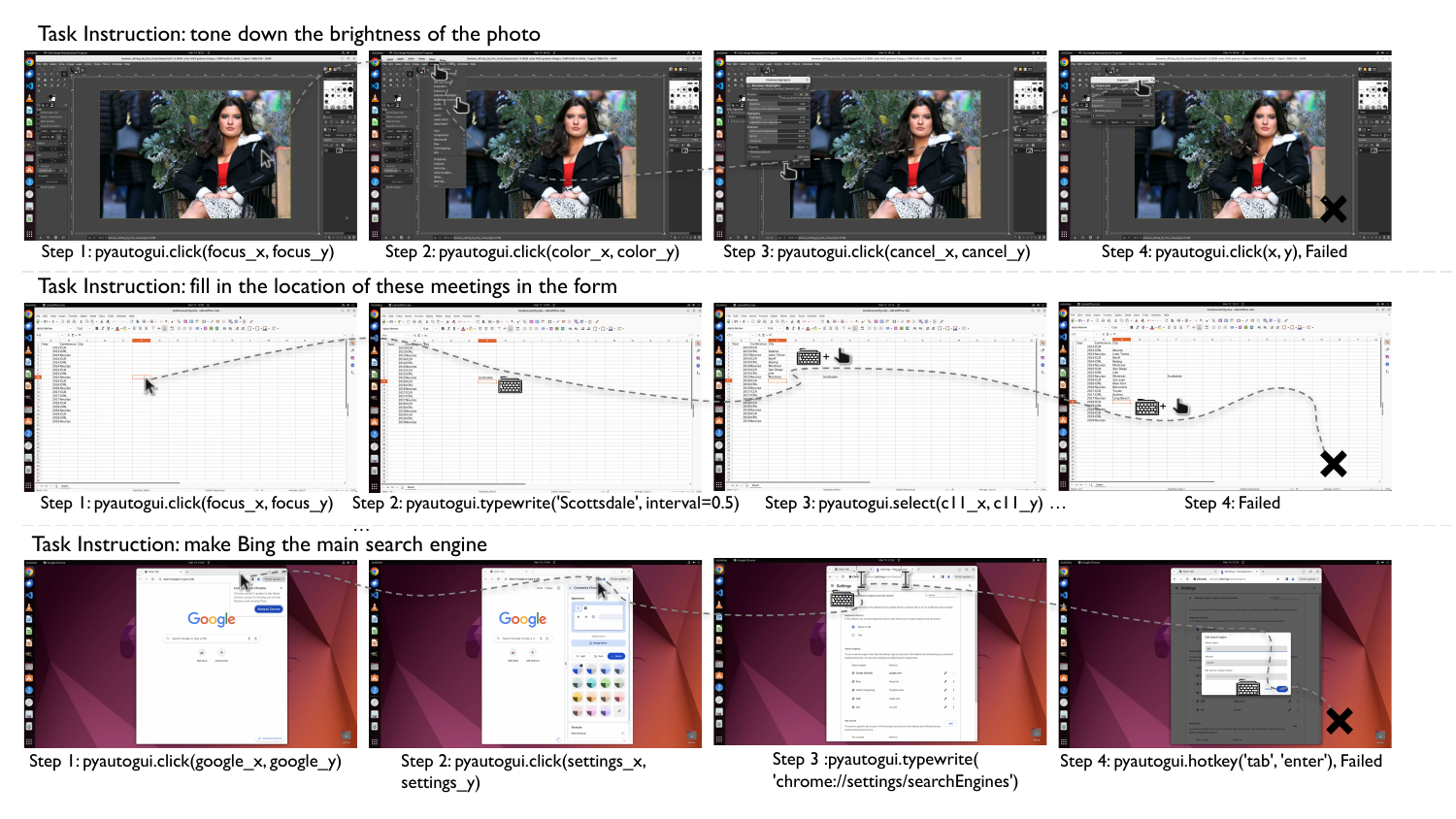}
    \caption{
    The figure shows examples of usage errors in three types of software. The first row is for GIMP, the second row is for LibreOffice Calc, and the third row is for Google Chrome. Detailed code execution processes are provided below.
    }
    \label{fig:model_lack_knowledge}
\end{figure}

\begin{tcolorbox}[title={GIMP Task: \textit{Could you tone down the brightness of my photo?}}]

\begin{Verbatim}[commandchars=\\\{\}]
\textcolor{purple}{# Action 1}
import pyautogui
import time
\textcolor{blue}{# Coordinates for the image icon on the desktop}
\textcolor{blue}{# Adding half of the width and height to get the center}
image_icon_x, image_icon_y = 1793 + 60, 920 + 17  
\textcolor{blue}{# Double-click on the image icon}
pyautogui.doubleClick(image_icon_x, image_icon_y)

\textcolor{purple}{# Action 2}
\textcolor{blue}{# Coordinates for the `Colors' menu}
\textcolor{blue}{# Adding half of the width and height to get the center}
colors_menu_x, colors_menu_y = 344 + 28, 64 + 12  
\textcolor{blue}{# Click on the `Colors' menu}
pyautogui.click(colors_menu_x, colors_menu_y)

\textcolor{purple}{# Action 3}
\textcolor{blue}{# Coordinates for the `Brightness-Contrast...' option in the `Colors' dropdown}
\textcolor{blue}{# Assuming each menu item is approximately 28 pixels apart}
brightness_contrast_option_x, brightness_contrast_option_y 
= 344 + 28, 64 + 12 + 28 * 6  
\textcolor{blue}{# Click on the `Brightness-Contrast...' option}
pyautogui.click(brightness_contrast_option_x, brightness_contrast_option_y)

\textcolor{purple}{# Action 4}
\textcolor{blue}{# Coordinates for the `Cancel' button in the `Shadows-Highlights' dialog}
\textcolor{blue}{# Adding half the width and height to get the center}
cancel_button_x, cancel_button_y = 375 + 42, 625 + 16  
\textcolor{blue}{# Click the `Cancel' button}
pyautogui.click(cancel_button_x, cancel_button_y)
\textcolor{blue}{# Coordinates for the `Brightness-Contrast...' option in the `Colors' dropdown}
\textcolor{blue}{# Adjusted index for Brightness-Contrast option}
brightness_contrast_option_x, brightness_contrast_option_y 
= 344 + 28, 64 + 12 + 28 * 5  
pyautogui.click(brightness_contrast_option_x, brightness_contrast_option_y)

\textcolor{purple}{# Action 5}
\textcolor{blue}{# Coordinates for the `Cancel' button in the `Exposure' dialog}
\textcolor{blue}{# Adding half the width and height to get the center}
cancel_button_x, cancel_button_y 
= 375 + 42, 391 + 16  
pyautogui.click(cancel_button_x, cancel_button_y)
\textcolor{blue}{# Coordinates for the `Brightness-Contrast...' option in the `Colors' dropdown}
\textcolor{blue}{# Adjusted index for Brightness-Contrast option}
brightness_contrast_option_x, brightness_contrast_option_y 
= 344 + 28, 64 + 12 + 28 $*$ 6  
pyautogui.click(brightness_contrast_option_x, brightness_contrast_option_y)

\textcolor{purple}{# Action 6}
\textcolor{blue}{# Coordinates for the `Cancel' button in the `Color Balance' dialog}
\textcolor{blue}{# Adding half the width and height to get the center}
cancel_button_x, cancel_button_y 
= 375 + 42, 588 + 16  
pyautogui.click(cancel_button_x, cancel_button_y)
\textcolor{blue}{# Since the `Brightness-Contrast...' option was not correctly opened before,}
\textcolor{blue}{we will use an estimated position based on the dropdown menu.}
\textcolor{blue}{# Assuming `Brightness-Contrast...' is the second option in the `Colors' menu.}
brightness_contrast_option_x, brightness_contrast_option_y 
= colors_menu_x, colors_menu_y + 28 * 2
pyautogui.click(brightness_contrast_option_x, brightness_contrast_option_y)

\textcolor{purple}{# Action 7}
\textcolor{blue}{# Coordinates for the `Cancel' button in the `Color Temperature' dialog}
\textcolor{blue}{# Adding half the width and height to get the center}
cancel_button_x, cancel_button_y = 375 + 42, 403 + 16  
pyautogui.click(cancel_button_x, cancel_button_y)
\textcolor{blue}{# Since the `Brightness-Contrast...' option was not correctly opened before, }
\textcolor{blue}{we will use an estimated position based on the dropdown menu.}
\textcolor{blue}{# Assuming `Brightness-Contrast...' is the second option in the `Colors' menu.}
brightness_contrast_option_x, brightness_contrast_option_y 
= colors_menu_x, colors_menu_y + 28 * 2
\textcolor{blue}{# Click on the `Brightness-Contrast...' option}
pyautogui.click(brightness_contrast_option_x, brightness_contrast_option_y)

\textcolor{blue}{# ... More Wrong Duplicate Actions; Can not find the right coordination.}

\textcolor{purple}{# Failed}
\end{Verbatim}
\end{tcolorbox}

\begin{tcolorbox}[title={Libreoffice calc Task: \textit{I now want to count the meeting cities of the three machine learning conferences in the past ten years from 2013 to 2019(including 2013 and 2019). I have listed the names and years of the conferences in excel. Please fill in the vacant locations.}}]

\begin{Verbatim}[commandchars=\\\{\}]
\textcolor{purple}{# Action 1}
\textcolor{blue}{# Move to cell C2 which is the first empty cell under `City' for ICLR 2013}
pyautogui.click(319, 222)  
\textcolor{blue}{# Coordinates of cell C2}
time.sleep(0.5)
\textcolor{blue}{# Type the city name for ICLR 2013}
pyautogui.write(`Scottsdale')
\textcolor{blue}{# Press Enter to move to the next cell}
pyautogui.press('enter')

\textcolor{purple}{# Action 2}
\textcolor{blue}{# Move to cell C3 which is the city cell for ICML 2013}
pyautogui.click(319, 268)  
\textcolor{blue}{# Coordinates of cell C3}
\textcolor{blue}{# Type the city name for ICML 2013}
pyautogui.write(`Atlanta')
\textcolor{blue}{# Press Enter to move to the next cell}
pyautogui.press('enter')

\textcolor{purple}{# Action 3}
\textcolor{blue}{# Move to cell C4 which is the city cell for NeurIPS 2013}
pyautogui.click(319, 289)  
\textcolor{blue}{# Coordinates of cell C4}
\textcolor{blue}{# Type the city name for NeurIPS 2013}
pyautogui.write(`Lake Tahoe')
\textcolor{blue}{# Press Enter to move to the next cell}
pyautogui.press('enter')"

\textcolor{blue}{... More copy-and-paste operations}

\textcolor{purple}{# Action 15}
\textcolor{blue}{# Move to cell C16 which is the city cell for NeurIPS 2017}
pyautogui.click(319, 541)  
\textcolor{blue}{# Coordinates of cell C16}
\textcolor{blue}{# Type the city name for NeurIPS 2017}
pyautogui.write(`Long Beach')
\textcolor{blue}{# Press Enter to move to the next cell}
pyautogui.press('enter')

\textcolor{purple}{Failed.}
\end{Verbatim}
\end{tcolorbox}

\begin{tcolorbox}[title={Chrome Task: \textit{Can you make Bing the main search thingy when I look stuff up on the internet?}}]

\begin{Verbatim}[commandchars=\\\{\}]
\textcolor{purple}{# Action 1}
pyautogui.click(x=1280, y=50)  
\textcolor{blue}{# Approximate coordinates for the `More' button}
pyautogui.move(0, 150)  
\textcolor{blue}{# Moving the cursor down to `Settings' in the drop-down menu}
pyautogui.click()  
\textcolor{blue}{# Clicking on `Settings'}

\textcolor{purple}{# Action 2}
pyautogui.click(tag_31)  
\textcolor{blue}{# Click on the `Finish update' button to}
\textcolor{blue}{approximate the location of `More' button}
pyautogui.move(-50, 0)  
\textcolor{blue}{# Move left from the `Finish update' button to approximate the `More' button}
pyautogui.click()  
\textcolor{blue}{# Click on the `More' button}
time.sleep(0.5)
pyautogui.move(0, 150)  
\textcolor{blue}{# Move down to the `Settings' option in the drop-down menu}
pyautogui.click()  
\textcolor{blue}{# Click on `Settings'}

\textcolor{purple}{# Action 3}
\textcolor{blue}{# Click on the `Customise Chrome' button which should be the `More' button}
time.sleep(1)  
\textcolor{blue}{# Wait for the dropdown menu to appear}
\textcolor{blue}{# The `Settings' option is usually located towards the bottom of}
\textcolor{blue}{the dropdown menu, so I'll move the cursor down to reach it}
pyautogui.move(0, 200)  
\textcolor{blue}{# Move down to the `Settings' option in the drop-down menu}
pyautogui.click()  
\textcolor{blue}{# Click on `Settings'}

\textcolor{purple}{# Action 4}
pyautogui.click(tag_47)  
\textcolor{blue}{# Click on the `Close' button to close the `Customise Chrome' menu}
pyautogui.click(tag_25)  
\textcolor{blue}{# Click on the `New Tab' button to open a new tab}
time.sleep(1)  
\textcolor{blue}{# Wait for the new tab to open}
pyautogui.typewrite(`chrome://settings/searchEngines')  
\textcolor{blue}{# Type the direct URL to the search engine settings page}
pyautogui.press(`enter')

\textcolor{purple}{# Action 5}
\textcolor{blue}{# Simulate pressing the `Tab' key multiple times to navigate through the options}
for _ in range(10):  
\textcolor{blue}{# Assuming it might take around 10 tabs to reach the search engine list}    
    pyautogui.press(`tab')    
    time.sleep(0.5)  
\textcolor{blue}{# Wait for the focus to move to the next element}
\textcolor{blue}{# Simulate pressing the `Enter' key to select an option,}
\textcolor{blue}{which would be done once we reach the Bing option}
pyautogui.press('enter')

\textcolor{purple}{# Action 6}
pyautogui.click(tag_42)  
\textcolor{blue}{# Click on the `Cancel' button to exit the edit mode for the '360' search engine}

\textcolor{purple}{Failed.}
\end{Verbatim}
\end{tcolorbox}

\newpage

\subsection{Common Errors by GPT-4V Agents }
\label{appendix:common_errors}

\begin{figure}[htbp]
    \centering
    \includegraphics[width=0.8\linewidth]{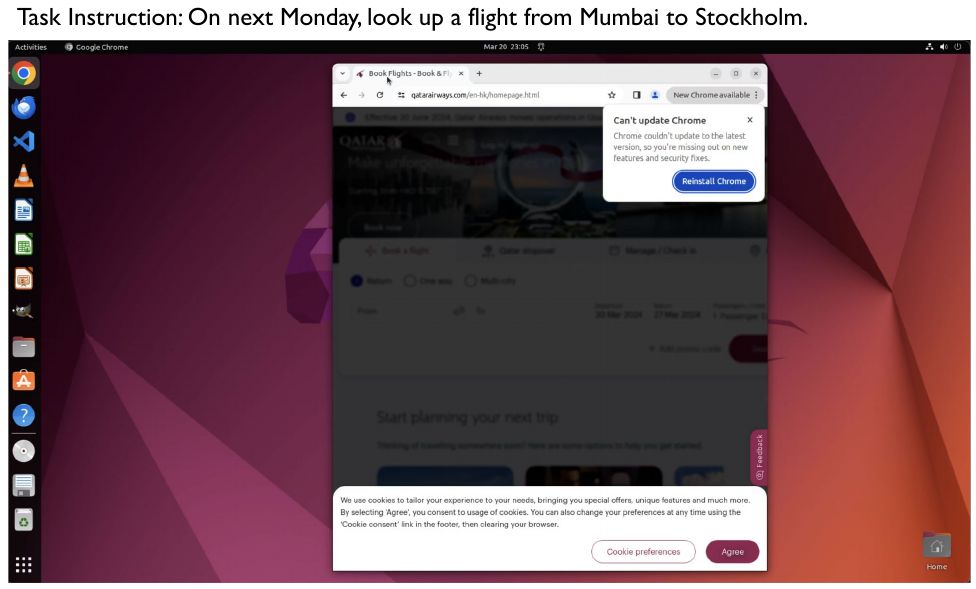}
    \caption{
    The error shown in the figure is due to mouse click inaccuracies. The agent was supposed to click on the product category images of the Nike official store and follow the instructions to search for women's jerseys priced over 60 dollars. However, due to a coordinate calculation error, it mistakenly clicked on the favorite button, creating a significant discrepancy between the current state and the target state. Under these circumstances, the agent is unable to backtrack to the previous state and start over.
    }
    \label{fig:wrong_click}
\end{figure}

\begin{figure}[htbp]
    \centering
    \includegraphics[width=0.8\linewidth]{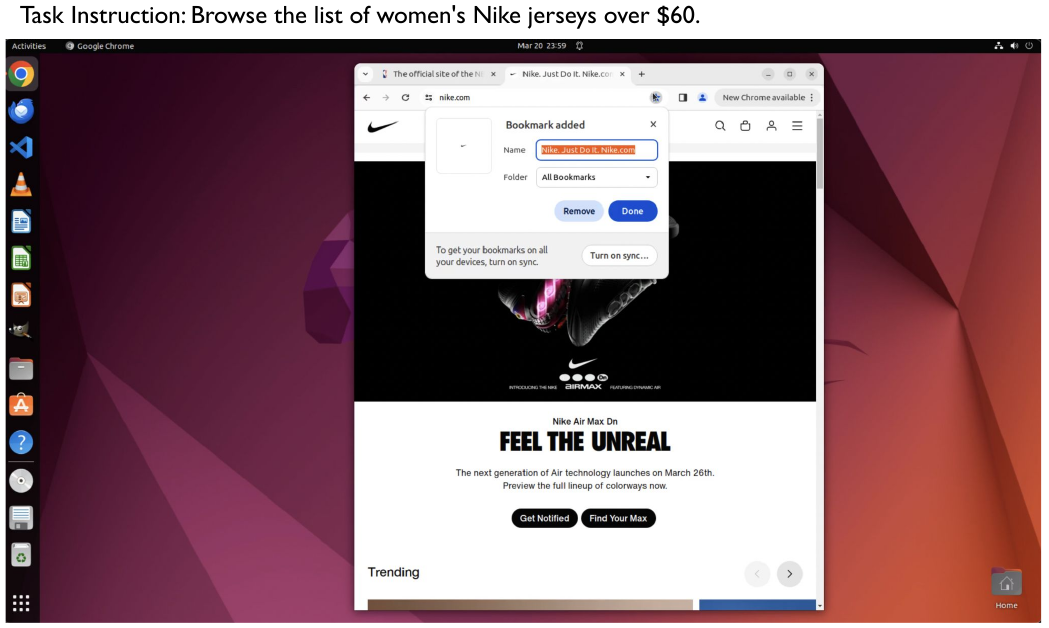}
    \caption{
    The error shown in the figure is due to inadequate handling of environmental noise. When the agent encounters web pages with cookies or pop-up ads, its unfamiliarity with the structure of such pages leads to attempts to interact directly with the web page without bypassing the cookies or pop-ups. Since the pop-ups or cookie requests are not closed, all actions are ineffective.
    }
    \label{fig:cookie_block}
\end{figure}

In the following two figures ( Figure \ref{fig:cookie_block} and Figure \ref{fig:wrong_click}), we take the web page tasks as examples to show the two most common types of errors in GPT4-V:
\textit{mouse click inaccuracies} and \textit{inadequate handling of environmental noise}.

\newpage

\subsection{Discrepancies in Task Difficulty between Agent and Human}
\label{appendix:human_vs_agent}

\begin{figure}[htbp]
    \centering
    \includegraphics[width=\linewidth]{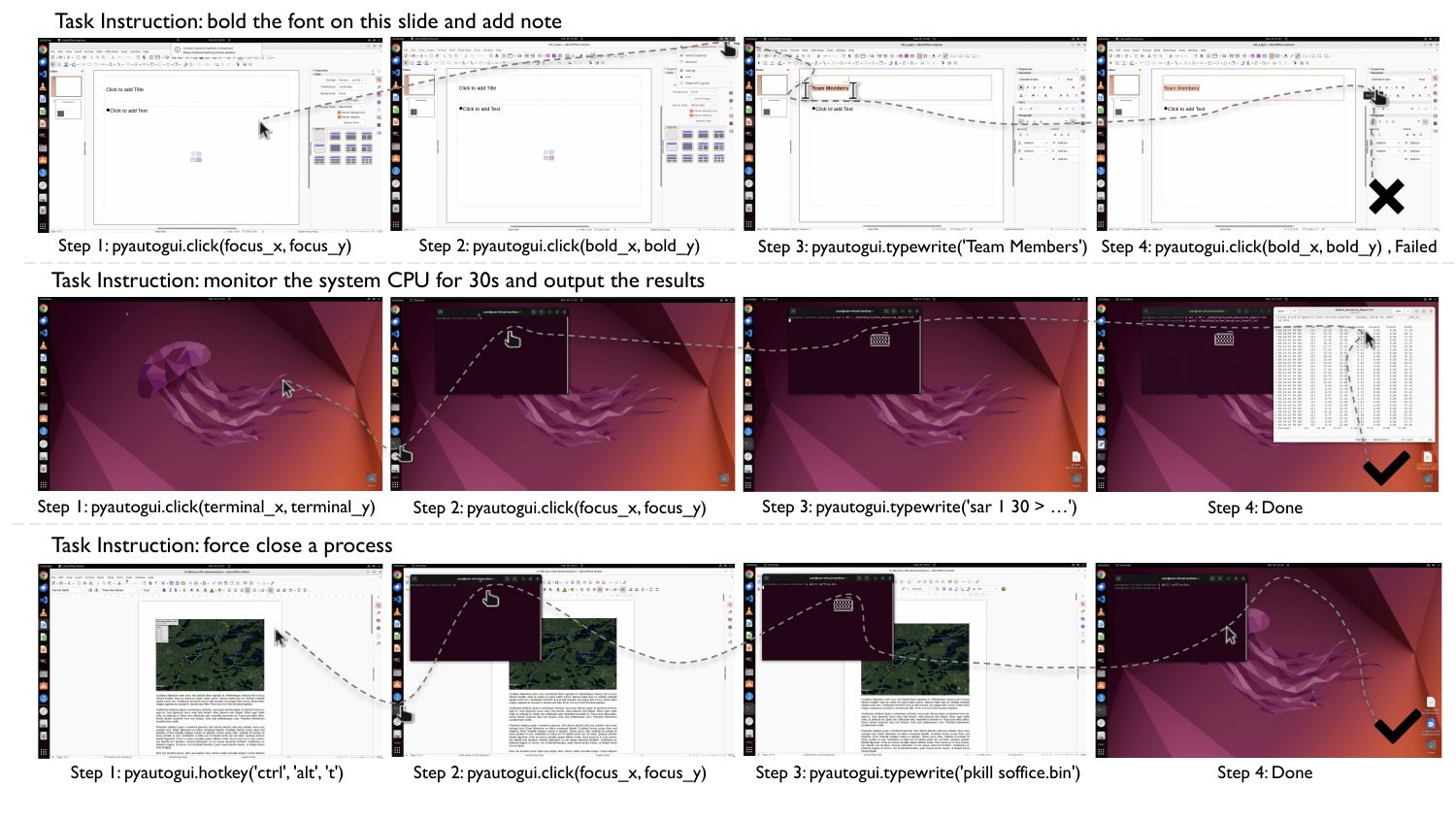}
    \caption{
    Supplementary examples of tasks performed by humans and agents.
    }
    \label{fig:human_agent_supplement}
\end{figure}

\begin{tcolorbox}[title={Human > Agent Task: \textit{erase all the highlighted marks in this document}}]

\small
\begin{Verbatim}[commandchars=\\\{\}]
\textcolor{purple}{# Action 1}
import pyautogui
import time
\textcolor{blue}{# Move to the `Character Highlighting Color' button and click}
pyautogui.moveTo(1020, 122)
pyautogui.click()

\textcolor{purple}{# Action 2}
\textcolor{blue}{# Assuming the `No Color' option is the first option in the dropdown}
\textcolor{blue}{# Move the mouse slightly below the `Character Highlighting Color' button}
\textcolor{blue}{to click on `No Color'}
pyautogui.moveTo(1020, 180)
time.sleep(0.5)
pyautogui.click()

\textcolor{purple}{# Action 3}
\textcolor{blue}{# Move to the `Save' button and click}
pyautogui.moveTo(184, 108) 
time.sleep(0.5)
pyautogui.click()

\textcolor{purple}{# Failed.}
\end{Verbatim}
\end{tcolorbox}

\begin{tcolorbox}[title={Human > Agent Task: \textit{On page 2, add a note same to the title into the slide. Make the font of title bold.}}]

\begin{Verbatim}[commandchars=\\\{\}]
\textcolor{purple}{# Action 1}
\textcolor{blue}{# Click on the vertical scrollbar to scroll down and reveal more slides}
pyautogui.click(tag_43)  

\textcolor{purple}{# Action 2}
\textcolor{blue}{# Click on the second slide thumbnail to select it}
pyautogui.click(tag_2) 

\textcolor{purple}{# Action 3}
\textcolor{blue}{# Click on the text box to add text}
pyautogui.click(tag_42)  
pyautogui.typewrite(`Team Members', interval=0.1)  
\textcolor{blue}{# Type the title}
pyautogui.hotkey(`ctrl', `a')  
\textcolor{blue}{# Select all text}
pyautogui.hotkey(`ctrl', `b')  
\textcolor{blue}{# Make the text bold}

\textcolor{purple}{# Action 4}
\textcolor{blue}{# Click on the Bold button to apply bold formatting}
pyautogui.click(tag_68)  

\textcolor{purple}{Failed.}
\end{Verbatim}
\end{tcolorbox}

\begin{tcolorbox}[title={Agent > Human Task: \textit{use GIMP to cut out the 2s to 4s part of a video}}]

\small
\begin{Verbatim}[commandchars=\\\{\}]
\textcolor{purple}{# Action 1}
import pyautogui
import time
\textcolor{blue}{# Move the mouse to the assumed terminal icon position}
\textcolor{blue}{at the bottom of the dock and click to open it}
\textcolor{blue}{# Assuming the terminal icon is at the bottom of the dock}
pyautogui.moveTo(70, 1350, duration=1)  
pyautogui.click()
\textcolor{blue}{# Type the command to use ffmpeg to trim the video}
ffmpeg_command = `ffmpeg -ss 00:00:02 -to 00:00:04 -i /home/user/fullvideo.mp4 
-c copy /home/user/trimmedvideo.mp4'
pyautogui.write(ffmpeg_command, interval=0.05)
pyautogui.press('enter')

\textcolor{purple}{# Action 2}
\textcolor{blue}{# Close the search dialog}
pyautogui.press('esc')
\textcolor{blue}{# Open the terminal using a keyboard shortcut (Ctrl+Alt+T is a common shortcut}
\textcolor{blue}{for opening a terminal in many Linux distributions)}
pyautogui.hotkey('ctrl', 'alt', 't')

\textcolor{purple}{# Action 3}
\textcolor{blue}{# Move the mouse to the terminal and click to focus on the terminal window}
pyautogui.moveTo(260 + 722/2, 103 + 434/2, duration=1)  
pyautogui.click()
\textcolor{blue}{# Type the command to use ffmpeg to trim the video}
pyautogui.write(ffmpeg_command, interval=0.05)
pyautogui.press(`enter')

\textcolor{purple}{# Done, but not follow the demanding in instruction.}
\end{Verbatim}
\end{tcolorbox}

\begin{tcolorbox}[title={Agent > Human Task: \textit{I want to understand the resource usage of my Ubuntu system under normal workloads. Please use the `sar` command in the `sysstat` toolkit to monitor system activity, evaluate the status once every second for 30 seconds, output the results to "System Resources Report.txt" under Desktop.}}]
\begin{Verbatim}[commandchars=\\\{\}]
\textcolor{purple}{# Action 1}
import pyautogui
import time
\textcolor{blue}{# Press the keyboard shortcut to open the terminal}
pyautogui.hotkey(`ctrl', `alt', `t')
time.sleep(1)  
\textcolor{blue}{# Wait a moment for the terminal to open}

\textcolor{purple}{# Action 2}
\textcolor{blue}{# Click inside the terminal to focus on it}
pyautogui.click(600, 300)  
\textcolor{blue}{# Assuming the terminal is in the upper left area of the screen}
\textcolor{blue}{n# Type the command to monitor system activity}
pyautogui.typewrite(`sar 1 30 > ~/Desktop/System Resources Report.txt')

\textcolor{purple}{# Action 3}
\textcolor{blue}{# Click on the `System Resources Report.txt' file to select it}
pyautogui.click(1850, 940)  
\textcolor{blue}{# Coordinates taken from the label position and size}
time.sleep(0.5)
\textcolor{blue}{# Right-click to open the context menu}
pyautogui.click(1850, 940, button=`right')
\textcolor{blue}{# Move to the `Open With Other Application' option in the context menu}
pyautogui.move(0, 70)  
\textcolor{blue}{# Assuming `Open With Other Application' is 70 pixels}
\textcolor{blue}{down from the right-click position}
\textcolor{blue}{# Click on `Open With Other Application'}
pyautogui.click()
time.sleep(0.5)
\textcolor{blue}{# Type `gedit' to select the text editor}
pyautogui.typewrite(`gedit')
\textcolor{blue}{# Press `Enter' to open the file with gedit}
pyautogui.press(`enter')

\textcolor{purple}{# Action 4}
\textcolor{blue}{# Click inside the terminal to ensure it's in focus}
pyautogui.click(500, 200)  
\textcolor{blue}{# Coordinates are estimated to be within the terminal window}
\textcolor{blue}{# Type the command to open the file with gedit}
pyautogui.typewrite(`gedit ~/Desktop/System Resources Report.txt')

\textcolor{purple}{Done!}
\end{Verbatim}
\end{tcolorbox}

\begin{tcolorbox}[title={Agent > Human Task: \textit{Hey, my LibreOffice Writer seems to have frozen and I can't get it to close normally. Can you help me force quit the application from the command line? I'm on Ubuntu and I don't want to restart my computer or lose any other work I have open.}}]

\begin{Verbatim}[commandchars=\\\{\}]
\textcolor{purple}{# Action 1}
import pyautogui
import time
\textcolor{blue}{# Open a terminal window}
pyautogui.hotkey(`ctrl', `alt', `t')
time.sleep(1)
\textcolor{blue}{# Type the command to kill all instances of LibreOffice Writer}
pyautogui.typewrite(`killall soffice.bin')

\textcolor{purple}{# Action 2}
\textcolor{blue}{# Open a terminal window}
pyautogui.hotkey(`ctrl', `alt', `t')
time.sleep(1)
\textcolor{blue}{# Type the command to kill all instances of LibreOffice Writer}
pyautogui.typewrite(`killall soffice.bin')

\textcolor{purple}{# Action 3}
\textcolor{blue}{# Ensure the terminal window is focused}
pyautogui.click(500, 200)  
\textcolor{blue}{# Assuming the terminal is at this position}
\textcolor{blue}{# Type the pkill command to forcefully kill the LibreOffice Writer application}
pyautogui.typewrite(`pkill -9 soffice.bin')

\textcolor{purple}{Done!}
\end{Verbatim}
\end{tcolorbox}

\subsection{Comparative Analysis: Claude-3 vs. GPT-4V}
\label{appendix:claude3}

\begin{figure}[htbp]
    \centering
    \includegraphics[width=\linewidth]{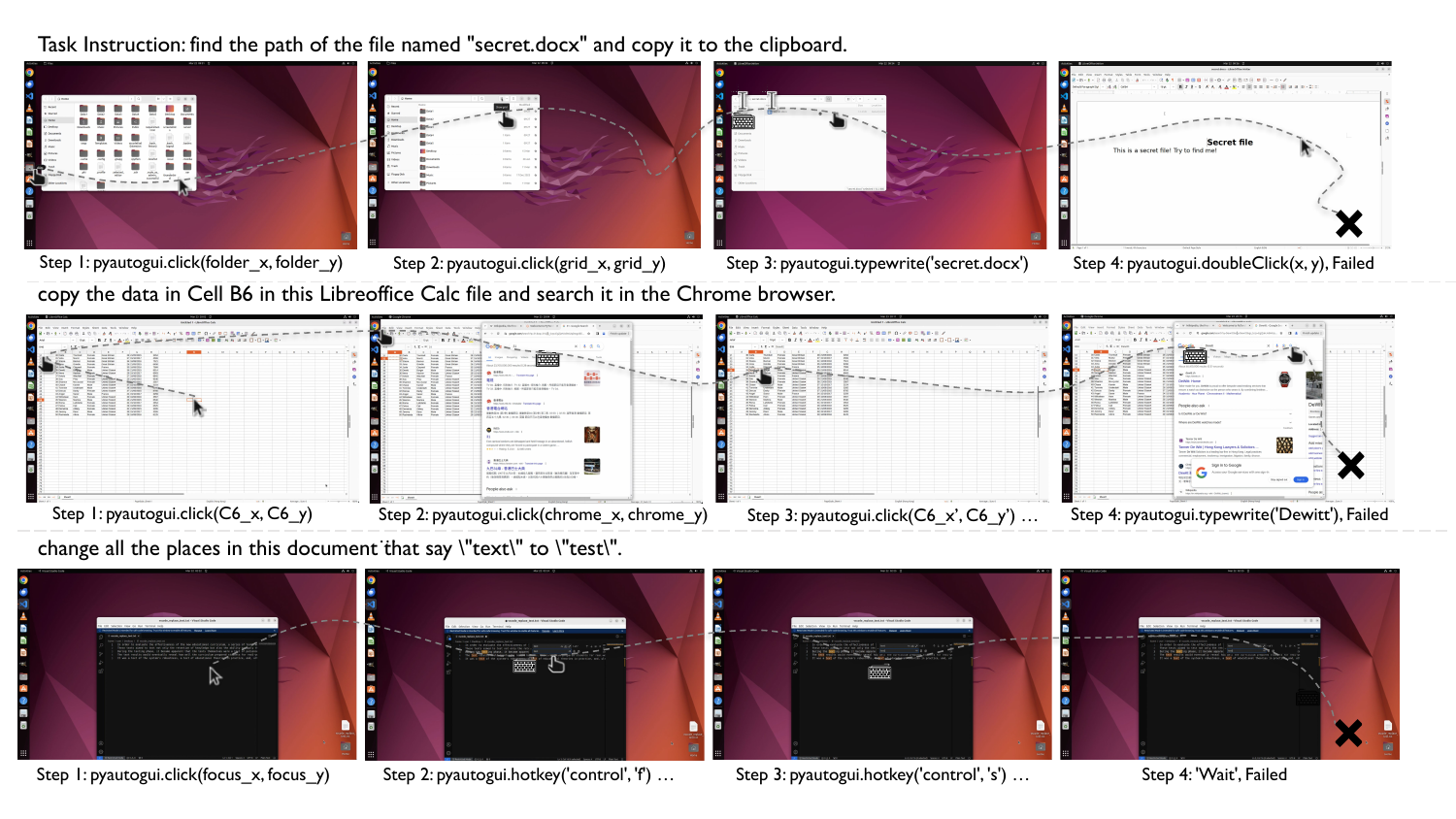}
    \caption{
    Screenshots of error example analysis of Claude-3 Opus.
    }
    \label{fig:claude_error}
\end{figure}

\begin{tcolorbox}[title={Claude3 Failed Task1: \textit{I remember there is a file named "secret.docx" on this computer, but I can't remember where it is. Please find the path where this file is stored and copy it to the clipboard.}}]

\begin{Verbatim}[commandchars=\\\{\}]
\textcolor{purple}{# Action 1}
pyautogui.click(tag_10) 
\textcolor{blue}{# click the Files icon to open the file manager}

\textcolor{purple}{# Action 2}
pyautogui.click(tag_21) 
\textcolor{blue}{# click the search icon to open the search bar}

\textcolor{purple}{# Action 3}
pyautogui.typewrite(`secret.docx') 
\textcolor{blue}{# type the file name to search for}
pyautogui.press(`enter') 
\textcolor{blue}{# press enter to start the search}

\textcolor{purple}{# Action 4}
pyautogui.doubleClick(tag_52) 
\textcolor{blue}{# double click the secret.docx file to select it}
pyautogui.hotkey(`ctrl', 'c') 
\textcolor{blue}{# press Ctrl+C to copy the selected file path to clipboard}

\textcolor{purple}{Failed.}
\end{Verbatim}
\end{tcolorbox}

\begin{tcolorbox}[title={Claude3 Failed Task2: \textit{copy the data in Cell B6 in this Libreoffice Calc file and search it in the Chrome browser.}}]

\begin{Verbatim}[commandchars=\\\{\}]
\textcolor{purple}{# Action 1}
import pyautogui
import time
\textcolor{blue}{# Scroll down to view more search results if needed}
pyautogui.press(`pagedown')
time.sleep(1)
pyautogui.press(`pagedown') 
time.sleep(2)
\textcolor{blue}{# The answer appears to be in the snippet under the first search result}
\textcolor{blue}{# `TV 31. u76f4u64adu4e2d.' which translates to `TV 31. Live.'}
\textcolor{blue}{# So channel 31 in Hong Kong is RTHK TV 31}
print(`Found the answer. Channel 31 in Hong Kong is RTHK TV 31.')
\textcolor{blue}{# The original task of copying cell B36 from the spreadsheet}
\textcolor{blue}{and searching it in the browser is complete}
print(`Task complete.')

\textcolor{purple}{# Action 2}
\textcolor{blue}{# Activate the Chrome browser window}
pyautogui.click(tag_3)
time.sleep(1)
\textcolor{blue}{# Click on the address/search bar}
pyautogui.click(tag_30)
\textcolor{blue}{# Paste the copied value `Dewitt'}
pyautogui.hotkey(`ctrl', `v')
\textcolor{blue}{# Press Enter to perform the search}
pyautogui.press(`enter')
print(`Searched for `Dewitt' in the Chrome browser.')

\textcolor{purple}{# Action 3}
print(`DONE')

\textcolor{purple}{# Action 4}
print(`DONE')

\textcolor{purple}{Failed.}
\end{Verbatim}
\end{tcolorbox}

\begin{tcolorbox}[title={Claude3 Failed Task3: \textit{Please help me change all the places in this document that say `text' to `test'.}}]

\begin{Verbatim}[commandchars=\\\{\}]
\textcolor{purple}{# Action 1}
pyautogui.hotkey(`ctrl', `h')
pyautogui.write(`text')
pyautogui.press(`tab')
pyautogui.write(`test')
pyautogui.press(`enter')

\textcolor{purple}{# Action 2}
pyautogui.press(`enter')

\textcolor{purple}{# Action 3}
pyautogui.hotkey(`ctrl', `s')

\textcolor{purple}{Failed.}
\end{Verbatim}
\end{tcolorbox}

\subsection{Analysis on SoM Effectiveness}

\begin{tcolorbox}[title={Task w/o SoM: \textit{turning off the code checking function of the python plugin in vscode}}]

\small
\begin{Verbatim}[commandchars=\\\{\}]
\textcolor{purple}{# Action 1}
import pyautogui
import time
\textcolor{blue}{# Move the mouse to the gear icon (Manage) and click it}
pyautogui.moveTo(472, 1034, duration=0.5) 
pyautogui.click()

\textcolor{purple}{# Action 2}
\textcolor{blue}{# Move the mouse to the \"Settings\" menu item and click it}
\textcolor{blue}{# A rough central point of the `Settings' menu item}
pyautogui.moveTo(497 + 385/2, 832 + 25/2, duration=0.5)  
pyautogui.click()

\textcolor{purple}{# Action 3}
\textcolor{blue}{# Move the mouse to the search settings bar and click it}
\textcolor{blue}{# A rough central point of the search settings bar}
pyautogui.moveTo(783 + 587/2, 390 + 19/2, duration=0.5)  
pyautogui.click()
\textcolor{blue}{# Type the query to search for Python-related settings}
pyautogui.write('python', interval=0.05)

\textcolor{purple}{# Action 4}
\textcolor{blue}{# Move the mouse to the `Edit in settings.json' button and click it}
\textcolor{blue}{# A rough central point of the button}
pyautogui.moveTo(990 + 116/2, 624 + 15/2, duration=0.5)  
pyautogui.click()

\textcolor{purple}{# Action 5}
\textcolor{blue}{# Move the cursor to the correct position}
pyautogui.moveTo(889 + 1/2, 432 + 1/2, duration=0.5)  
pyautogui.click()
\textcolor{blue}{# Type the configuration to disable Python missing imports error reporting}
pyautogui.write(`reportMissingImports: `none'', interval=0.05)
pyautogui.press(`enter')
\textcolor{blue}{# Save the changes}
pyautogui.hotkey(`ctrl', `s')

\textcolor{purple}{# Done}
\end{Verbatim}

\end{tcolorbox}

\begin{tcolorbox}[title={Task w/ SoM: \textit{  turning off the code checking function of the python plugin in vscode}}]
    
\small
\begin{Verbatim}[commandchars=\\\{\}]
\textcolor{purple}{# Action 1}
import pyautogui
import time
pyautogui.hotkey(`ctrl', `shift', `p')

\textcolor{purple}{# Action 2}
pyautogui.write(`Preferences: Open Settings (UI)', interval=0.05)
time.sleep(0.5)
pyautogui.press(`enter')

\textcolor{purple}{# Action 3}
\textcolor{blue}{# Click on the search settings input field}
pyautogui.click(x=600, y=175)  
\textcolor{blue}{# These coordinates may need to be adjusted for your screen resolution}
\textcolor{blue}{# Type `Python' into the search field}
pyautogui.write(`Python', interval=0.05)
pyautogui.press(`enter')

\textcolor{purple}{# Action 4}
\textcolor{blue}{# Click on the search settings bar and type in the query again}
pyautogui.click(x=800, y=150)  
\textcolor{blue}{# Click on the search bar, the coordinates may need adjustment}
pyautogui.typewrite(`Python missing imports', interval=0.25)  

\textcolor{blue}{# ... Can not find the right click area}

\textcolor{purple}{# Failed}
\end{Verbatim}

\end{tcolorbox}







\subsection{Analysis on SoM Effectiveness}
\begin{figure}[htbp]
    \centering
    \vspace{-5pt}
    \includegraphics[width=\linewidth]{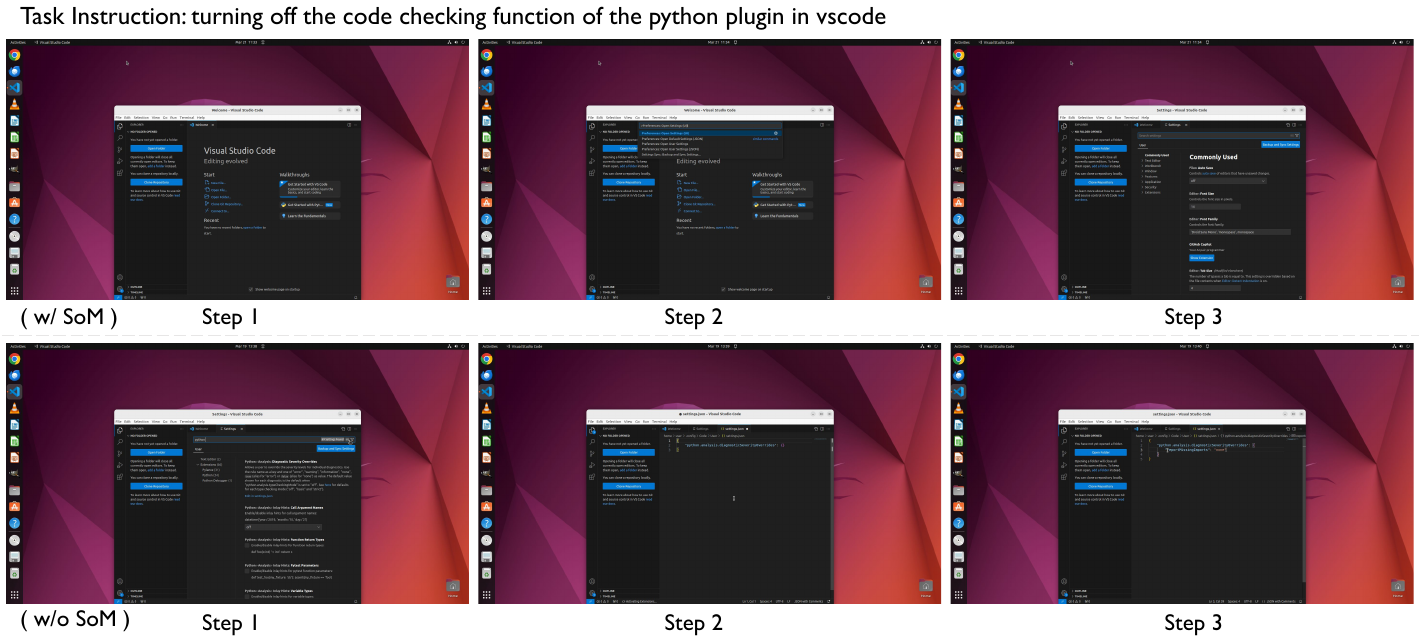}
    \caption{
    A task showcase where the SoM-equipped agent tried to find interactive settings, while the screen-a11ytree-equipped agents directly modified the value in the setting.json file.
    }
    \label{fig:som_analysis}
    \vspace{-10pt}
\end{figure}

The SoM, effective for web tasks focusing only on labeled bounding boxes, \textbf{did not} excel in our diverse tasks that required more expertise and flexibility. It shortens the action space of the agent and thus hinders the agent's exploration and adaptability.
For example in Figure \ref{fig:som_analysis}, in the task of "turning off the code checking function of the python plugin." in VS Code, agents without SoM succeeded by editing the settings.json, whereas SoM-equipped agents struggled with finding and deselecting the checkbox in the settings. 
The latter has a longer action path and fails due to incorrect mouse clicks. 
Furthermore, SoM's labeling of elements in professional software also has such errors, that agents without SoM can use a11y tree to calculate and adjust coordinates, while the elements' blocks under SoM can be hard to change during the task.

\end{document}